\pdfoutput=1
\documentclass[twoside,11pt]{article}

\usepackage[preprint]{dmlr2e}
\usepackage[T1]{fontenc}

\usepackage{booktabs}
\usepackage{multirow}
\usepackage{subfig}
\usepackage{amsmath}
\usepackage{amsfonts}
\usepackage{nicefrac}
\usepackage{xfrac}
\usepackage[table]{xcolor}
\usepackage{microtype}
\usepackage{url}
\usepackage{lastpage}
\usepackage{wrapfig}

\usepackage{upquote}   %
\usepackage{listings}
\usepackage[most]{tcolorbox}

\lstdefinestyle{promptstyle}{%
  basicstyle=\ttfamily\footnotesize,
  breaklines=true,
  breakautoindent=false,
  breakindent=4em,   %
  columns=flexible,
  keepspaces=true,
  frame=none,
  aboveskip=0pt,
  belowskip=0pt,
  xleftmargin=0pt,
  xrightmargin=0pt,
}

\newtcolorbox{promptbox}[1][]{%
  breakable,
  colback=black!4,
  colframe=black!25,
  colbacktitle=black!10,
  coltitle=black,
  fonttitle=\bfseries\small,
  boxrule=0.4pt,
  arc=2pt,
  left=6pt, right=6pt, top=6pt, bottom=6pt,
  #1}

\usepackage[section]{placeins}

\newcommand{\rvlcdipnSize}{1,002}
\newcommand{\rvlcdipoSize}{2,915}
\newcommand{\totalSize}{3,917}
\newcommand{\errorRate}{10.5\%}      %
\newcommand{\dupRate}{35.2\%}        %
\newcommand{\nModels}{24}           %
\newcommand{\gradient}[1]{\cellcolor{blue!#1!white}#1}

\DeclareRobustCommand{\cat}[1]{{\ttfamily\def\_{\textunderscore\allowbreak}#1}}

\dmlrheading{1}{2026}{1-\pageref{LastPage}}{X/26}{X/26}{26-0000}{Larson and Leach}
\ShortHeadings{Data Quality in RVL-CDIP}{Larson and Leach}
\firstpageno{1}

\begin{document}

\title{Data Quality in RVL-CDIP: A Systematic Audit}

\author{\name Stefan Larson \email stefan.larson@vanderbilt.edu \\
        \addr Department of Computer Science \\
        Vanderbilt University \\
        Nashville, TN 37235, USA
        \AND
        \name Kevin Leach \email kevin.leach@vanderbilt.edu \\
        \addr Department of Computer Science \\
        Vanderbilt University \\
        Nashville, TN 37235, USA}

\editor{TBD}

\maketitle

\begin{abstract}%
RVL-CDIP is the \emph{de facto} standard benchmark for document classification, and reported accuracy on it is now 98.8\%.
In this paper, we audit RVL-CDIP for data quality issues and find that approximately \errorRate\ of its 400,000 documents have erroneous labels and that \dupRate\ of the test set has a duplicate or near-duplicate in the training set.
Identification codes on the pages are sufficient to classify at 69\% accuracy, and a state-of-the-art model loses 11 points without them.
Image-only models trained on RVL-CDIP lose 16 to 28 points on documents from other sources, and the choice of OCR engine changes accuracy by 4.1 to 8.3 points.
We find that nearly 16,000 RVL-CDIP documents contain sensitive personal data, and produce de-identified versions of these documents.
We re-benchmark 25 models, and with all defects remediated, the 13 that we retrain are 0.9 to 2.9 points more accurate than in the original condition; the most accurate of them, LayoutLMv3, is 96.5\% accurate.
We release the corrected corpus, the annotations and the out-of-distribution data.
\end{abstract}

\begin{keywords}
  document classification, data quality, benchmark auditing, label errors, data leakage, spurious correlations, de-identification, data privacy
\end{keywords}

\section{Introduction}\label{sec:intro}

The RVL-CDIP corpus \citep{harley2015icdar-rvlcdip} was introduced in 2015 as a computer vision benchmark and has since become what prior work has called the \emph{de facto} standard benchmark for document page classification \citep{larson-etal-2023-evaluation}, a description that our own survey of the literature confirms.
Its 400,000 labeled document images across 16 categories made it large enough to train deep architectures at a time when comparable resources did not exist, and the community adopted it accordingly.
Modern document understanding models, among them LayoutLMv3 \citep{layoutlmv3-huang-2022}, Donut \citep{kim2022donut}, and UDOP \citep{tang-2023-udop}, report document classification results on RVL-CDIP alone.
Of the 34 papers we surveyed that report document classification results on RVL-CDIP, 23 evaluate classification on that dataset and nothing else, and 29 evaluate only on datasets derived from the archive from which RVL-CDIP was sampled (Section~\ref{sec:background-leaderboard}).
Of the five that use a dataset from another source, two use proprietary in-house collections that cannot be reproduced.
The other three all use NIST tax forms (Special Database 2 or 6, two sets of the same 1988 IRS forms), on which they report 99.99 to 100\% accuracy; these datasets are therefore saturated.
RVL-CDIP is the most popular benchmark for this task and effectively the only standardized benchmark of comparable scale, so that a claim about document classification performance is in practice usually a claim based on RVL-CDIP.

Performance on this benchmark has increased steadily and now appears close to saturation.
Early convolutional models reported accuracies between 88\% and 91\% \citep{afzal-2017}; contemporary multi-modal architectures report the high 90s, with LayoutLLM \citep{fujitake-2024-layoutllm} reporting 98.8\%.
Progress of this kind is normally interpreted as evidence that the underlying task is being solved, which depends on an assumption that is rarely examined: that the benchmark measures what it is taken to measure, and that successive improvements on it are improvements in document understanding and not an increasingly effective use of the properties of the benchmark itself.

There is good reason to examine that assumption: a body of work across natural language processing and computer vision has found that widely used benchmarks frequently contain substantial label noise \citep{pervasive-label-errors-2021}, significant overlap between test and training data \citep{allamanis-2019-adverse-duplication, elangovan-etal-2021-memorization}, and collection artifacts that permit models to shortcut the intended task \citep{gururangan-etal-2018-annotation, mccoy-etal-2019-right, niven-kao-2019-probing}.
RVL-CDIP is a plausible candidate for such problems, because its documents are sampled from IIT-CDIP \citep{iit-cdip-2006}, a collection of roughly seven million documents released through litigation against the American tobacco industry, which was assembled for legal discovery and was never curated for machine learning.
Harley et al. themselves noted that the source collection had ``missing or erroneous tags'' and acknowledged that ``the final categories are not perfectly distinct'', but did not publish category definitions against which those concerns could be assessed.

This paper reports a systematic audit of RVL-CDIP.
Our central finding is that the benchmark is compromised along several independent dimensions simultaneously, and that a reported accuracy on RVL-CDIP is therefore a composite quantity: partly a measurement of document classification ability, and partly a measurement of a model's ability to learn the benchmark's defects.
We organize the audit around five axes, four of which concern the dataset itself: the validity of its labels, the integrity of its train and test splits, the presence of spurious visual cues, and the narrowness of its distribution.
The fifth concerns the evaluation protocol: because RVL-CDIP is distributed as images without a text layer, every text-based and multi-modal model must obtain its text from an optical character recognition engine of its own choice, and the choice of OCR engine, which is uncontrolled across the literature, has a larger effect on accuracy than the architectural differences on which these models are compared.

One further defect is separate from these five, and we treat it first: RVL-CDIP contains a large quantity of sensitive personal data (Social Security numbers, dates and places of birth, home addresses), which does not affect what a score on a document classification benchmark means but does determine what can responsibly be released.
We quantify that exposure, de-identify the affected documents by replacing each sensitive entity with a synthetic but plausible one, and show that doing so changes none of the predictions of a DiT classifier on 445 resumes (Section~\ref{sec:deid}).
Because de-identification has almost no effect on classification, we run the entire audit on the de-identified corpus, so that the version of RVL-CDIP we measure model performance on is the same version that we release publicly.

\begin{figure}[t]
  \centering
  \includegraphics[width=\linewidth]{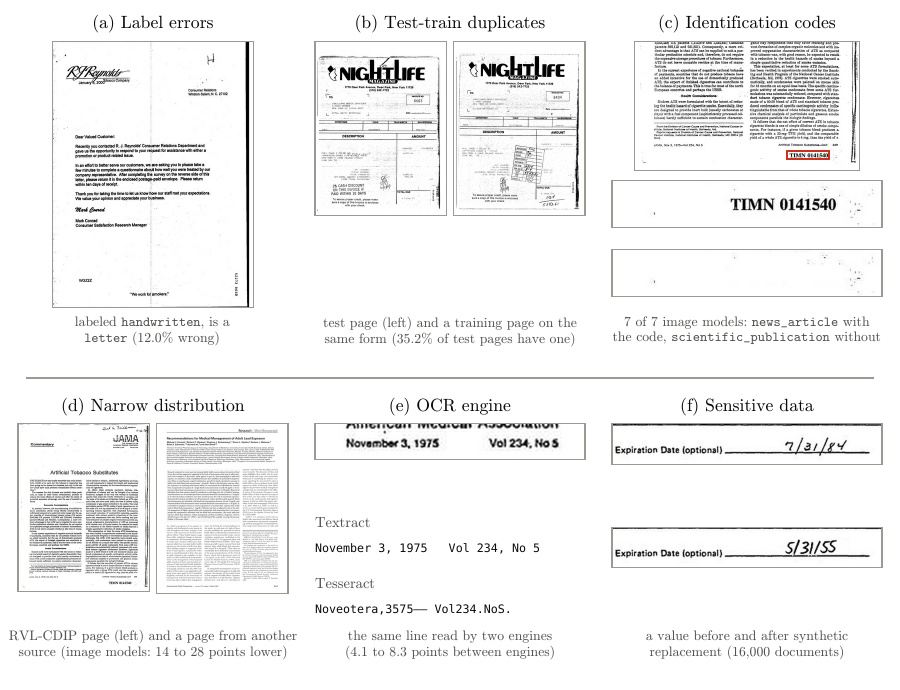}
  \caption{The defects of RVL-CDIP examined in this paper, each with an example
           page and the corresponding result. The addressee of the letter in
           (a) is masked, and the image models in (c) are those trained on the
           original labels.}
  \label{fig:teaser}
\end{figure}

Figure~\ref{fig:teaser} shows an example page of each defect together with the corresponding result.
Each of these defects has, in isolation, a bounded effect that could be dismissed as a known limitation.
Considered together they raise a question that cannot be answered by studying any one of them: what does an RVL-CDIP score measure when all of them are controlled at once?
We construct a corpus in which all five axes are remediated at once and re-benchmark \nModels{} models against it.
We also test whether the defects are independent of one another, since defects that co-occur in the same documents cannot be corrected one at a time and their effects on accuracy are not additive.

This paper is a benchmark audit: we do not propose a new model or a new training procedure.
It consolidates and substantially extends five of our earlier conference papers \citep{larson-etal-2022-rvlcdipn, larson-etal-2023-evaluation, larson-etal-2024-de, larson-etal-2025-spurious, larson-etal-2026-revising}, each of which examined a single axis in isolation; the joint remediation and defect-interaction analyses and the OCR protocol analysis are new here.

\noindent We make the following contributions:
\begin{enumerate}
  \item An audit of sensitive personal data in RVL-CDIP, finding 15,956 affected documents (4.0\% of the corpus) including 2,342 US Social Security numbers, and a de-identification pipeline, based on synthetic replacement, that changes none of the predictions of a DiT classifier on 445 resumes and on whose output the remainder of this paper's experiments are run (Section~\ref{sec:deid}).
  \item An exhaustive, 400,000-document review of RVL-CDIP establishing a label error rate of approximately \errorRate, nearly nine times the error rate of the leading models, with per-category rates from 1.5\% (\cat{resume}) to 31.3\% (\cat{letter}). The rate is measured against published category criteria, absent from the original release (Sections~\ref{sec:guidelines} and~\ref{sec:axis1}).
  \item A filter-and-refine method for detecting test--train overlap in scanned document collections, finding that approximately \dupRate\ of RVL-CDIP's test set has a near-duplicate counterpart in training. The validation split shows nearly the same rate, which indicates that the leakage was introduced when the corpus was assembled and not when it was split (Section~\ref{sec:axis2}).
  \item A quantification of litigation ID codes as spurious cues: these codes appear on 92\% of the corpus, and a classifier given only features of the codes classifies RVL-CDIP documents with 69\% accuracy, and a state-of-the-art model loses 11.0 accuracy points when the codes are removed. We release an automatic detector and code-free versions of RVL-CDIP (Section~\ref{sec:axis3}).
  \item Two out-of-distribution evaluation sets, RVL-CDIP-\emph{N} and RVL-CDIP-\emph{O}, on which RVL-CDIP-trained image-only models lose 14 to 28 accuracy points, against 3 to 4 for text models, and the image-only models' ability to separate in-domain from out-of-domain input decreases toward chance (Section~\ref{sec:axis4}).
  \item An analysis of OCR engine choice as an uncontrolled confound in the multi-modal document classification literature, where holding architecture fixed and varying only the engine changes accuracy by 4.1 to 8.3 points, against 0.19 to 0.82 points between successive LayoutLM generations (Section~\ref{sec:axis5}).
  \item A joint remediation experiment measuring model performance with all confounds controlled simultaneously (Section~\ref{sec:joint}), and a defect-interaction analysis establishing whether they are independent (Section~\ref{sec:interaction}).
  \item Public release of all annotations, corrected corpus variants, and evaluation data (Appendix~\ref{sec:artifacts}).
\end{enumerate}

The paper is organized as follows: Section~\ref{sec:background} describes RVL-CDIP, the collection from which it is derived and the benchmark-auditing literature; Section~\ref{sec:deid} quantifies the corpus's sensitive personal data and defines the de-identified corpus on which every later measurement is made; Section~\ref{sec:guidelines} provides the category criteria that the original release omitted, on which every subsequent measurement depends; Sections~\ref{sec:axis1} through~\ref{sec:axis5} examine the five axes in turn; and Sections~\ref{sec:joint} and~\ref{sec:interaction} report the joint remediation and defect-interaction analyses.

\section{Background and Related Work}\label{sec:background}

In this section, we first describe RVL-CDIP itself: its origin in the IIT-CDIP litigation archive, its composition and standard splits (Section~\ref{sec:background-rvlcdip}), and the accuracies reported against it (Section~\ref{sec:background-leaderboard}).
We then review the prior work on which our audit is based, covering label errors and test--train duplication, spurious correlations, out-of-distribution evaluation, and sensitive data and de-identification (Section~\ref{sec:background-related}).

\subsection{RVL-CDIP and IIT-CDIP}\label{sec:background-rvlcdip}

\begin{figure}[tbp]
  \centering
  \includegraphics[width=\linewidth]{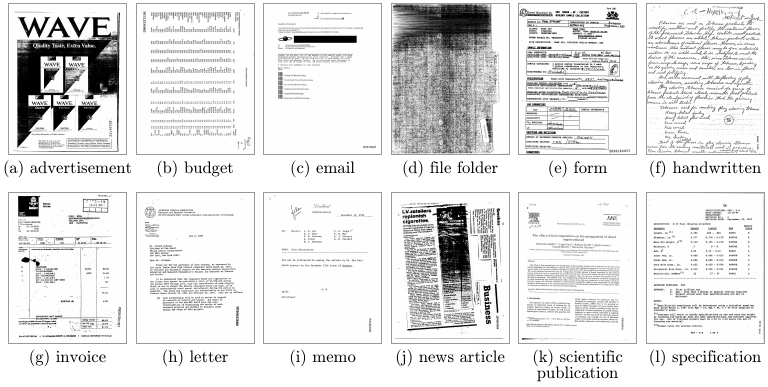}
  \caption{Example pages from twelve of the 16 RVL-CDIP categories, taken from the de-identified test split.}
  \label{fig:category-examples}
\end{figure}

\begin{sloppypar}
RVL-CDIP \citep{harley2015icdar-rvlcdip} consists of 400,000 grayscale document images distributed evenly across 16 categories: 
\cat{advertisement}, \cat{budget}, \cat{email}, \cat{file\_folder}, \cat{form}, \cat{handwritten}, \cat{invoice}, \cat{letter}, \cat{memo}, \cat{news\_article}, \cat{presentation}, \cat{questionnaire}, \cat{resume}, \cat{scientific\_publication}, \cat{scientific\_report}, and \cat{specification}.
Each image is a single page, and in most cases the first page of its source document; Figure~\ref{fig:category-examples} shows example pages from twelve of the categories.
The standard split allocates 320,000 documents to training and 40,000 each to validation and test; per-category counts are given in Table~\ref{tab:rvlcdip-statistics}.
The benchmark is used in two settings: supervised models are trained on the training split and evaluated on the test split, while zero-shot models are evaluated on the test split alone.
\end{sloppypar}

\begin{wraptable}{R}{0.47\textwidth}
  \centering
  \footnotesize
  \setlength{\tabcolsep}{3pt}
  \caption{RVL-CDIP per-category train, validation, and test counts.}
  \label{tab:rvlcdip-statistics}
  \begin{tabular}{lccc}
  \toprule
  \textbf{Category} & \textbf{Train} & \textbf{Val.} & \textbf{Test} \\
  \midrule
  \cat{advertisement}          & 19,963 & 2,522 & 2,515 \\
  \cat{budget}                 & 20,010 & 2,485 & 2,505 \\
  \cat{email}                  & 19,954 & 2,530 & 2,516 \\
  \cat{file\_folder}           & 20,022 & 2,451 & 2,527 \\
  \cat{form}                   & 19,957 & 2,537 & 2,506 \\
  \cat{handwritten}            & 20,034 & 2,434 & 2,532 \\
  \cat{invoice}                & 19,947 & 2,576 & 2,477 \\
  \cat{letter}                 & 20,106 & 2,430 & 2,464 \\
  \cat{memo}                   & 19,975 & 2,533 & 2,492 \\
  \cat{news\_article}          & 20,011 & 2,526 & 2,463 \\
  \cat{presentation}           & 20,043 & 2,468 & 2,489 \\
  \cat{questionnaire}          & 20,048 & 2,517 & 2,435 \\
  \cat{resume}                 & 20,037 & 2,426 & 2,537 \\
  \cat{scientific\_publication} & 19,902 & 2,526 & 2,572 \\
  \cat{scientific\_report}     & 19,994 & 2,508 & 2,498 \\
  \cat{specification}          & 19,997 & 2,531 & 2,472 \\
  \bottomrule
  \end{tabular}
\end{wraptable}

The documents are sampled from IIT-CDIP \citep{iit-cdip-2006}, a collection of approximately seven million documents (some 40 million pages) made public as a result of litigation and settlement agreements involving several US tobacco companies.
This provenance has three properties that are relevant to the audit: the documents are overwhelmingly scans of physical paper, with few born-digital files; they date from roughly the 1950s to 2006, the year IIT-CDIP was released, and concern a single industry; and they were assembled for legal discovery and indexed retrospectively, and were not selected or curated for machine learning evaluation.
One consequence is that the research community has been benchmarking against a dataset for which no authoritative statement of what each category contains has ever existed, which Section~\ref{sec:guidelines} addresses.

Other datasets from the same collection include DocVQA \citep{docvqa-2021}, FUNSD \citep{jaume2019-funsd}, Tobacco3482 \citep{kumar-2013-tobacco3482}, and Tobacco800 \citep{zhu-2007-tobacco800}, which are taken wholly or in part from IIT-CDIP or its parent library and are used for visual question answering, form understanding, classification, and signature detection respectively.
Our focus throughout this paper is RVL-CDIP, and we refer to these other datasets where a defect that we identify also occurs in them, which is principally the case for sensitive personal data (Section~\ref{sec:deid}) and for ID codes (Section~\ref{sec:axis3}).

\subsection{Reported Performance on RVL-CDIP}\label{sec:background-leaderboard}

Table~\ref{tab:model-scores} lists reported accuracy on RVL-CDIP across the literature, with the modalities that each model uses: image, text, and layout.
Early image-only convolutional models reported accuracies between 88\% and 91\% \citep{afzal-2017}.
The incorporation of text and layout signals, and subsequently of transformer architectures pre-trained on document data, raised reported accuracy to the mid 90s.
To our knowledge, the current state of the art is LayoutLLM at 98.80\% \citep{fujitake-2024-layoutllm}, followed by EAML at 97.70\% \citep{eaml-bakkali-2021} and Bi-VLDoc at 97.17\% \citep{bi-vldoc-2025-luo}.

\begin{table}[tbp]
  \centering
  \caption{Reported accuracy on RVL-CDIP by model and modality (I: image,
           T: text, L: layout), ordered by reported score.}
  \label{tab:model-scores}
\scalebox{0.7}{
  \setlength{\tabcolsep}{4pt}
  \begin{tabular}{llc@{\hspace{2em}}llc}
  \toprule
  \textbf{Model (Reported by)} & \textbf{Mod.} & \textbf{Acc.} &
  \textbf{Model (Reported by)} & \textbf{Mod.} & \textbf{Acc.} \\
  \midrule
  LayoutLLM \citep{fujitake-2024-layoutllm}          & I, T, L & 98.80 & MGDoc \citep{mgdoc-2022-emnlp}                        & I, T, L & 93.64 \\
  EAML \citep{eaml-bakkali-2021}                     & I, T    & 97.70 & Dessurt \citep{dessurt-2022}                          & I       & 93.60 \\
  Bi-VLDoc \citep{bi-vldoc-2025-luo}                 & I, T, L & 97.17 & Bigbird-base \citep{pham-2022-long}                   & T       & 93.48 \\
  ERNIE-Layout-large \citep{ernie-layout-2022}       & I, T, L & 96.27 & \citet{pramanik-2022}                                 & I, T, L & 93.36 \\
  UDOP-Dual \citep{tang-2023-udop}                   & I, T, L & 96.22 & Bigbird-large \citep{pham-2022-long}                  & T       & 93.34 \\
  DocFormer-base \citep{Appalaraju_2021_ICCV-docformer} & I, T, L & 96.17 & Multimodal Ensemble \citep{dauphinee-2019}         & I, T    & 93.07 \\
  StructuralLM-large \citep{li-etal-2021-structurallm} & T, L  & 96.08 & SelfDoc \citep{selfdoc-2021}                          & I, T, L & 92.81 \\
  UDOP \citep{tang-2023-udop}                        & I, T, L & 96.00 & LadderNet \citep{laddernet-2019}                      & I       & 92.77 \\
  LayoutLMv3-large \citep{layoutlmv3-huang-2022}     & I, T, L & 95.93 & \citet{Zingaro2021}                                   & I, T    & 92.70 \\
  LiLT-base \citep{wang-etal-2022-lilt}              & T, L    & 95.68 & DiT-large \citep{li-2022-dit-image-transformer}                           & I       & 92.69 \\
  LayoutLMv2-large \citep{xu-etal-2021-layoutlmv2}   & I, T, L & 95.65 & VLCDoC \citep{vlcdoc}                                 & I, T    & 92.64 \\
  BROS-base \citep{wang-etal-2022-lilt}              & T, L    & 95.58 & InceptionResNetV2 \citep{xu-etal-2021-layoutlmv2}     & I       & 92.63 \\
  TILT-large \citep{TILT-2021}                       & I, T, L & 95.52 & EfficientNet \citep{ferrando-2020}                    & I       & 92.31 \\
  DocFormer-large \citep{Appalaraju_2021_ICCV-docformer} & I, T, L & 95.50 & Region Ensemble \citep{das-2018}                  & I       & 92.21 \\
  LayoutLMv3-base \citep{layoutlmv3-huang-2022}      & I, T, L & 95.44 & DiT-base \citep{li-2022-dit-image-transformer}                            & I       & 92.11 \\
  Donut \citep{kim2022donut}                         & I, T    & 95.30 & MAE-base \citep{li-2022-dit-image-transformer}                            & I       & 91.42 \\
  Wukong-Reader-large \citep{wukong-reader-2022}     & I, T, L & 95.26 & Stacked CNN Single \citep{das-2018}                   & I       & 91.11 \\
  \citet{pham-2022-long}                             & T, L    & 95.25 & BEiT-base \citep{li-2022-dit-image-transformer}                           & I       & 91.09 \\
  TILT-base \citep{TILT-2021}                        & I, T, L & 95.25 & VGG-16 \citep{afzal-2017}                             & I       & 90.97 \\
  LayoutLMv2-base \citep{xu-etal-2021-layoutlmv2}    & I, T, L & 95.25 & \citet{csurka-2016}                                   & I       & 90.70 \\
  UDoc-star \citep{udoc-2021}                        & I, T, L & 95.05 & ResNext-101 \citep{li-2022-dit-image-transformer}                         & I       & 90.65 \\
  Wukong-Reader-base \citep{wukong-reader-2022}      & I, T, L & 94.91 & \citet{audebert-2019}                                 & I, T    & 90.60 \\
  StrucTexTv2-large \citep{yu2023structextv}         & I       & 94.62 & ResNet-50 \citep{afzal-2017}                          & I       & 90.40 \\
  LayoutLMv1-base \citep{xu-2020-layoutlm}           & I, T, L & 94.43 & RoBERTa-large \citep{li-etal-2021-structurallm}       & T       & 90.11 \\
  LayoutLMv1-large \citep{xu-2020-layoutlm}          & I, T, L & 94.42 & RoBERTa-base \citep{li-etal-2021-structurallm}        & T       & 90.06 \\
  MATrIX \citep{matrix-2022}                         & I, T, L & 94.20 & BERT-large \citep{li-etal-2021-structurallm}          & T       & 89.92 \\
  DocXClassifier-xl \citep{Saifullah2024-docxclassifier} & I   & 94.17 & BERT-base \citep{li-etal-2021-structurallm}           & T       & 89.81 \\
  DocXClassifier-large \citep{Saifullah2024-docxclassifier} & I & 94.15 & \citet{tensmeyer-2017}                              & I       & 89.31 \\
  DocXClassifier-base \citep{Saifullah2024-docxclassifier} & I  & 94.00 & GoogLeNet \citep{afzal-2017}                        & I       & 89.02 \\
  UDoc \citep{udoc-2021}                             & I, T, L & 93.96 & AlexNet \citep{afzal-2017}                            & I       & 88.60 \\
  Longformer-base \citep{pham-2022-long}             & T       & 93.85 &                                                       &         &       \\
  Longformer-large \citep{pham-2022-long}            & T       & 93.73 &                                                       &         &       \\
  \bottomrule
  \end{tabular}}
\end{table}

Two features of this table are relevant to the rest of the paper.
First, the differences between the highest-ranked models are very small: the improvement from LayoutLMv2-base to LayoutLMv3-base is 0.19 accuracy points, and from LayoutLMv1-base to LayoutLMv2-base 0.82 points, so that claims of progress in this literature are routinely based on differences of well under a point.
Second, the error rate of the leading models (1.2\% for LayoutLLM) is now about one ninth of the rate of label error we measure in the benchmark itself (Section~\ref{sec:axis1}).
A model cannot be correct more often than the labels against which it is scored, so the differences between the highest-ranked models do not appear to be differences in classification accuracy in any straightforward sense.

The papers behind Table~\ref{tab:model-scores} evaluate document classification almost exclusively on RVL-CDIP.
Table~\ref{tab:survey-datasets} in Appendix~\ref{app:survey-datasets} records, for each of the 34 papers cited there, which datasets it evaluates document classification on, how many datasets it evaluates on for every other task, and what document corpus it pre-trains on.
Twenty-three of the 34 use RVL-CDIP and nothing else for classification.
Eight use Tobacco3482 \citep{kumar-2013-tobacco3482}, which is taken from the same litigation archive; for six of them that is the only addition, so 29 of the 34 evaluate classification wholly on datasets derived from IIT-CDIP.
Five report classification results on a dataset that is not derived from IIT-CDIP, and for two of those the dataset is private and cannot be reproduced \citep{tensmeyer-2017, pramanik-2022}.
Three papers remain (\citealt{csurka-2016}, \citealt{laddernet-2019}, and \citealt{vlcdoc}) whose classification results can be checked against public data of independent origin.
In contrast, the same papers routinely evaluate information extraction and question answering across four, five, or thirteen datasets, while using a single dataset for classification.
Twenty-two of the 34 also pre-train on IIT-CDIP or a corpus derived from it, so for most of this leaderboard the evaluation set and the pre-training corpus originate in the same collection of tobacco litigation documents.

\subsection{Related Work}\label{sec:background-related}

\paragraph{Label errors and test--train overlap.}
Our work belongs to an established literature that studies benchmark datasets empirically and does not assume that they are fixed ground truth.
Label errors have been documented in text, image and document corpora alike \citep{rucker-akbik-2023-cleanconll, pervasive-label-errors-2021, vu-2020-funsd-quality, lim-2025-tobacco3482-quality}, and \citet{pervasive-label-errors-2021} found that correcting test set labels can reorder model leaderboards, so label quality is not a peripheral concern.
A parallel line of work examines overlap between evaluation and training data: \citet{allamanis-2019-adverse-duplication} showed that duplication inflates reported performance because part of the test set was effectively available during training, and the same holds in question answering \citep{lewis-etal-2021-question} and document information extraction \citep{laatiri-2023-funsd-sroie-overlap}.
Others have argued more generally that meaningful evaluation requires data that was not available in training \citep{elangovan-etal-2021-memorization}.
Sections~\ref{sec:axis1} and~\ref{sec:axis2} examine label validity and split integrity respectively.

\paragraph{Spurious correlations and shortcut learning.}
Models frequently achieve strong benchmark performance by exploiting features that correlate with the target label but are irrelevant to the intended task, as documented in natural language inference \citep{gururangan-etal-2018-annotation, mccoy-etal-2019-right}, image classification \citep{neuhaus-2023-cvpr-spurious-imagenet}, and medical imaging \citep{bissoto-2019-deconstructing-bias}.
A shortcut is by definition a real statistical regularity of the dataset from which it was learned, and it is useless, or misleading, on data from any other source.
Closest to our setting, \citet{saifullah-2024-explainability} applied explainability methods to document classifiers trained on RVL-CDIP and Tobacco3482 and observed that identification codes stamped on the documents were often highly relevant features, an observation that they did not pursue further and that Section~\ref{sec:axis3} examines in depth.

\paragraph{Out-of-distribution evaluation.}
Evaluating beyond the training distribution involves two distinct problems: classifying inputs that belong to the label set but have a different source, and abstaining on inputs that do not belong to any label.
Benchmarks for both exist in other domains, among them WILDS \citep{koh2021wilds} and corruption benchmarks such as ImageNet-C \citep{hendrycks-imagenet-c-2019} for the first, and CLINC150 \citep{larson-clinc150} and natural adversarial inputs \citep{hendrycks-2021-natural-adversarial-examples} for the second.
Prior to our own work \citep{larson-etal-2022-rvlcdipn}, no published evaluation data existed for either setting in document classification, and RVL-CDIP results were reported without any accompanying measure of generalization; Section~\ref{sec:axis4} treats the two separately.

\paragraph{Sensitive data and de-identification.}
Further work on datasets concerns harmful content, including undesirable social bias, explicit material, and personally identifiable information \citep{birhane-large-image-datasets-2021}.
Exposure of personal data in public corpora increases the risk of fraud and identity theft \citep{sweeney-2006}, models have been shown to leak their training data \citep{carlini21extracting}, and community norms reflect the concern: the NeurIPS Code of Ethics states that ``datasets should minimize the exposure of any personally identifiable information, unless informed consent from those individuals is provided to do so''.\footnote{\url{https://neurips.cc/public/EthicsGuidelines}, accessed October 2026.}
De-identification is well developed in adjacent fields, particularly face obfuscation in computer vision \citep[e.g.,][]{yang-imagenet-face-obfuscation-2022} and clinical natural language processing \citep[e.g.,][]{vakili-dalianis-2022-utility}, where anonymization is often a legal precondition of use, but much less developed for document images.
The principal exception applies black redaction boxes to RVL-CDIP documents \citep{pagel-redactions-2024} and finds that doing so degrades classifier performance, which motivates the utility-preserving approach of Section~\ref{sec:deid}.

\section{Sensitive Personal Data and De-Identification}\label{sec:deid}

RVL-CDIP contains a substantial quantity of sensitive personal data: Social Security numbers, dates and places of birth, home addresses, and more.
The rest of this audit produces data configurations of RVL-CDIP, such as the corrected labels and the de-duplicated test split, that we release so that our results can be reproduced.
We cannot responsibly release them with this personal data in place, and we therefore de-identify the corpus before the audit begins.

This section quantifies the exposure of sensitive personal data in RVL-CDIP and describes a de-identification method that replaces each sensitive entity with a synthetic but plausible one and leaves classifier predictions unchanged; the conventional method, which covers the entity with a redaction rectangle, is used as a baseline.
We run the experiments from Section~\ref{sec:axis1} onwards on the synthetic-replacement corpus, including the OCR of Section~\ref{sec:axis5}, and we distribute this corpus as the default version of RVL-CDIP; Appendix~\ref{sec:artifacts} lists the links to the released data.

\subsection{Problems with Automated Detection}

Automated detection would be preferable to reading 400,000 documents, and we evaluated four personally identifiable information (PII) detectors against our annotations: Presidio, Google DLP, Microsoft Azure's language service, and Amazon Comprehend.
Two measurements show that they are unsuitable, one of coverage and one of recall.
Their coverage is narrow: of the eleven sensitive entity types we record, only Social Security number and age are supported by all four, seven are detected by none, and the address and telephone detectors that they do provide do not distinguish between a home address or number and an employer's (Table~\ref{tab:pii-entities}).
Their recall on the most important type is also uneven: measured as document-level recall over 1,773 annotated documents, Presidio finds 0.97 of Social Security numbers and Google DLP 0.93, but Amazon finds 0.77 and Azure 0.70, the latter missing roughly 30\% of the documents that contain one.
Appendix~\ref{app:pii} shows the per-tool results.
Large language models are a further option, and \citet{xu-etal-2026-sensitive} evaluate several of them for sensitive entity extraction, including on RVL-CDIP; they find performance uneven across entity types and below that of a pattern-based baseline on Social Security numbers.

We therefore annotated the documents by hand.
Manual annotation enables the position of each entity on the page to be recorded, which subsequent redaction requires and document-level detection does not provide.

\subsection{Annotation}

We extended the four entity types identified in our earlier work \citep{larson-etal-2023-evaluation} (Social Security numbers, dates of birth, places of birth, and marital status) to the fuller set in Table~\ref{tab:pii-entities}, taking as reference the sensitivity classifications used by large institutions and by the US Department of Homeland Security's guidance on personally identifiable information \citep{dhs-pii-2017}.

Person names are not among the types we record, because names occur throughout this corpus (on letterheads, in signature blocks, in distribution lists) and most of them are not sensitive, so that annotating them would mark a large fraction of the corpus without addressing the risk.
A name becomes a risk when it appears beside a Social Security number or a date of birth, so we annotate those entities and do not annotate the name.

\begin{wraptable}{r}{0.37\textwidth}
  \centering
  \footnotesize
  \caption{The eleven sensitive entity types recorded during annotation.}
  \label{tab:pii-entities}
  \begin{tabular}{l}
    \toprule
    \textbf{Entity type} \\
    \midrule
    US Social Security number \\
    Age \\
    Date of birth \\
    Sex/gender \\
    Place of birth \\
    Marital/parental/spousal status \\
    Home address \\
    Home telephone number \\
    Citizenship/nationality \\
    Religious affiliation \\
    Health status \\
    \bottomrule
  \end{tabular}
\end{wraptable}

We distinguish \emph{home} addresses and \emph{home} telephone numbers, which we treat as sensitive, from business addresses and telephone numbers, which we do not.
The general categories would match every letterhead and business card in the corpus, which are not sensitive and were mostly published deliberately.
Entities that are sensitive but are outside the enumerated types, such as a photograph on a scanned driver's license, were recorded as \emph{other}.

The annotation was carried out as part of our earlier study \citep{larson-etal-2024-de} by a team of nine people, five of them co-authors of that paper; the lead author, an author of the present paper, served as expert annotator and organized the process.
FUNSD and Tobacco3482 were inspected by at least two annotators including the expert; DocVQA and Tobacco800 by the expert; and approximately half of RVL-CDIP by two annotators, the remainder by the expert alone.
Inter-annotator agreement on the binary judgment of whether a document contains sensitive data, measured over five annotators on Tobacco3482 and FUNSD, yielded a Fleiss's kappa of 0.918.
We did not use crowdsourcing, because in addition to its cost it would have exposed this material to a large number of additional people, which is contrary to the purpose of the annotation.

\subsection{Prevalence of Sensitive Personal Data}

\begin{wraptable}{R}{0.38\textwidth}
  \centering
  \footnotesize
  \setlength{\tabcolsep}{3pt}
  \caption{Sensitive entity counts in RVL-CDIP. Counts for all five
           IIT-CDIP-derived datasets are given in
           Table~\ref{tab:pii-counts-full}.}
  \label{tab:pii-counts}
  \begin{tabular}{lr}
    \toprule
    Dataset size            & 400,000 \\
    \midrule
    US Social Security no.  & 2,342  \\
    Date of birth           & 12,800 \\
    Place of birth          & 6,125  \\
    Age                     & 219    \\
    Home address            & 3,908  \\
    Home telephone          & 1,801  \\
    Marital/parental status & 4,228  \\
    Citizenship/nationality & 2,647  \\
    Sex/gender              & 602    \\
    Health status           & 60     \\
    Religious affiliation   & 43     \\
    \midrule
    Documents affected      & 15,956 \\
    \bottomrule
  \end{tabular}
\end{wraptable}

Table~\ref{tab:pii-counts} shows the counts for each entity type in RVL-CDIP, and Table~\ref{tab:pii-counts-full} in Appendix~\ref{app:pii-counts} shows them for all five datasets.
We identified 15,956 RVL-CDIP documents containing sensitive personal data, 4.0\% of the corpus, among them 2,342 US Social Security numbers.
We applied the same annotation and de-identification to four further datasets sampled from IIT-CDIP (DocVQA, Tobacco3482, Tobacco800 and FUNSD), since a Social Security number on a source page recurs in each dataset built from that page, and Social Security numbers occur in all five datasets, most numerously in RVL-CDIP (Appendix~\ref{app:pii-counts}).
The most frequent entity types overall are dates and places of birth, home addresses, citizenship and nationality statuses, and marital, parental, and spousal statuses; the rarest are health status and religious affiliation, which appear only in RVL-CDIP and DocVQA.

\begin{figure}[tbp]
  \centering
  \includegraphics[width=0.85\linewidth]{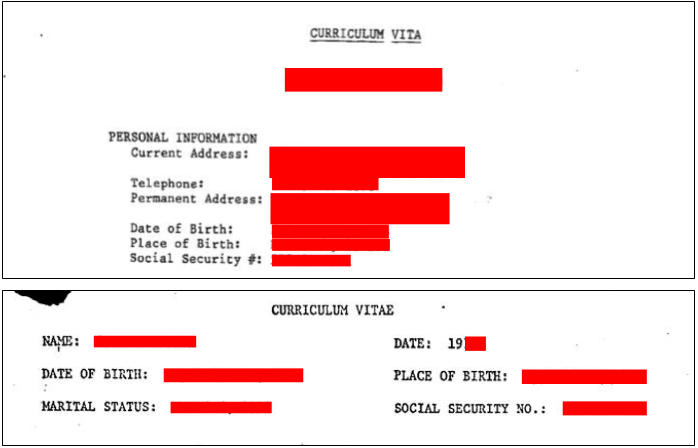}
  \caption{Documents from RVL-CDIP containing sensitive personal information.
           Sensitive fields have been obscured by us; they appear unobscured in
           the corpus as distributed.}
  \label{fig:pii-examples}
\end{figure}

The affected documents are ordinary ones: resumes, invoices, employment forms, and, in a number of cases, scanned photographs of driver's licenses and Social Security cards.
Figure~\ref{fig:pii-examples} shows representative examples, with the sensitive fields obscured by us.
Within RVL-CDIP the concentration is highest in \cat{resume}, where our earlier sampling found that 7.7\% of test documents contain a Social Security number, 43.6\% a date of birth, 19.9\% a place of birth, 11.4\% a marital or family status, and 8.9\% a citizenship status.
Because these types co-occur, the risk is greater than any single entity implies: a document containing a name together with a date and place of birth, a home address, and a Social Security number provides most of the information that identity fraud requires \citep{sweeney-2006}.

The period in which these documents were created partly explains why so many of them contain Social Security numbers and identity documents: Social Security numbers were used routinely as employee identifiers at that time, so their appearance on personnel paperwork was unremarkable, and individuals also gave identity documents to employers for ordinary administrative reasons.
Neither the employees nor their employers could have foreseen that these documents would become permanently and publicly available as part of a machine learning corpus.

We therefore do not reproduce any document containing sensitive data in this paper, or identify any person, beyond the examples in Figure~\ref{fig:pii-examples}, whose sensitive fields we have obscured.

\subsection{De-Identification with Synthetic Replacement}

We de-identify sensitive documents by synthetic replacement: each annotated entity is erased from the page and a generated value of the same type is written in its place.
Because the sensitive data annotation process recorded bounding boxes, the position of every sensitive entity on the page is known, which makes replacement possible.
The annotated regions are small, occupying less than 2\% of the page in most affected documents, and they are few relative to the corpus: 15,956 of RVL-CDIP's 400,000 pages contain an annotation, and those pages contain 39,435 regions in total, so 96\% of the benchmark is left unaltered.
We implemented this replacement pipeline as a command-line tool, which we release together with the fonts and word lists it uses (Appendix~\ref{sec:artifacts}).

\begin{figure}[tbp]
  \centering
  \includegraphics[width=\linewidth]{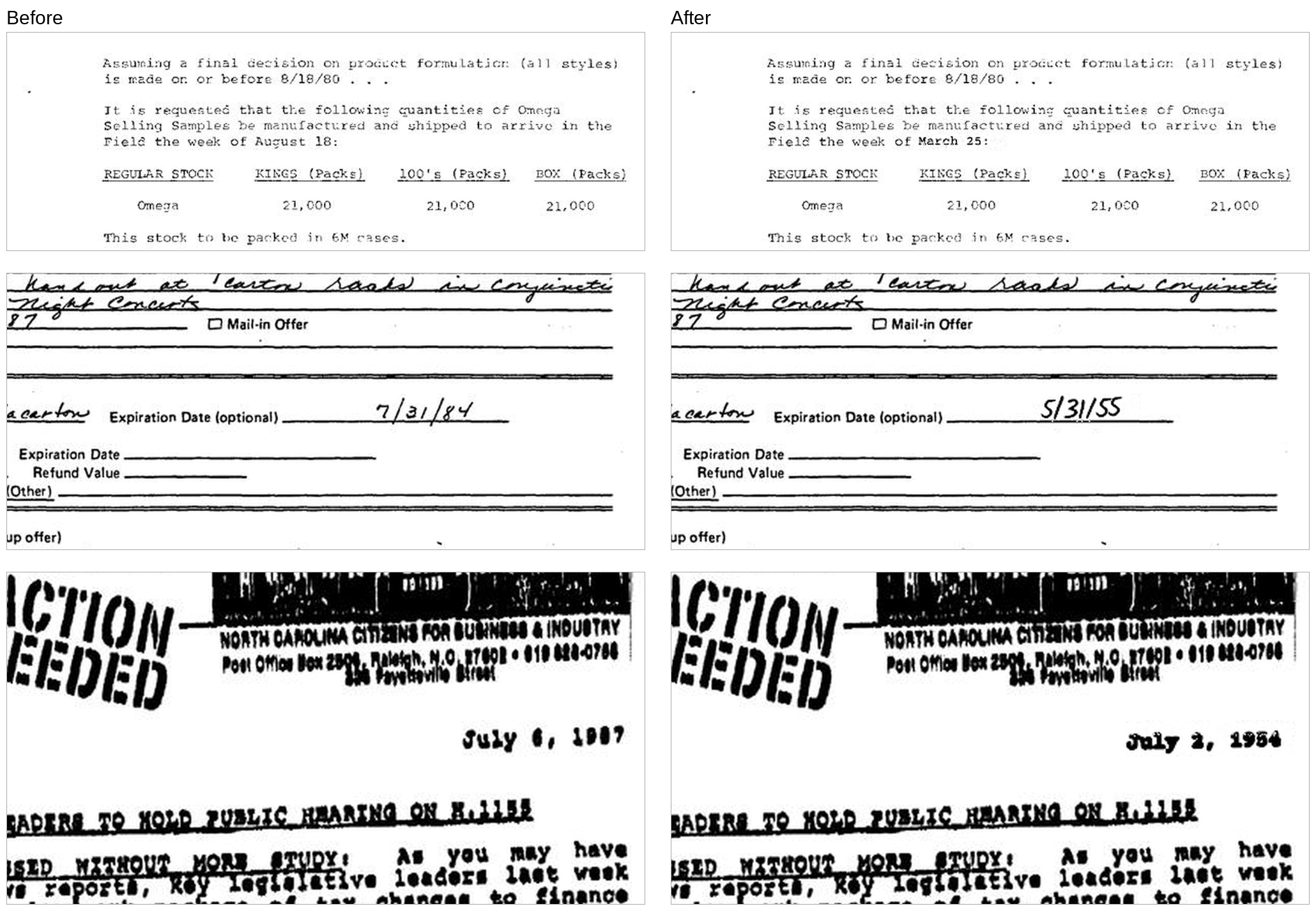}
  \caption{Synthetic replacement of non-PII dates for illustrative purposes, top to bottom:
           \texttt{August 18}~$\rightarrow$~\texttt{March 25},
           \texttt{7/31/84}~$\rightarrow$~\texttt{5/31/55},
           \texttt{July 6, 1987}~$\rightarrow$~\texttt{July 2, 1954}.}
  \label{fig:redaction-pairs}
\end{figure}

Replacement values have two sources: places, nationalities, religions and months are taken from gazetteer lists included with the tool, and the patterned types are generated directly, so that each value has the internal structure its type requires.
For example, a Social Security number has the format that the Social Security Administration issues, a telephone number conforms to the North American numbering plan, and a ZIP code is taken from the range belonging to its state.

The conventional alternative is to cover each annotated region with a black or white redaction rectangle.
We generate a replacement because these corpora are also used for tasks other than classification: a rectangle placed over the answer to a visual question answering item makes that item unanswerable, and removes the target of an information extraction item, whereas a generated value of the same type preserves both while removing the original value.
Replacement also makes any entity our annotation missed harder to find, since a reader can no longer assume that a plausible Social Security number on one of these pages is a real one.

Figure~\ref{fig:redaction-pairs} shows examples of annotated regions before and after replacement, from which it can be seen that a replacement must match the typography of the page it is written into.
We measure the typeface, size, stroke weight, baseline and skew of the original text on the page itself: for each annotated region, the pipeline locates the surrounding words in the page's Amazon Textract OCR and then carries out the steps below.
Appendix~\ref{app:redact} describes the full implementation.

\begin{enumerate}
  \item \textbf{Local context.} The Textract words lying at least 70\% inside the annotated bounding box are grouped into text lines.
  The measurements in the remaining steps are taken from those words; where the box does not enclose a complete word, the nearest words on the same line are used instead.

  \item \textbf{Typeface.} The pipeline includes nineteen open-license fonts, chosen to cover the serif, sans-serif and typewriter faces that predominate in the corpus.
  Each candidate font is used to render the words that Textract read in the box, at the width of the scanned words, and the scan and each rendering are then despeckled, thinned to one-pixel skeletons, and compared by chamfer distance \citep{barrow-1977-chamfer}.
  The comparison is repeated over as many as three words from the box, and the font with the best mean rank across them is selected.
  Regions that Textract labels \texttt{HANDWRITING} are matched against handwriting fonts instead, and the face chosen for the page's longest handwritten words is reused for every handwritten region on that page, so that all handwritten replacements in a document are drawn in the same handwriting font.

  \item \textbf{Size.} Size is estimated by re-rendering the scanned words in the selected font and scaling until their widths agree.
  Widths are used rather than heights because Textract pads its boxes vertically, by roughly 8\% of box height above the ink on these datasets.

  \item \textbf{Stroke weight.} Measured as ink area divided by half the boundary length, on the scanned words and on the same words re-rendered in the selected font. 
  The replacement is drawn with whichever outline width brings the two measurements closest; fonts already heavier than the scan are not thinned.

  \item \textbf{Baseline and skew.} Textract's polygons understate skew (a median of 0.05$^\circ$ against 0.44$^\circ$ measured from the pixels of Tobacco3482), so both are measured from the image.
  A word's baseline is the lowest row in which ink is dense, computed after long horizontal runs are removed so that an underline or a printed form line is not mistaken for the baseline.
  A line's slope is fitted by least squares through the baselines of its words where at least three words span at least 150 pixels, and is accepted if it is at most 5$^\circ$.
  Otherwise the page's skew is used, which is the median of the slopes fitted to the page's lines of at least three words spanning at least 300 pixels, and is therefore also measured from the pixels.

  \item \textbf{Erase and refill.} The region is filled with paper sampled from around it rather than with white, so that its color, grain and specks continue across the erased area.
  Ruled lines crossing the region, such as the printed lines of a form or an underline, are detected and redrawn: candidate rows are found by thresholding just below the paper's noise level, at three standard deviations, bridging the one-pixel steps of slightly tilted lines, and keeping only runs longer than the region is tall.
  Each candidate is then checked for clean paper above and below it and for ink in the gaps between words, which rejects a row of letter bottoms that resembles a ruled line.

  \item \textbf{Draw.} The text is rendered at four times the output resolution and downsampled.
  Supersampling allows the outline width of step 4 to change stroke thickness in steps of half a pixel, which is finer than the output grid.

  \item \textbf{Degrade.} IIT-CDIP documents are scans of physical paper and have the corresponding print and scanning noise, so a cleanly rendered replacement would be conspicuous.
  We add noise to the rendered text with Augraphy's \texttt{LowInkRandomLines} \citep{augraphy} and then darken the strongest strokes, so that the low-ink lines do not wash out the text.
\end{enumerate}

We apply two constraints to the generated values.
Formats are copied and values are not: \texttt{7/31/84} is replaced by another five-digit slash-delimited date, \texttt{MARRIED} by an upper-case status, and a date annotated without its year, such as \texttt{August 18} in Figure~\ref{fig:redaction-pairs}, is replaced by another month and day without a year.
Values with internal structure are also generated consistently, so that a ZIP code agrees with the state printed beside it, a telephone number obeys the North American numbering plan, and street and city names are real US names; a replaced address that could not exist would be as conspicuous as an undegraded one.

Fifteen of the entity categories receive replacement text, covering 96\% of annotated regions.
The remainder (sex and gender, signature, health status, and the residual \emph{other}) are erased and refilled but not written back, because these do not appear in standard formats.
A health status might read ``good health'' in one document and give height, weight and eye color in another, and none of our data generators covers that range; a language model could plausibly provide replacements here, but we did not pursue that direction.
Those regions, 3.8\% of the annotated total, are therefore blanked, and the claims above about replacement preserving answerability do not apply to them.

\subsection{Effect of De-Identification on Downstream Models}

Whether de-identification damages the data determines what we can responsibly release, so we measure its effect in two ways, against black and white box redaction as baselines.

We first compute document similarity between de-identified documents and their originals.
We take 445 \cat{resume} documents from RVL-CDIP, each with an original, black, white and synthetic-replacement version, embed all four with CLIP using the ViT-B/32 backbone \citep{clip-2021}, and compute cosine similarity and Euclidean distance between each redacted version and its original.
Black redaction produces the documents least similar to their originals.
White and synthetic replacement are both closer and behave alike, with synthetic replacement slightly closer: it is more similar than white for 61.6\% of documents by cosine similarity, less similar for 28.3\%, and tied for the remaining 10.1\%, and for no document was the black version the closest of the three.
Similarity decreases as the redacted area of a page increases.

We then measure the effect on a classifier.
Using DiT-base fine-tuned on the RVL-CDIP training split \citep{li-2022-dit-image-transformer}, we compare predictions and softmax confidence scores on the original and redacted versions of the same 445 documents.
Across all three treatments, \emph{no document changed its predicted label}.
Mean absolute change in confidence was 0.0036 for black, 0.0022 for white, and 0.0024 for synthetic replacement.
Eleven documents showed larger shifts; in the most extreme, the confidence for a document with roughly 3\% of its area redacted decreased by about 0.30, and the median shift among these eleven was 0.045.
Prior work applying black redactions to RVL-CDIP reported degraded classifier performance \citep{pagel-redactions-2024}; we attribute the difference to redacting only the sensitive regions we had localized, rather than larger or less targeted areas.

The conclusion we draw concerns release policy: de-identification of these corpora has almost no effect on classification, so that there is no research argument for preferring the unredacted original data, and the de-identified versions should be the default distribution.
We adopt that policy for the artifacts described in Appendix~\ref{sec:artifacts}.

We also adopt it for this paper, as stated at the start of this section, and give the text-based and multi-modal models of Section~\ref{sec:axis5} the OCR of the replaced pages.
A black or white box removes the words beneath it, so an OCR comparison on box-redacted pages would measure the effect of the redaction together with that of the engine.
A replaced page keeps words of the same types in the same places, with different values, and we therefore run the OCR comparison on replaced pages.
Appendix~\ref{app:redact-encoding} describes how the released pages are encoded and how the pipeline's use of Textract relates to that comparison.

\section{Label Guidelines for RVL-CDIP Categories}\label{sec:guidelines}

In this section we describe the criteria that we use to decide which of the 16 RVL-CDIP categories a document belongs to.
The original release of RVL-CDIP did not state what the categories contain;
Harley et al. report that the categories were selected in part because each had at least 25,000 representatives in IIT-CDIP, but did not publish definitions, criteria, or annotation rules for them.
It is also not documented who assigned the original category labels in the source collection; the available evidence suggests they were applied by staff at the library that hosted the tobacco documents \citep{d-lib-document-library, digital-archives-2022}.
Because the error analysis in Section~\ref{sec:axis1} and the annotation of out-of-distribution data in Section~\ref{sec:axis4} require a definition of each category, we wrote the label criteria in Table~\ref{tab:label-guidelines}, and we hope that others who audit or extend the corpus will use them as well.
We developed the label criteria in two stages: an initial review of 16,000 training documents, and revisions made during the exhaustive review of all 400,000 labels described in Section~\ref{sec:axis1}.

\subsection{Method}

We derived an initial set of criteria by sampling 1,000 documents from each of the 16 categories in the training split, and reviewing the resulting 16,000 documents category by category.
The purpose of this first pass was descriptive: we sought to characterize what each category contains, before judging whether individual documents belong to it.

The first pass showed that most categories are internally heterogeneous and comprise several distinct sub-types.
The \cat{resume} category, for instance, contains at least four sub-types: conventional resumes and curricula vitae, ``Biographical Sketch'' documents of the kind required for United States National Institutes of Health grant applications, third-person executive biographies, and scanned business cards.
The \cat{advertisement} category likewise includes print advertisements from newspapers and magazines, scripts for television and radio spots, and advertisement insertion orders.
A single category label thus describes a document less precisely than its name suggests, which we discuss further in Section~\ref{sec:discussion} because it affects zero-shot prompting and clustering analyses, both of which assume that each label describes a coherent set of documents.

\begin{figure}[tbp]
  \centering
  \includegraphics[width=\linewidth]{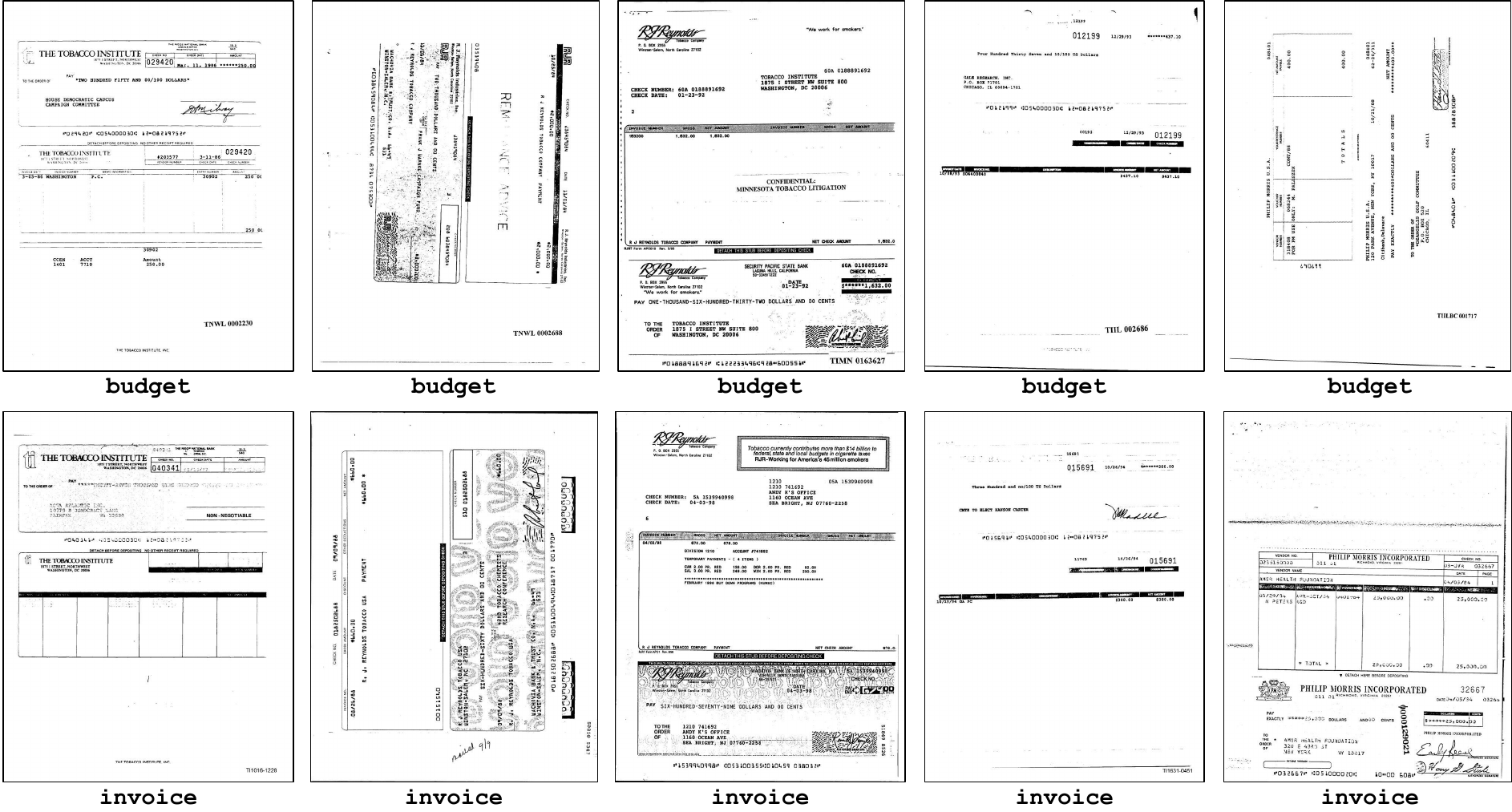}
  \caption{Check images labeled \cat{budget} (top row) and \cat{invoice}
           (bottom row) in RVL-CDIP.}
  \label{fig:checks}
\end{figure}

The first pass also showed that the boundary between some pairs of categories has to be decided for both categories together.
For instance, scanned checks (Figure~\ref{fig:checks}) and political contribution requests appear in both \cat{budget} and \cat{invoice}, so a rule for either category requires a decision about which of the two categories these documents belong to.
We resolved such cases empirically, by annotating the sub-types wherever they occurred and comparing their relative frequencies: check images accounted for 8.8\% of the \cat{budget} documents we reviewed but only 3.8\% of \cat{invoice} documents, and our criteria accordingly assign checks to \cat{budget}.
We set several rules in Table~\ref{tab:label-guidelines} in this way.

\begin{wrapfigure}{R}{0.25\textwidth}
  \centering
  \includegraphics[width=\linewidth]{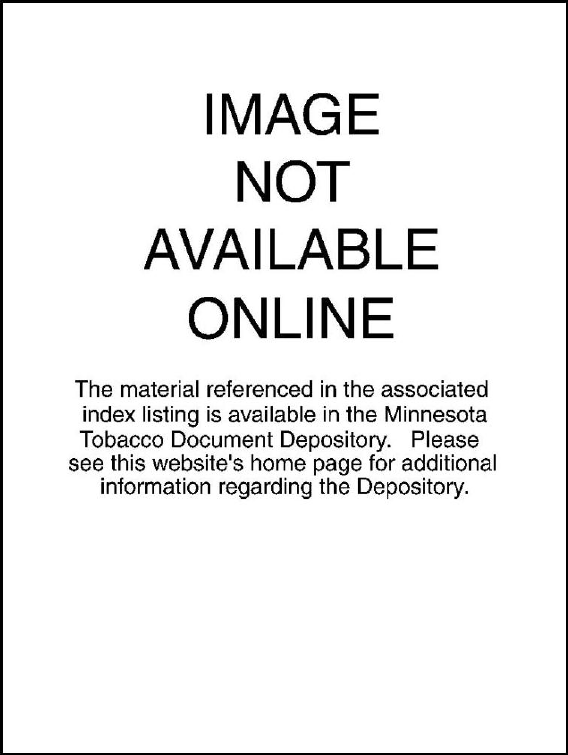}
  \caption{A placeholder page from the \cat{advertisement} category.}
  \label{fig:image-not-available}
\end{wrapfigure}

We also excluded from the criteria certain documents that recur in the corpus but do not have a semantic relation to the category in which they occur.
The clearest instance is a placeholder page reading ``IMAGE NOT AVAILABLE ONLINE'' (Figure~\ref{fig:image-not-available}), which states that the material is held in the Minnesota Tobacco Document Depository.
The Textract text of 494 RVL-CDIP pages contains the phrase ``image not available'', and 375 of these pages are labeled \cat{advertisement}.
Each of the 45 classifiers that we trained on the original labels predicts \cat{advertisement} for all 54 such pages in the test split, including the 11 pages labeled with other categories; the review of Section~\ref{sec:axis1} marked all 54 pages as unknown, and we remove them in the fixed conditions of Section~\ref{sec:joint}.

We revised the label criteria periodically during the roughly two years of exhaustive annotation, whenever annotators encountered boundary cases that the label criteria did not resolve.
Table~\ref{tab:label-guidelines} shows the final version of the label criteria.
The label criteria describe what each category contains in the corpus, and we did not try to reconstruct what the categories were originally intended to contain.

\subsection{Scope Conventions}

In addition to the label criteria for each category in Table~\ref{tab:label-guidelines}, we follow three conventions that apply to all 16 categories.

\emph{Single label.} RVL-CDIP is a single-label benchmark, and where a document could plausibly belong to two categories the corpus is largely consistent about which one is assigned: a handwritten letter, for example, is labeled \cat{handwritten} and not \cat{letter}.
We adopt the corpus's own resolutions where they are consistent, and record documents as \emph{mixed} where they are not (Section~\ref{sec:axis1}).

\emph{First page.} Most images are the first page of a document.
The systematic exceptions are charts and tables, typically labeled \cat{scientific\_report}, and presentation slides, labeled \cat{presentation}, both of which may be taken from any position within a document.

\begin{figure}[tbp]
  \centering
  \includegraphics[width=\linewidth]{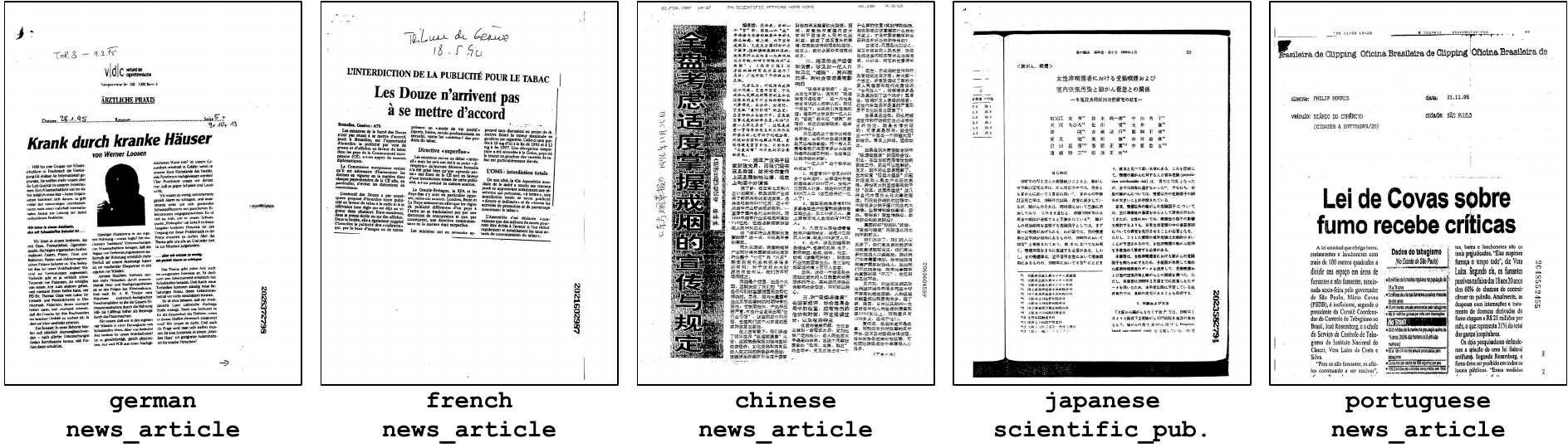}
  \caption{Non-English examples in RVL-CDIP, with their language and
           category label.}
  \label{fig:non-english}
\end{figure}

\emph{Language.} We assign a document to a category by its type, whatever its language.
Almost all documents containing legible text are in English, and our review also found documents in German, Dutch, French, Spanish, Portuguese, Italian, Japanese, Chinese, Arabic, and Hebrew, such as German, French, and Portuguese news clippings labeled \cat{news\_article} (examples are displayed inFigure~\ref{fig:non-english}).

\begin{table}[tbp]
  \centering
  \caption{Operational criteria for the 16 RVL-CDIP categories, with notes
           recording boundary cases encountered during annotation.}
  \label{tab:label-guidelines}
  \scalebox{0.72}{
  \begin{tabular}{l m{0.58\linewidth} m{0.41\linewidth}}
    \toprule
    \textbf{Category} & \textbf{Description} & \textbf{Notes} \\
    \midrule
    \cat{advertisement} & Advertisements from print media such as newspapers and magazines; a smaller number of scripts for television or radio advertisements; a small number of advertisement order and insertion documents. & Excludes the recurring ``IMAGE NOT AVAILABLE'' page, which appears predominantly in this category. \\
    \midrule
    \cat{budget} & Expense, spending, sales, cash, and accounting reports and forecasts; budgets; quotes and estimates; income and bank statements. Also receipt-like documents including political campaign contribution requests, other receipts, and checks and check stubs. & Checks are assigned here rather than to \cat{invoice}: 8.8\% of reviewed \cat{budget} documents were checks, against 3.8\% of \cat{invoice}. \\
    \midrule
    \cat{email} & Scanned images of printed email messages. & \\
    \midrule
    \cat{file\_folder} & Scanned images of folders and binders, often characterized by vertically oriented text indicating a folder label. & A moderate number contain handwritten notes. Some scans are indistinguishable from blank pages. \\
    \midrule
    \cat{form} & Documents with ruled lines or boxes indicating fillable fields; also documents composed largely of key--value pair elements, such as ``Size: Large'', whether or not the fields are ruled. & Substantial overlap with \cat{questionnaire}, many of which contain form-like elements. \\
    \midrule
    \cat{handwritten} & Handwritten documents, including handwritten letters and scientific notes. & Takes precedence over the content-based categories: a handwritten letter is \cat{handwritten}. \\
    \midrule
    \cat{invoice} & Invoices, bills, and account statements. & Excludes checks; see \cat{budget}. \\
    \midrule
    \cat{letter} & Letters, typically bearing letterhead and commonly opening with a ``Dear\ldots'' salutation. & The boundary with \cat{memo} is frequently unclear and accounts for much of this category's error rate. \\
    \midrule
    \cat{memo} & Memoranda and inter-office correspondence, often with explicit ``TO'', ``FROM'', and ``SUBJECT'' headings. & \\
    \midrule
    \cat{news\_article} & News articles as clippings from newspapers and other print news media, and a small number from the web. Also press releases. & Press releases occur in both this category and \cat{presentation}, more often in the latter. \\
    \midrule
    \cat{presentation} & Presentation and overhead slides; transcripts of speeches and statements. Also a large number of press releases. & See \cat{news\_article}. \\
    \midrule
    \cat{questionnaire} & Customer surveys and questionnaires; survey prompts written for surveyors; questionnaires forming part of legal proceedings and investigations. & Many contain substantial form-like elements; see \cat{form}. \\
    \midrule
    \cat{resume} & Resumes and curricula vitae; ``Biographical Sketch'' documents; third-person executive biographies; a small number of business cards. & Contains a high concentration of sensitive personal data; see Section~\ref{sec:deid}. \\
    \midrule
    \cat{scientific\_publication} & Papers and articles from scientific journals and book chapters; book title pages; articles from science newsletters. & Newsletter articles closely resemble \cat{news\_article}. \\
    \midrule
    \cat{scientific\_report} & Bioassay, pathology, and test reports; charts, graphs, and tables; research reports including progress reports; research proposals, abstracts, and paper drafts. & Reports and abstracts resemble \cat{scientific\_publication}; test results resemble \cat{specification}. \\
    \midrule
    \cat{specification} & Data sheets, including safety data sheets; product, material, and test specifications; specification change reports. & \\
    \bottomrule
  \end{tabular}}
\end{table}

\section{Axis I: Label Validity}\label{sec:axis1}

Using the label criteria of Section~\ref{sec:guidelines}, we determine how many of RVL-CDIP's 400,000 labels are correct.

\subsection{Automated Error Detection}

One way to estimate how many labels are wrong without reviewing every document is automated label-error detection.
Confident learning \citep{northcutt2021confidentlearning}, as implemented in CleanLab, is one such method; it identifies likely errors from the disagreement between a model's predicted class probabilities and the assigned label, and it has been applied successfully to benchmarks across several modalities \citep{pervasive-label-errors-2021}.
Other methods rank documents by the area under the margin during training \citep{mislabeled-pleiss-2020} or by the loss of a fine-tuned pre-trained language model \citep{chong-etal-2022-detecting}.
However, all three methods assume that label errors are rare, because each of them judges a label by the predictions of a model trained on the labels under audit, and these predictions are informative only if most of those labels are correct.
In RVL-CDIP the errors are frequent and systematic, and because they arise from category boundaries that were never specified, they recur in the same direction across all three splits.
A model trained under these conditions thus learns the erroneous convention, and its confidence scores are high for the erroneous labels, as we show for confident learning in Appendix~\ref{app:cleanlab}.
Establishing the error rate therefore requires human review.

\subsection{Annotation Protocol}

We manually inspected all 400,000 documents in RVL-CDIP.
Annotation was carried out by a team of seven volunteer undergraduate researchers over approximately two years, working from the criteria in Table~\ref{tab:label-guidelines}.
Documents were distributed in batches of 1,000 taken from a single category, accompanied by that category's criteria.
For each document, annotators recorded whether it satisfied the criteria for its assigned category: ``yes'' where it clearly did, ``no'' where it clearly did not, and ``unsure'' in ambiguous cases.
Every ``no'' and every ``unsure'' was reviewed by the project lead, along with a sample of ``yes'' judgments, both to correct individual annotations and to identify where the criteria themselves needed revision.
Documents marked ``no'' were assigned a corrected label wherever one could be determined; these corrections are released with the paper and are the basis of the remediated corpus of Section~\ref{sec:joint}.

\begin{figure}[tbp]
  \centering
  \includegraphics[width=\linewidth]{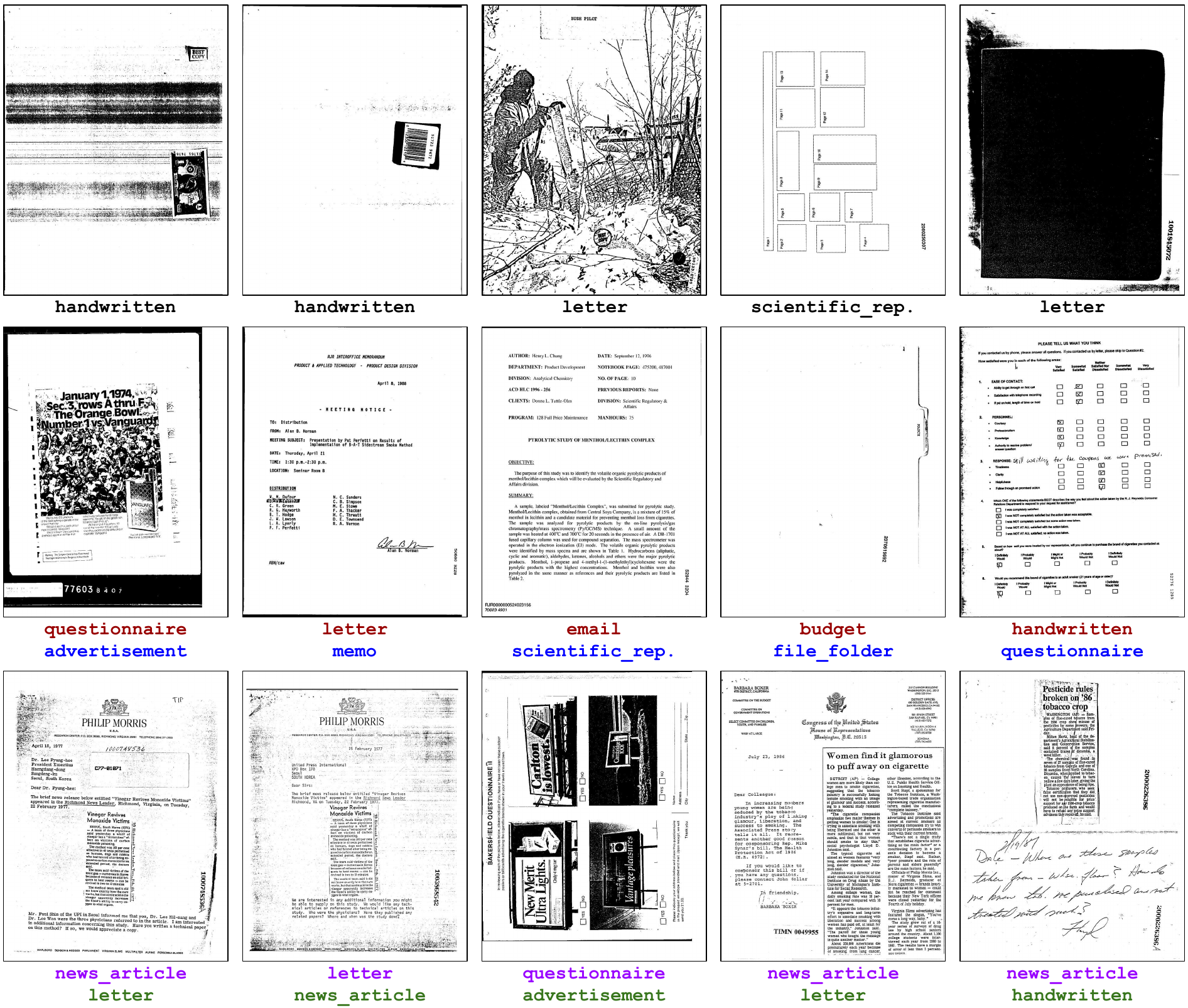}
  \caption{Examples of each error type. Top row: \emph{unknown}, documents that
           do not satisfy the criteria for any RVL-CDIP category. Middle row:
           \emph{wrong}, where a different category is correct. Bottom row:
           \emph{mixed}, where more than one category legitimately applies.}
  \label{fig:error-examples}
\end{figure}

We distinguish three kinds of label error, and display examples of each label error type in Figure~\ref{fig:error-examples}.
A document is \emph{unknown} where it does not satisfy the criteria for any RVL-CDIP category, so that no correct label for it exists within the label set.
It is \emph{wrong} where its assigned label is incorrect but a different, correct label is available.
It is \emph{mixed} where it satisfies the criteria for two or more categories, including its assigned one, which the single-label format of the benchmark cannot represent.
A document is \emph{unsure} where an annotator marked it as ambiguous and the review did not settle whether its label is correct, and we report these documents separately from both the error count and the correct count.

Inter-rater agreement was assessed on a sample of 1,500 documents from five categories (\cat{resume}, \cat{letter}, \cat{memo}, \cat{handwritten}, and \cat{email}), which gave a Fleiss's kappa of 0.74, in the range conventionally described as substantial agreement \citep{landis-koch-1977}.

\subsection{Findings}

\begin{table}[tbp]
  \centering
  \caption{Label error rates by category and error type. \emph{Any Error} is the
           sum of \emph{unknown}, \emph{wrong}, and \emph{mixed};
           the final column additionally counts \emph{unsure} cases as errors.}
  \label{tab:label-errors}
  \scalebox{0.85}{
  \begin{tabular}{lcccccc}
    \toprule
    \textbf{Category} & \textbf{Unknown} & \textbf{Wrong} & \textbf{Mixed} & \textbf{Any Error} & \textbf{Unsure} & \textbf{Any\,+\,Unsure} \\
    \midrule
    \cat{advertisement}           & 5.89\% & 0.60\% & 1.62\% & 8.12\% & 1.50\% & 9.61\% \\
    \cat{budget}                  & 4.93\% & 8.48\% & 3.94\% & 17.35\% & 1.66\% & 19.00\% \\
    \cat{email}                   & 8.12\% & 1.27\% & 0.12\% & 9.51\% & 0.50\% & 10.02\% \\
    \cat{file\_folder}            & 3.33\% & 0.16\% & 0.52\% & 4.01\% & 0.08\% & 4.09\% \\
    \cat{form}                    & 4.54\% & 3.60\% & 6.10\% & 14.25\% & 1.04\% & 15.29\% \\
    \cat{handwritten}             & 9.50\% & 2.12\% & 1.03\% & 12.66\% & 0.00\% & 12.66\% \\
    \cat{invoice}                 & 5.83\% & 8.52\% & 0.60\% & 14.94\% & 1.16\% & 16.10\% \\
    \cat{letter}                  & 15.32\% & 15.51\% & 0.42\% & 31.26\% & 0.04\% & 31.30\% \\
    \cat{memo}                    & 3.11\% & 1.72\% & 1.12\% & 5.96\% & 1.91\% & 7.87\% \\
    \cat{news\_article}           & 3.51\% & 4.50\% & 0.54\% & 8.55\% & 0.96\% & 9.52\% \\
    \cat{presentation}            & 3.26\% & 1.34\% & 0.87\% & 5.47\% & 2.08\% & 7.55\% \\
    \cat{questionnaire}           & 7.38\% & 5.98\% & 1.09\% & 14.44\% & 1.27\% & 15.72\% \\
    \cat{resume}                  & 1.47\% & 0.01\% & 0.00\% & 1.48\% & 0.01\% & 1.49\% \\
    \cat{scientific\_publication} & 2.15\% & 2.39\% & 0.00\% & 4.54\% & 2.72\% & 7.26\% \\
    \cat{scientific\_report}      & 2.90\% & 4.09\% & 5.17\% & 12.16\% & 0.45\% & 12.60\% \\
    \cat{specification}           & 1.55\% & 1.13\% & 0.74\% & 3.42\% & 0.24\% & 3.66\% \\
    \midrule
    RVL-CDIP                        & 5.17\% & 3.84\% & 1.49\% & \textbf{10.51\%} & 0.98\% & 11.48\% \\
    \bottomrule
  \end{tabular}}
\end{table}

We find that approximately \errorRate\ of RVL-CDIP is erroneously labeled (Table~\ref{tab:label-errors}): 5.17\% \emph{unknown}, 3.84\% \emph{wrong}, and 1.49\% \emph{mixed}, or 11.48\% if \emph{unsure} cases are counted as errors.

The rate differs greatly between categories: \cat{letter} is erroneously labeled 31.26\% of the time, close to a third of the category, with the errors concentrated at its boundary with \cat{memo}, a distinction that our own criteria describe as frequently unclear.
\cat{budget} is the only other category above 15\% (17.35\%), followed by \cat{invoice} (14.94\%), \cat{questionnaire} (14.44\%), \cat{form} (14.25\%), \cat{handwritten} (12.66\%), and \cat{scientific\_report} (12.16\%).
At the other extreme, \cat{resume} is erroneously labeled only 1.48\% of the time, and \cat{specification}, \cat{file\_folder}, and \cat{scientific\_publication} are all below 5\%.
The \emph{wrong} labels also have a direction: of the 3,878 \cat{letter} documents that we relabeled, 3,253 (83.9\%) are memos, of the 2,129 relabeled \cat{invoice} documents 1,846 (86.7\%) are budgets, and of the 597 relabeled \cat{scientific\_publication} documents 548 (91.8\%) are news articles; Table~\ref{tab:label-confusion} in Appendix~\ref{app:label-confusion} shows the original and corrected labels of all 15,366 relabeled documents.
As a consequence, a model's aggregate accuracy on RVL-CDIP is an average over categories whose labels differ in reliability by more than a factor of twenty, and per-category results reported against this benchmark are not comparable with one another.

The errors are also systematic within categories, and the exhaustive review showed several patterns that a sample would not have shown.
Press releases, for example, occur in both \cat{news\_article} and \cat{presentation}, and are the more common of the two in \cat{presentation}, which became apparent only after enough of both categories had been reviewed to perceive the imbalance.

\subsection{Sampled and Exhaustive Error Rates}\label{sec:axis1-sampling}

Our own earlier work \citep{larson-etal-2023-evaluation} estimated RVL-CDIP's label error rate at 8.1\%, with a further 1.7\% of documents having multiple valid labels, from a sample of 1,000 training documents from each of the 16 categories.
On the training split, the exhaustive review finds 8.97\% without and 10.45\% with the \emph{mixed} documents, so that the sampled estimates were 0.9 and 0.7 percentage points lower, a difference larger than the sampling error of the earlier estimate (about 0.4 points at 95\% confidence).
We see two possible causes: annotators who review a whole category see more documents near its boundaries than annotators who review a sample, and some errors, such as the press-release asymmetry described above, are only apparent across many documents.
Because the difference is small, we conclude only that a sampled estimate can understate the error rate.

\section{Axis II: Split Integrity}\label{sec:axis2}

The test split of a benchmark is informative to the extent that its documents were not available in training, because where evaluation data recurs in training, reported accuracy measures the recall of specific instances as well as generalization to the task, in proportions that are unknown.
\citet{allamanis-2019-adverse-duplication} describes the resulting scores as ``inflated'', and similar findings have been reported for question answering \citep{lewis-etal-2021-question}, image retrieval \citep{radenovic-2018-revisiting-oxford-paris}, and document information extraction \citep{laatiri-2023-funsd-sroie-overlap}.
This section measures how much of RVL-CDIP's test split recurs in its training split.

\subsection{Definition of Duplicates}

We treat two documents as duplicates or near-duplicates in either of two cases.
In the first case the two samples are copies of the same underlying document that differ only in scanning artifacts, noise, or geometric distortion.
The second case is a \emph{template match}: the two documents are distinct instances of the same underlying form or letter, differing in the values filled into a shared structure.
A purchase order form completed for two different orders, or the same circular letter addressed to two recipients, are template matches.

\begin{figure}[tbp]
  \centering
  \includegraphics[width=\linewidth]{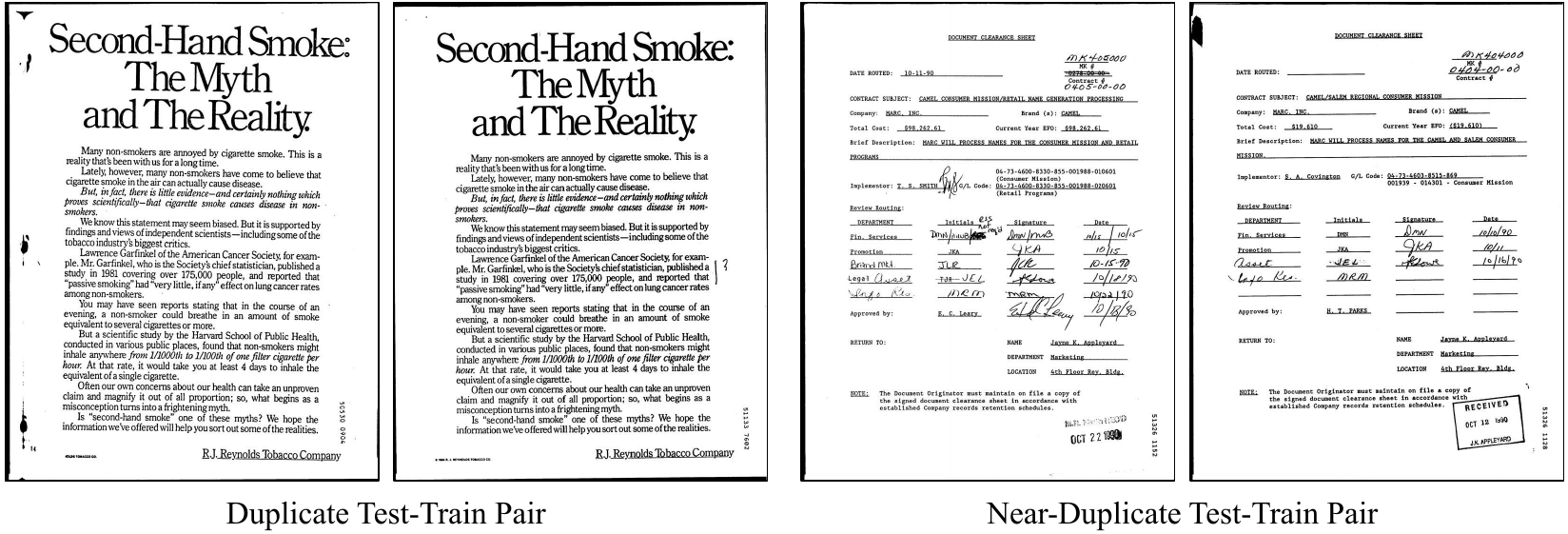}
  \caption{Example test--train pairs: the first is a duplicate, the second a
           template-match near-duplicate. Near-duplicates need not be
           pixel-identical: they share an underlying document template while
           differing in filled fields or scanning conditions.}
  \label{fig:duplicate-types}
\end{figure}

The second case requires justification, since the two documents are in an ordinary sense not the same document, and we count it as leakage for two reasons.
The first reason is the task: RVL-CDIP requires a document's \emph{type}, and document type is very largely a function of layout and structure, which is what the two documents of a template match have in common, whereas the variable content that distinguishes them is irrelevant to the task, so that a model that was trained on one instance has been shown the discriminative structure of the other.
The second reason is empirical and is developed in Section~\ref{sec:axis3}: models on this benchmark rely on structural and incidental page features and not on document semantics, so that structural repetition is the kind of repetition that increases their scores.
Figure~\ref{fig:duplicate-types} shows examples of both types.

\subsection{Limitations of Perceptual Hashing}

The standard tools for duplicate detection in image datasets use perceptual hashing (pHash and related methods, as used in CleanVision and fastdup), which assigns near-identical hashes to near-identical images and is unsuitable for both of the cases above.
In the first case, scanner noise and geometric distortion between two scans of the same page are frequently large enough in pixel terms to produce different hashes, so recall is low, and in the second case hashing fails by construction, because template matches differ in their filled content, to which a perceptual hash is designed to be sensitive.
A method that treats RVL-CDIP's duplicates as an image-hashing problem will therefore substantially undercount them.

\subsection{Filter-and-Refine Detection}

We therefore use a two-stage filter-and-refine pipeline: an inexpensive approximate stage proposes candidate pairs, and a more discriminative stage adjudicates them.

\paragraph{Filter.}
We embed every document with CLIP \citep{clip-2021}, giving a 512-dimensional representation per page, and $\ell_2$-normalize.
Stacking the train and test embeddings into matrices $X_{\mathrm{train}}$ and $X_{\mathrm{test}}$, the full matrix of cosine similarities is a single product $S = X_{\mathrm{train}}X_{\mathrm{test}}^{\top}$, which makes exhaustive comparison across the corpus tractable.
For each test document we retain its ten most similar training documents as candidates.
This stage is deliberately permissive: its purpose is recall, and its output is not used as evidence of duplication.

\begin{figure}[tbp]
  \centering
  \includegraphics[width=\linewidth]{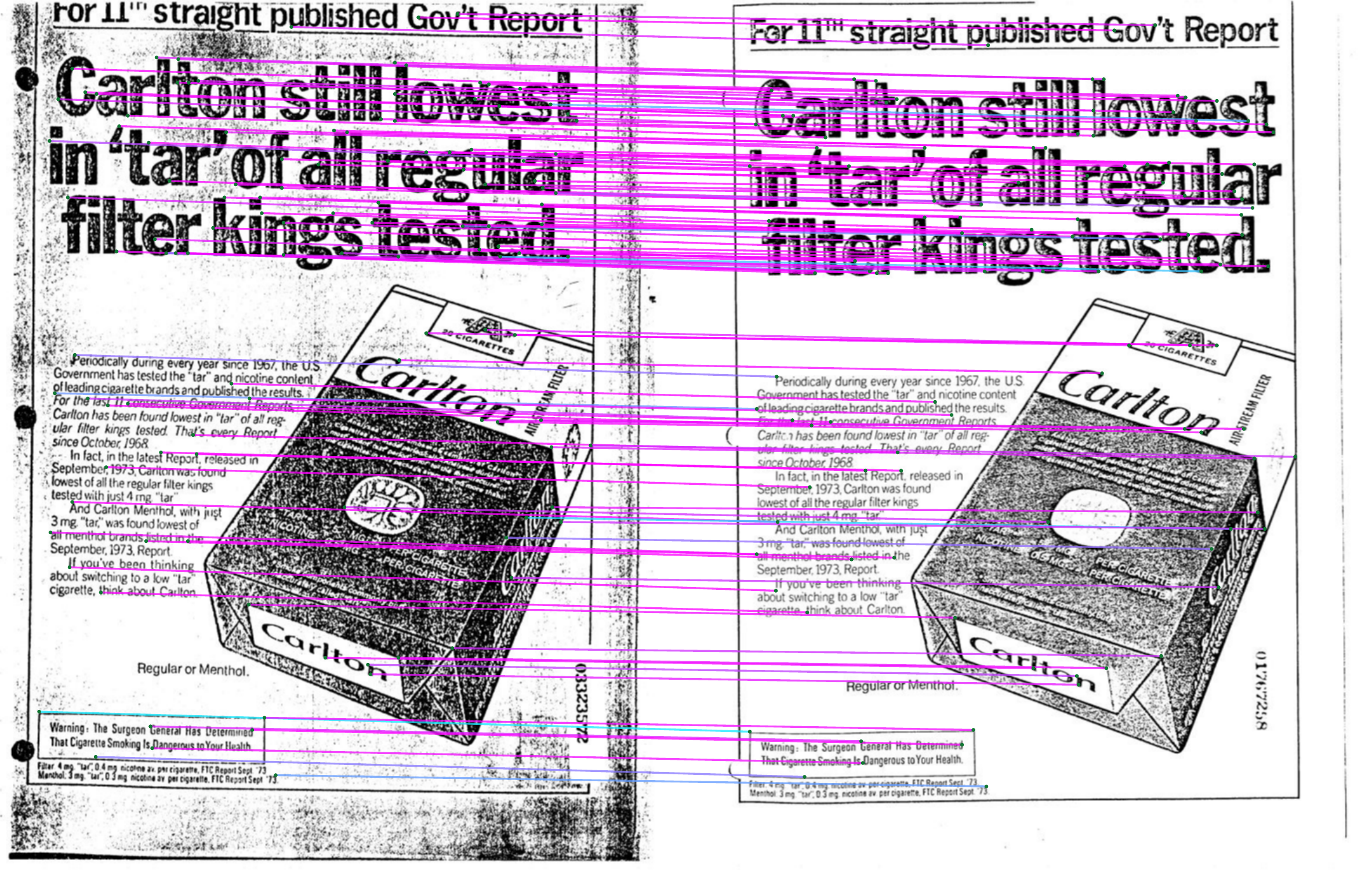}
  \caption{Matched SuperPoint keypoints of a test document (left) and its duplicate in the training split (right), a random sample of 120 of the 599 matches.}
  \label{fig:match-example}
\end{figure}

\paragraph{Refine.}
Each candidate (test, train) pair is then assessed by two complementary methods, one that compares the page images and one that compares their recognized text.
For the images we use local feature matching: SuperPoint \citep{superpoint-2018} detects keypoints and computes descriptors, and LightGlue \citep{lindenberger2023lightglue} matches them across the pair (Figure~\ref{fig:match-example}).
SuperPoint was designed for robustness to the geometric transformations that distinguish two scans of one page, and the printed structure that the two documents of a template match share also produces many matched keypoints.
For the text we use sequence similarity, computed with the Ratcliff/Obershelp algorithm \citep{difflib-gestalt} over the documents' recognized text.
Because this method gives a high score to long common subsequences, it detects the shared body of text of both duplicate types and tolerates the small regions in which the two documents differ.
A pair is judged a duplicate where its sequence similarity is at least $0.2$ and its feature match count is at least $100$.

\paragraph{Calibration.}
We tuned these thresholds against a reference set of candidate pairs that the authors judged by hand.
For 50 test documents from each of the 16 categories, we took two candidates from the filter stage, one chosen at random and the one with the highest CLIP similarity, which yields 1,528 distinct pairs.
We marked 56 of these pairs as unsure and evaluate on the other 1,472, of which 403 are true matches.
At the operating point above, the pipeline achieves precision of 100\%, recall of 82.9\%, and $F_1$ of 90.6\%.
We calibrated for precision, so that every pair we report is a genuine duplicate, with the consequence that we miss roughly one pair in six and \emph{all duplication rates in this section are lower bounds}.
The reference pairs and our judgments are available on the project website (Appendix~\ref{sec:artifacts}), so that others can assess the threshold choice directly.

\begin{figure}[tbp]
  \centering
  \includegraphics[width=\linewidth]{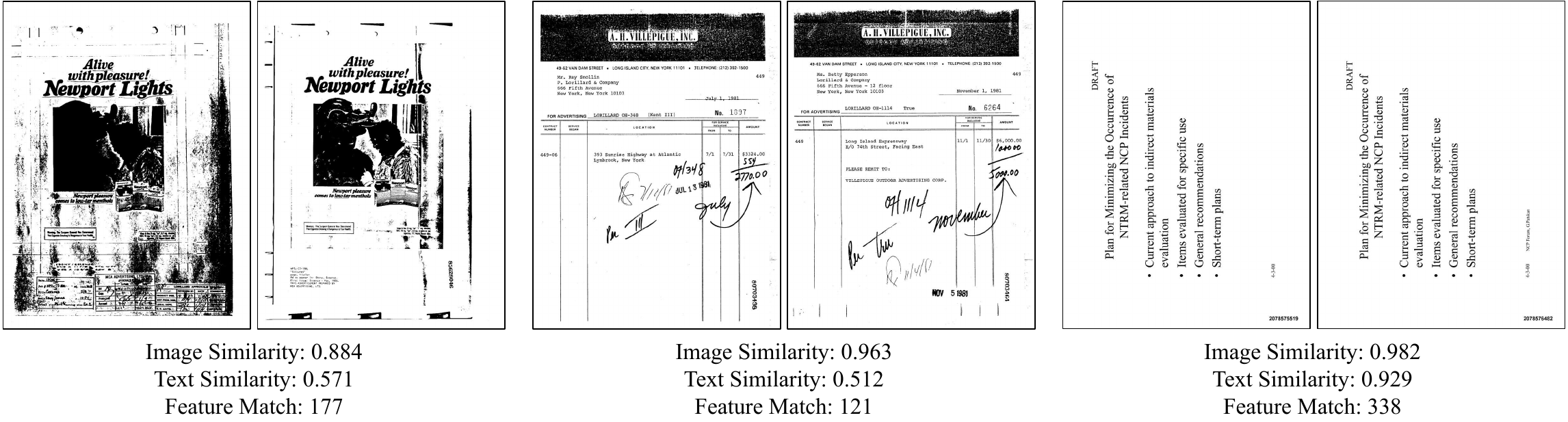}
  \caption{Further test--train pairs detected as duplicates, each annotated
           with its CLIP similarity (\emph{Image Similarity}), text sequence
           similarity (\emph{Text Similarity}) and SuperPoint match count
           (\emph{Feature Match}).}
  \label{fig:test-train-pairs}
\end{figure}

\subsection{Findings}

Table~\ref{tab:duplication} reports the results, and Figure~\ref{fig:test-train-pairs} shows further examples of the pairs the pipeline identifies.
Approximately \dupRate\ of RVL-CDIP's test split has a duplicate or near-duplicate counterpart in its training split.

\begin{wraptable}{R}{0.55\textwidth}
  \centering
  \caption{Test--train and validation--train duplication by category. The
           percentage is the proportion of that split having a duplicate or
           near-duplicate counterpart in the training split.}
  \label{tab:duplication}
  \scriptsize
  \setlength{\tabcolsep}{2pt}
  \begin{tabular}{lcccc}
    \toprule
    \textbf{Category} & \multicolumn{2}{c}{\textbf{Test}} & \multicolumn{2}{c}{\textbf{Validation}} \\
    \cmidrule(lr){2-3}\cmidrule(lr){4-5}
    & \textbf{\# Dup.} & \textbf{\% Dup.} & \textbf{\# Dup.} & \textbf{\% Dup.} \\
    \midrule
    \cat{advertisement}           & 856   & 34.04\% & 843   & 33.43\% \\
    \cat{budget}                  & 1,479 & 59.04\% & 1,450 & 58.35\% \\
    \cat{email}                   & 175   & 6.96\%  & 190   & 7.51\%  \\
    \cat{file\_folder}            & 50    & 1.98\%  & 56    & 2.28\%  \\
    \cat{form}                    & 1,458 & 58.18\% & 1,455 & 57.35\% \\
    \cat{handwritten}             & 102   & 4.03\%  & 112   & 4.60\%  \\
    \cat{invoice}                 & 1,703 & 68.75\% & 1,803 & 69.99\% \\
    \cat{letter}                  & 292   & 11.85\% & 259   & 10.66\% \\
    \cat{memo}                    & 457   & 18.34\% & 440   & 17.37\% \\
    \cat{news\_article}           & 501   & 20.34\% & 514   & 20.35\% \\
    \cat{presentation}            & 710   & 28.53\% & 688   & 27.88\% \\
    \cat{questionnaire}           & 1,499 & 61.56\% & 1,515 & 60.19\% \\
    \cat{resume}                  & 1,461 & 57.59\% & 1,360 & 56.06\% \\
    \cat{scientific\_publication} & 445   & 17.30\% & 450   & 17.81\% \\
    \cat{scientific\_report}      & 856   & 34.27\% & 899   & 35.85\% \\
    \cat{specification}           & 2,032 & 82.20\% & 2,117 & 83.64\% \\
    \midrule
    Total                            & 14,076 & \textbf{35.19\%} & 14,151 & \textbf{35.38\%} \\
    \bottomrule
  \end{tabular}
\end{wraptable}

As with label errors, the rate varies widely between categories.
Six categories exceed 50\%: \cat{specification} (82.20\%), \cat{invoice} (68.75\%), \cat{questionnaire} (61.56\%), \cat{budget} (59.04\%), \cat{form} (58.18\%), and \cat{resume} (57.59\%); eight of the sixteen exceed one third, whereas \cat{file\_folder} (1.98\%), \cat{handwritten} (4.03\%), and \cat{email} (6.96\%) have almost no overlap.
The ordering has an explanation: the heavily affected categories consist of standardized documents (specifications, invoices, forms, questionnaires), in which the same template recurs across an organization's records, while the unaffected ones consist of documents that are structurally unique to their author, as handwriting necessarily is.
For \cat{specification}, more than four fifths of the test split has a counterpart in training, which should be considered when a per-category accuracy for it is read.

The duplication rate does not depend on the particular split: applying the same pipeline to the validation split yields an overlap with training of 35.38\%, against 35.19\% for test, a difference of less than two tenths of a point.
Had duplication arisen from an unlucky partition, the two splits would not agree so closely, and their agreement indicates that duplicates are distributed evenly throughout the corpus, and therefore that duplication is a property of how RVL-CDIP was assembled from IIT-CDIP and is independent of how it was subsequently divided.
Re-partitioning the existing corpus therefore cannot remove the duplication, and duplicate pairs must be identified and removed, which is what the remediated variants of RVL-CDIP in Section~\ref{sec:joint} do.

\subsection{Duplication and Label Error}\label{sec:axis2-vs-axis1}

Because Sections~\ref{sec:axis1} and~\ref{sec:axis2} report per-category rates over the same 16 categories, we can ask directly whether the two defects coincide, and across categories they do not: label error rate and duplication rate are almost uncorrelated (Pearson $r = -0.03$; Spearman $\rho = 0.14$).
The extreme cases occur in both directions: \cat{specification} combines the corpus's highest duplication (82.20\%) with one of its lowest error rates (3.42\%), while \cat{letter} combines the highest error rate (31.26\%) with low duplication (11.85\%).

A plausible interpretation is that the two defects have different origins.
Duplication results from templating: categories composed of standardized documents contain many duplicates, and standardized documents are also easy to label correctly, since their type is evident from their structure.
Label error results from semantic proximity between categories, as in the \cat{letter}/\cat{memo} boundary, which is unrelated to whether documents recur.

If this interpretation is correct, the two confounds are independent and their effects on reported accuracy are cumulative.
We treat this as a hypothesis, because a correlation over 16 category-level aggregates is weak evidence and is a statement about categories and not about documents.
Section~\ref{sec:interaction} tests the question at the level of individual documents, where the relevant annotations permit a direct join.

\section{Axis III: Spurious Cues from Identification Codes}\label{sec:axis3}

The preceding two sections concern defects in RVL-CDIP's labels and in the division of its splits.
This section concerns a defect in the document images themselves: a visual feature that correlates strongly with the target label and is not related to the document type.
In this section we describe the identification codes stamped on the pages of RVL-CDIP, measure how well the codes alone predict the category of a document, and describe how we detect the codes and remove them from all 400,000 documents; Section~\ref{sec:joint} measures the effect of the codes on trained classifiers.

\subsection{Identification Codes}

\begin{wrapfigure}{R}{0.5\textwidth}
  \centering
  \includegraphics[width=\linewidth]{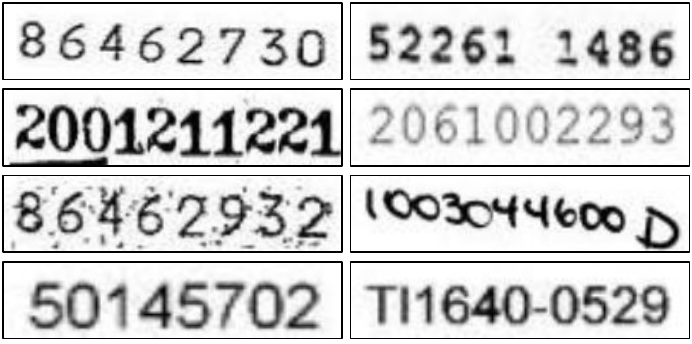}
  \caption{Identification codes from RVL-CDIP, illustrating the range of
           position, orientation, typeface, and format.}
  \label{fig:example-codes}
\end{wrapfigure}

Nearly every document in RVL-CDIP has a short alphanumeric code stamped onto the page (Figure~\ref{fig:example-codes}), which is a Bates number or a comparable identifier that was applied during legal discovery to index the document within the litigation.
The codes were added after the documents were created, by the parties that produced them, to track the documents through a legal process.

\begin{wrapfigure}{r}{0.5\textwidth}
  \centering
  \includegraphics[width=\linewidth]{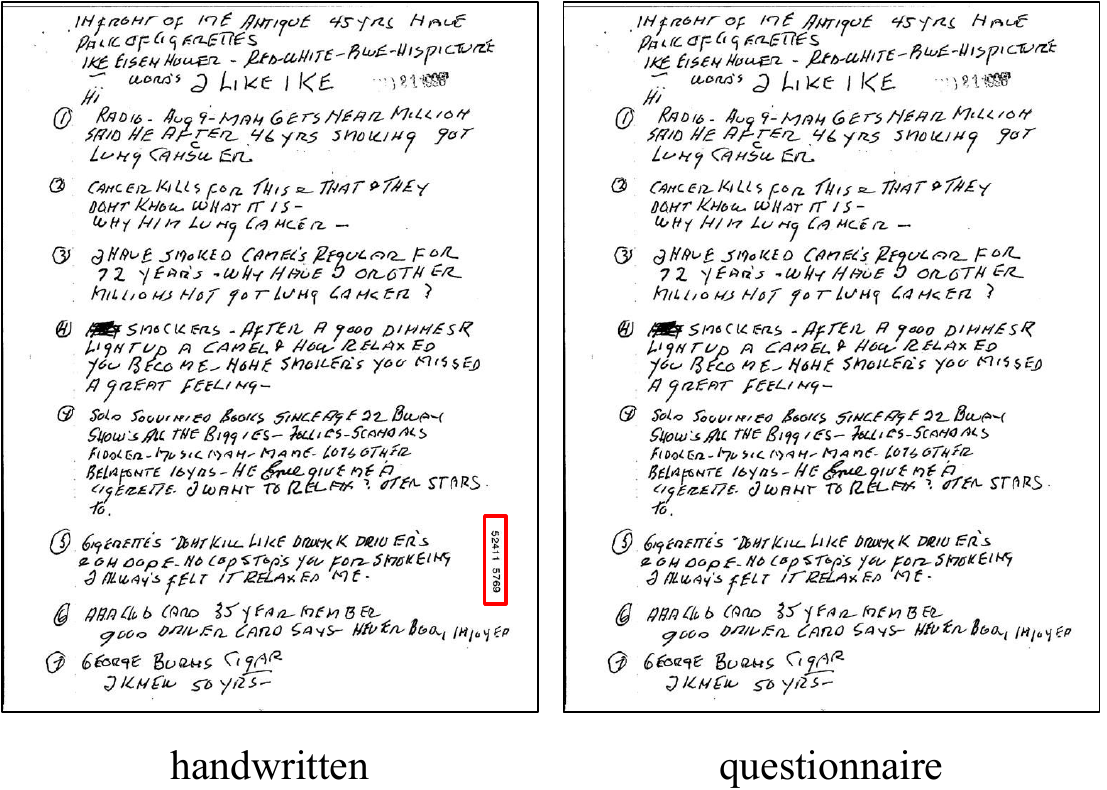}
  \caption{Document classifiers are sensitive to the removal of identification
           codes. After the code (marked in red) is erased from the page, a DiT
           classifier changes its prediction to an incorrect label.}
  \label{fig:idcode-teaser}
\end{wrapfigure}

The code of a document records where the document was placed in the numbering of the litigation, and the accuracy of a classifier that is given only the codes (Section~\ref{sec:axis3-signal}) shows that this position is correlated with the type of the document: documents with neighboring numbers tend to belong to the same category.
The release of RVL-CDIP does not record how the documents were collected and numbered, so we cannot establish the cause of this correlation; one possible cause is that documents of the same kind were kept in the same files and were numbered together when the files were produced.
A model that learns to use these codes therefore learns how the documents were numbered during the litigation, which is correlated with the label within RVL-CDIP and does not have any value for documents from other sources.
A classifier that relies on the codes can change its prediction when a code is erased from the page, as in the example of Figure~\ref{fig:idcode-teaser}.

Recent work applying explainability methods to document classifiers observed that these codes are highly relevant features \citep{saifullah-2024-explainability}, but did not pursue the observation.

\subsection{Annotation and Distribution}

We annotated identification codes in 11,200 documents sampled from RVL-CDIP and in all 3,482 documents of Tobacco3482, marking each code with a bounding box.
Regions were cropped, rotated to horizontal where necessary, and transcribed with Amazon Textract.

The codes are present on most documents: the RVL-CDIP sample contains 11,102 codes distributed over 10,306 documents, or 92.0\% of the sample; some documents have more than one code, and 894 have none.
Tobacco3482 is comparable at 92.8\%, so that the codes are available to a classifier on about nine of every ten inputs.

Both the position and the content of the codes vary systematically by category.
Positionally, the codes of each category have a distinct distribution over the page (Figure~\ref{fig:distributions}, and Figure~\ref{fig:distributions-all} in Appendix~\ref{app:idcodes} for all 16 categories): codes in \cat{email} documents cluster into roughly three regions, and the codes of \cat{handwritten} documents lie almost all in one narrow region at the right edge of the lower half of the page.
As to content, 10,099 of the 11,102 codes (91.0\%) begin with a digit, and the distribution over that first digit differs between categories (Table~\ref{tab:starting-digit-full} in Appendix~\ref{app:idcodes}).
The extreme case is \cat{handwritten}, in which approximately 99.9\% of codes begin with the digit \texttt{5}.
A single character of the code is thus nearly sufficient to identify one of the sixteen categories.

\begin{figure}[tbp]
  \centering
  \includegraphics[width=\linewidth]{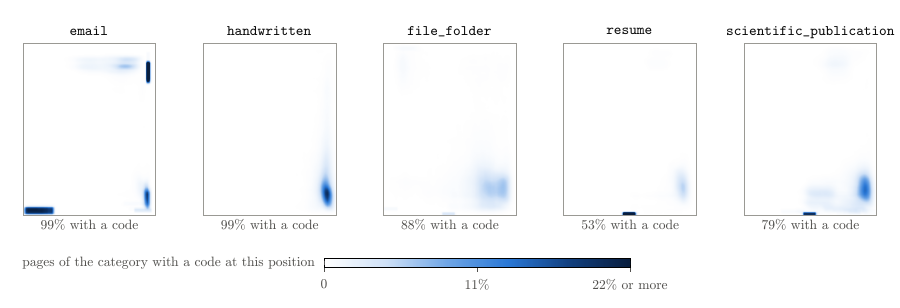}
  \caption{Position of the detected identification codes on the page for five
           RVL-CDIP categories, as the share of the category's pages with a code
           at each position; all 16 categories are in
           Figure~\ref{fig:distributions-all}.}
  \label{fig:distributions}
\end{figure}

\subsection{Predictive Signal of Identification Codes}\label{sec:axis3-signal}

To measure the information available in the codes alone, we train a classifier on features derived only from the code, without the rest of the document.
The features describe the position, dimensions, and orientation of its bounding box; the length and character composition of its text; the format of the text, by which we mean its pattern of digits and letters and its first one to five characters, each represented by the distribution over categories observed for it among the training documents; the categories of the training codes nearest to it in numeric value; the same description of a second code where the page has more than one; and the confidence of the OCR engine on the cropped code, together with whether OCR of the full page read the code and agreed with the reading of the crop (Appendix~\ref{app:idcode-features}).

A second configuration adds the class probabilities of a small convolutional network trained from scratch on the cropped code region, scaled to 32 by 256 pixels, so that the appearance of the code is available to the classifier as well as its transcription.
For RVL-CDIP the codes are located by the detector of Section~\ref{sec:axis3-removal} and transcribed by OCR of the cropped regions, and an XGBoost classifier is trained on the 320,000 training documents and evaluated on the 40,000 test documents; for Tobacco3482 we use the annotated boxes and their transcriptions and report the mean over ten stratified splits into 2,611 training and 871 evaluation documents.

\begin{wraptable}{R}{0.47\textwidth}
  \centering
  \caption{Accuracy of an XGBoost classifier on identification-code features,
           alone and with a CNN on the cropped code, and on the file name of
           the page. Zero-shot CLIP from \citet{siddiqui-zero-shot-2021}.}
  \label{tab:idcode-classification}
  \small
  \setlength{\tabcolsep}{3pt}
  \begin{tabular}{lcc}
    \toprule
    \textbf{Model} & \textbf{RVL-CDIP} & \textbf{Tob.3482} \\
    \midrule
    Random chance  & 6.25 & 17.8 \\
    Zero-shot CLIP & 39.2 & 49.5 \\
    \midrule
    Code feats.       & 69.2 & 56.5 \\
    Code feats.\ +CNN & 69.9 & 57.8 \\
    \midrule
    File name only      & 68.8 & 55.6 \\
    \bottomrule
  \end{tabular}
\end{wraptable}

On RVL-CDIP's 16-way task the classifier yields an accuracy of 69.2\% from the code features and 69.9\% with the convolutional network added, against a chance rate of approximately 6.25\%, and on Tobacco3482's 10-way task, where it has 2,611 training documents rather than 320,000, it yields accuracies of 56.5\% and 57.8\%, against a chance rate of 17.8\% (Table~\ref{tab:idcode-classification}).

For comparison, zero-shot CLIP, given the entire document image, achieves 39.2\% on RVL-CDIP and 49.5\% on Tobacco3482 in the evaluation of \citet{siddiqui-zero-shot-2021}, and 43.7\% on RVL-CDIP in our own evaluation of the CLIP ViT-L/14 model (Section~\ref{sec:joint}), so that a gradient-boosted model that is given only the features of the codes exceeds zero-shot CLIP by between 25.5 and 30.0 points on RVL-CDIP and by seven points on Tobacco3482.

A document's category can be predicted equally well---without the page image---from the identifier of the page in the dataset.
The original release of RVL-CDIP identifies each image by its file path, and its label files list the images by these paths, which are named after the codes: in a path such as \texttt{imagesq/q/o/c/qoc54c00/80035521.tif}, the folder is the document identifier of the source archive and the file name is the Bates number of the document, or the range of numbers for a multi-page document.
Applying the format, prefix, and nearest-number features to the file name alone, with no pixels, no detector, and no OCR, yields an accuracy of 68.8\% on RVL-CDIP, which is within half a point of the classifier that uses the codes on the page, and adding the folder identifier raises this to 70.3\%; the file names of Tobacco3482 follow the same convention and yield an accuracy of 55.6\% (last row of Table~\ref{tab:idcode-classification}).
The standard RVL-CDIP benchmark setup does not provide the file names to a model, so that a model can obtain this information only from the codes on the page, but the result shows that the information is in the numbering of the documents and does not depend on our detection and transcription, and that it can be reproduced from the paths of the public release (Appendix~\ref{app:idcode-analysis}).

\subsection{Detection and Removal at Scale}\label{sec:axis3-removal}

Establishing what the codes contribute to a trained classifier requires removing them from all 400,000 documents, for which we trained a YOLOv8 object detector on 8,000 annotated RVL-CDIP training documents.
On held-out data the detector achieves precision of 97.5\%, recall of 96.8\%, and $F_1$ of 97.1\% on RVL-CDIP, and precision of 97.8\%, recall of 93.7\%, and $F_1$ of 95.7\% on Tobacco3482, the latter without any Tobacco3482 training data, which indicates that the detector trained on RVL-CDIP is also accurate on Tobacco3482.
We applied it to the remaining 388,800 RVL-CDIP documents and replaced every detected region with white pixels, producing code-free versions of both corpora, which we release together with the detector weights and the detected regions (Appendix~\ref{sec:artifacts}).
The change in the accuracy of a classifier when the codes are removed is reported in Section~\ref{sec:joint}, with the other accuracy experiments, and Section~\ref{sec:interaction} examines how the removal interacts with label correction, de-duplication and the choice of OCR engine.

\section{Axis IV: Distributional Narrowness}\label{sec:axis4}

The preceding sections examine problems within RVL-CDIP; this section examines the range of documents the corpus does not cover and describes the two evaluation sets with which Section~\ref{sec:joint} measures how models trained on it behave outside that range.

\subsection{Distributional Coverage of RVL-CDIP}

RVL-CDIP is limited in three respects, all of which result from its sampling from IIT-CDIP.
In \emph{time}, every document predates 2006, and most were created decades earlier; in \emph{medium}, the documents are scans of physical paper with the noise of the scanning equipment of their era, and almost none are born-digital; and in \emph{domain}, the documents concern a single industry, and largely a single episode of litigation within it.

As we showed in Section~\ref{sec:background-leaderboard}, 23 of the 34 papers that we surveyed evaluate document classification only on RVL-CDIP, which consists of scanned tobacco-industry documents produced before 2006.
RVL-CDIP cannot show whether a model trained on it classifies a born-digital invoice from 2024, and because no alternative benchmark of comparable scale exists, this question has largely not been asked.

To examine it, we constructed evaluation data that differs from RVL-CDIP in time, medium and domain.

\subsection{Out-of-Distribution Evaluation: RVL-CDIP-\emph{N} and RVL-CDIP-\emph{O}}

Out-of-distribution behavior consists of two problems, both of which occur for a document classifier that is deployed on unconstrained input, and we constructed one evaluation set for each of them.
The first is \emph{generalization}: given a document that belongs to one of the sixteen categories but has a different source, does the model still classify it correctly?
This is covariate shift in the vocabulary of the domain-adaptation literature, in that the distribution over inputs changes while the labeling function relating input to label does not.
The second is \emph{abstention}: given a document belonging to none of the sixteen categories, is the model's confidence in its prediction low enough that the document can be identified and set aside?
In the terms of \citet{yang-2024-generalized-ood} this is semantic shift, since the input belongs to a class outside the label set, and it is distinct from concept shift, in which the labels of inputs from the same classes change.

\begin{figure}[tbp]
  \centering
  \includegraphics[width=\linewidth]{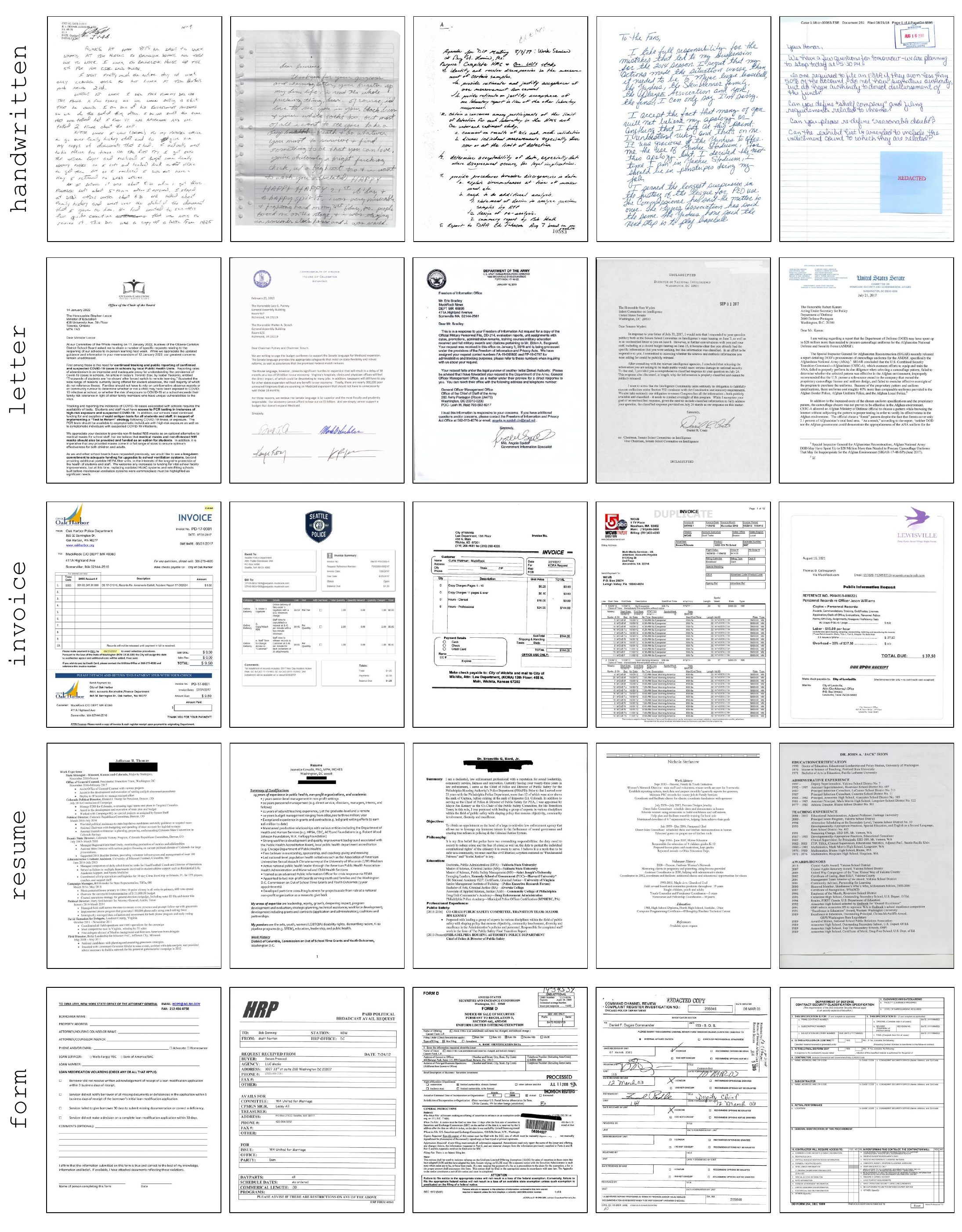}
  \caption{Documents from RVL-CDIP-\emph{N}, shown prior to grayscale
           conversion. These belong to the RVL-CDIP categories but are largely
           born-digital, post-2006, and unrelated to the tobacco industry.}
  \label{fig:rvlcdip-n-samples}
\end{figure}

\paragraph{RVL-CDIP-\emph{N}} contains \rvlcdipnSize\ documents that belong to the RVL-CDIP categories but are sampled from a different distribution (Figure~\ref{fig:rvlcdip-n-samples}).
Sources are the public DocumentCloud repository, which hosts government, legal, and public-service documents released under freedom-of-information requests, and general web search.
Roughly 65\% of the DocumentCloud documents and 95\% of the web-search documents are born-digital, whereas almost no RVL-CDIP document is, and the great majority postdate 2006.
Annotation of RVL-CDIP-\emph{N} followed the criteria of Section~\ref{sec:guidelines}, including RVL-CDIP's own single-label conventions, so that a handwritten letter is labeled \cat{handwritten} as it would be in the original corpus.
We admitted only unambiguous instances of each category.
The set covers twelve of the sixteen categories, because we did not find sufficient contemporary examples of \cat{advertisement}, \cat{file\_folder}, \cat{presentation} and \cat{scientific\_report}.

\begin{figure}[tbp]
  \centering
  \includegraphics[width=\linewidth]{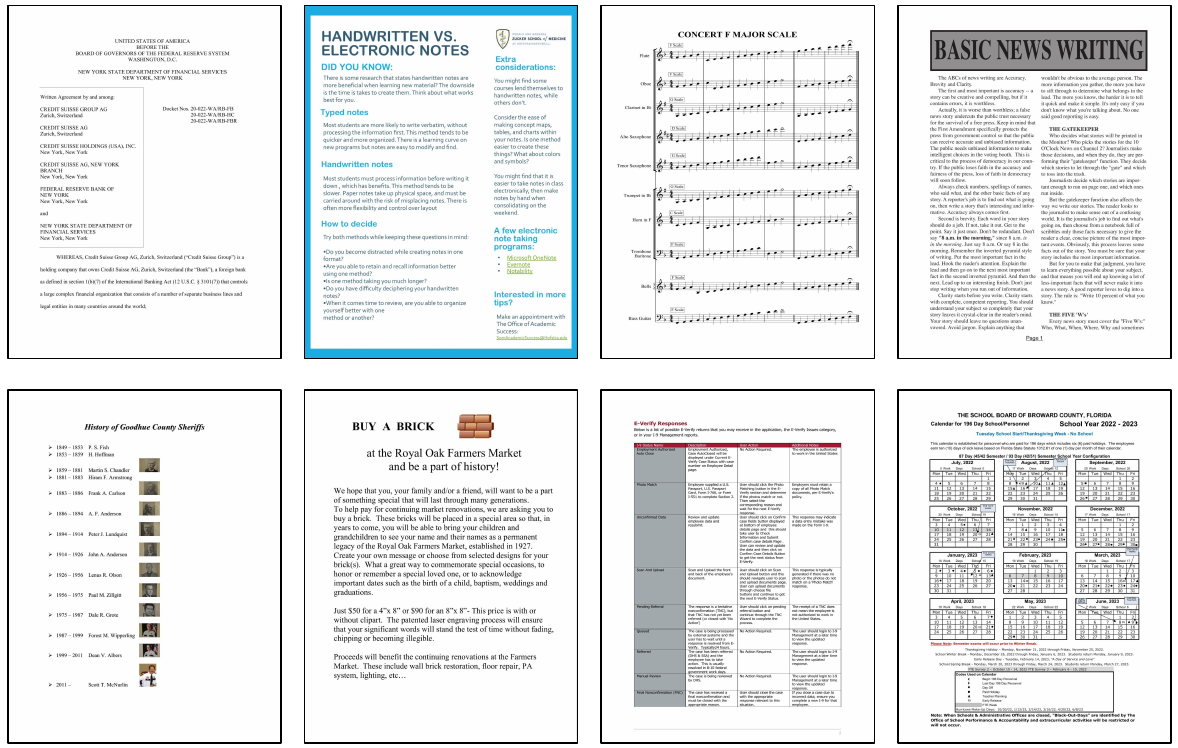}
  \caption{Documents from RVL-CDIP-\emph{O}, shown prior to grayscale
           conversion. Each is an unremarkable document that belongs to none of
           the 16 RVL-CDIP categories.}
  \label{fig:rvlcdip-o-samples}
\end{figure}

\paragraph{RVL-CDIP-\emph{O}} contains \rvlcdipoSize\ documents belonging to none of the sixteen categories: court filings, instructional guides, sheet music, timelines, schedules, and others (Figure~\ref{fig:rvlcdip-o-samples}).
These are sampled from the same two sources as RVL-CDIP-\emph{N}, DocumentCloud and general web search, and are ordinary documents of types that the label set of the benchmark does not cover.

Both sets follow RVL-CDIP's own conventions so that the comparison is fair: we take only the first page of each document, convert to grayscale, and scale so that the larger dimension is 1000 pixels.
Together they comprise \totalSize\ documents.

Since we released RVL-CDIP-\emph{N} and RVL-CDIP-\emph{O} \citep{larson-etal-2022-rvlcdipn}, three other studies have evaluated RVL-CDIP classifiers on out-of-distribution data.
\citet{he-etal-2023-dogood} alter RVL-CDIP pages synthetically, replacing words in the OCR text, merging word boxes, warping the page, or replacing its background with a natural photograph.
\citet{gu-etal-2023-critical} remove four RVL-CDIP categories (\cat{letter}, \cat{form}, \cat{scientific\_report} and \cat{presentation}) from training to serve as out-of-distribution data, and add scientific posters and the receipts of the CORD dataset \citep{park-etal-2019-cord}.
\citet{sheikh-etal-2025-zod} collected about 38,000 images from the web, roughly half of them in five RVL-CDIP categories and the rest in five categories that RVL-CDIP does not have, among them business cards, certificates and menus.
We do not evaluate our models on these sets.
Two of the document types that these studies treat as out of distribution are part of RVL-CDIP categories: \cat{budget} contains receipts and other receipt-like documents, and \cat{resume} contains business cards (Table~\ref{tab:label-guidelines}).
Because \citet{harley2015icdar-rvlcdip} did not publish definitions of the categories (Section~\ref{sec:guidelines}), the authors of these studies could not have known which document types each category includes, and part of the data that they use as out of distribution belongs to the label set of RVL-CDIP.

\subsection{Measuring Abstention}

To measure performance on RVL-CDIP-\emph{N}, we use accuracy, because the documents of RVL-CDIP-\emph{N} have correct labels; we report micro accuracy in Section~\ref{sec:joint-n}, alongside the remediation conditions, and macro accuracy in Appendix~\ref{app:joint}.

RVL-CDIP-\emph{O} requires different treatment, because the models were never trained to emit an ``other'' class.
We therefore assess whether a model's confidence separates in-domain from out-of-domain input.
For a document image $x$, let $f$ be a trained classifier and $z = f(x) \in \mathbb{R}^K$ its vector of logits over the $K=16$ categories, with $z_i$ the logit of category $i$; we consider two confidence functions of $z$.
The maximum softmax probability \citep{hendrycks-2017-baseline} is $\max_i s(z)_i$, the largest of the normalized logits
\[
  s(z)_i = \frac{e^{z_i}}{\sum_{j=1}^{K} e^{z_j}},
\]
and the energy score \citep{liu2020energy} is computed from the logits directly as
\[
  E(z; T) = T \cdot \log \sum_{j=1}^{K} e^{z_j / T},
\]
where $T$ is a temperature parameter, set to 1.
Higher values of either function indicate that the document is more likely to belong to one of the 16 categories.
Either can be thresholded to abstain, and we report the area under the ROC curve, which measures separability across all thresholds.
We report AUC in two settings: \emph{T-O}, separating the RVL-CDIP test split from RVL-CDIP-\emph{O}, and \emph{N-O}, separating RVL-CDIP-\emph{N} from RVL-CDIP-\emph{O}.
The second is the harder and the more realistic setting, because the model must distinguish documents it should classify from documents it should refuse when neither resembles its training data.
The results are reported in Section~\ref{sec:joint-abstention}; false positive rate at 95\% true positive rate, and two further confidence functions---MaxLogit \citep{hendrycks2022scaling} and GEN \citep{liu2023gen}---are reported in Appendix~\ref{app:ood}, but neither function changes any conclusion of that section.

\section{Axis V: OCR Engine Choice}\label{sec:axis5}

Unlike the four preceding axes, which concern properties of the corpus itself, Axis V concerns the protocol by which it is used: the corpus was released as a dataset of images, so every paper reporting a text or layout model has had to produce the tokens, and usually their bounding boxes, with an OCR engine of its own choosing, frequently without recording which.
``Accuracy on RVL-CDIP'' therefore does not denote a single quantity for these models, since two papers may report scores on the same 40,000 test images after giving their models materially different text, which is a failure of coordination between papers, for which no individual paper is at fault, and which has the largest consequences for the highest-ranked models, almost all of which have been multi-modal architectures since roughly 2020 (Table~\ref{tab:model-scores}).

\subsection{Evidence from Prior Work}\label{sec:axis5-prior}

Our earlier work gave two indications that the choice of engine has a substantial effect.
Both arose incidentally while we studied other questions, and neither is a controlled experiment, since both compare our own runs with published results.
In our out-of-distribution study \citep{larson-etal-2022-rvlcdipn}, LayoutLMv2 evaluated on the RVL-CDIP test split with Tesseract obtained 88.7\%, against the 95.3\% its authors reported with a proprietary Microsoft Azure OCR model; the architecture, weights and test images were the same, and the gap is 6.6 points.
In our label-error study \citep{larson-etal-2023-evaluation}, BERT and RoBERTa trained on AWS Textract output scored 2.63 and 2.65 points above their published results, although that comparison also involves different training runs.

\subsection{OCR Reporting in Prior Work}

Before we measure the effect of the OCR engine on the accuracy of a classifier, we ask how much the published papers report about the OCR engines that they used.
We extend the survey of the 34 RVL-CDIP papers of Section~\ref{sec:background} with three questions, which apply to the 24 of them that report a model using text or layout: whether the paper names the OCR engine that produced its RVL-CDIP text, whether it reports any measurement of OCR quality, and whether any of its experiments varies the engine (Table~\ref{tab:ocr-survey}).

\begin{table}[tbp]
  \centering
  \caption{OCR engine used for the RVL-CDIP text, as stated in each surveyed
           paper with a text or layout model. No paper reports a measurement of
           OCR quality on RVL-CDIP, and only UDoc compares engines.}
  \label{tab:ocr-survey}
  \footnotesize
  \setlength{\tabcolsep}{2pt}
  \begin{tabular}{@{}ll@{\hspace{1em}}ll@{}}
  \toprule
  \textbf{Paper} & \textbf{Engine} & \textbf{Paper} & \textbf{Engine} \\
  \midrule
  LayoutLLM (\citeauthor{fujitake-2024-layoutllm})          & EasyOCR & UDoc (\citeauthor{udoc-2021})                     & EasyOCR$^{a}$ \\
  EAML (\citeauthor{eaml-bakkali-2021})                     & Tesseract 4.0 & LayoutLM (\citeauthor{xu-2020-layoutlm})        & --$^{b}$ \\
  Bi-VLDoc (\citeauthor{bi-vldoc-2025-luo})                 & Tesseract & MATrIX (\citeauthor{matrix-2022})                     & Textract \\
  ERNIE-Layout (\citeauthor{ernie-layout-2022})             & Microsoft OCR & MGDoc (\citeauthor{mgdoc-2022-emnlp})             & EasyOCR \\
  UDOP (\citeauthor{tang-2023-udop})                        & -- & \citeauthor{pramanik-2022}                                 & Tesseract \\
  DocFormer (\citeauthor{Appalaraju_2021_ICCV-docformer})   & Textract & \citeauthor{dauphinee-2019}                          & Tesseract \\
  StructuralLM (\citeauthor{li-etal-2021-structurallm})     & Tesseract & SelfDoc (\citeauthor{selfdoc-2021})                   & Tesseract \\
  LayoutLMv3 (\citeauthor{layoutlmv3-huang-2022})           & Microsoft Read API & \citeauthor{Zingaro2021}                   & Tesseract 4.0$^{c}$ \\
  LiLT (\citeauthor{wang-etal-2022-lilt})                   & TextIn API & VLCDoC (\citeauthor{vlcdoc})                         & ``e.g.\ Tesseract'' \\
  LayoutLMv2 (\citeauthor{xu-etal-2021-layoutlmv2})         & Microsoft OCR & \citeauthor{audebert-2019}                      & Tesseract 4.0 \\
  TILT (\citeauthor{TILT-2021})                             & Microsoft Azure OCR & \citeauthor{pham-2022-long}               & -- \\
  Wukong-Reader (\citeauthor{wukong-reader-2022})           & in-house (Huawei) & Donut (\citeauthor{kim2022donut})             & none (OCR-free) \\
  \bottomrule
  \end{tabular}

  \smallskip
  \raggedright\scriptsize
  $^{a}$ Also reports 93.42, 93.96 and 94.10\% with Tesseract, EasyOCR and Google OCR.
  $^{b}$ Names Tesseract for the pre-training corpus only.
  $^{c}$ The text released by \citeauthor{audebert-2019}.
\end{table}

Of the 24 papers, 23 use an external OCR engine, Donut being the exception because it reads the page image directly.
Nineteen of the 23 name the engine that produced their RVL-CDIP text, VLCDoC names Tesseract only as an example of an off-the-shelf system, and three do not name it, among them LayoutLM, which names Tesseract for its pre-training corpus only.
The 19 named engines are Tesseract in eight papers, EasyOCR in three, Microsoft OCR services in four, Amazon Textract in two, the TextIn API in one and an in-house engine in one, so that eight of the 19 depend on a commercial or in-house system.
None of the 23 papers report a measurement of OCR quality on RVL-CDIP, and only UDoc \citep{udoc-2021} varies the engine, obtaining 93.42\%, 93.96\% and 94.10\% with Tesseract, EasyOCR and Google OCR, a range of 0.68 points that is larger than many of the differences between neighboring models in Table~\ref{tab:model-scores}.
Where the engine is not named, the reported number cannot be reproduced even given the model code and the benchmark, since a necessary input to the model is unspecified.

\subsection{Agreement Between OCR Engines}\label{sec:axis5-differ}

Downstream accuracy conflates OCR quality with a model's tolerance of OCR error, so before measuring the effect of the engines on a classifier we compare their output directly.
Since no ground-truth transcription exists for RVL-CDIP, the error rate of neither engine can be measured, and we therefore measure how far the Textract and Tesseract 4.1.1 text layers disagree, over all 400,000 pages.
For each page we count the words produced by each engine and compute two measures of disagreement, taking Textract as the reference only for normalization: the character edit distance divided by the length of the Textract text after whitespace is collapsed, and the $F_1$ overlap between the two multisets of words after lowercasing and stripping punctuation, which ignores reading order.

\begin{table}[tbp]
  \centering
  \caption{Agreement between the Textract and Tesseract 4.1.1 text layers on
           all RVL-CDIP pages (25,000 per category, 400,000 in all).
           \emph{Empty} pages do not have any text.
           Character distance is the edit distance normalized by the Textract
           length, and word overlap is the $F_1$ between the two bags of
           lowercased, punctuation-stripped words; both are medians over the
           374,439 pages on which neither engine is empty.}
  \label{tab:ocr-agreement}
  \scalebox{0.86}{
  \begin{tabular}{lrrrrrr}
    \toprule
    & \multicolumn{2}{c}{\textbf{Words per page}} & \multicolumn{2}{c}{\textbf{Empty (\%)}} & \textbf{Char.} & \textbf{Word} \\
    \cmidrule(lr){2-3}\cmidrule(lr){4-5}
    \textbf{Category} & \textbf{Textract} & \textbf{Tesseract} & \textbf{Textract} & \textbf{Tesseract} & \textbf{distance} & \textbf{overlap} \\
    \midrule
    \cat{advertisement}          &  97 &  51 &  3.1 & 11.9 & 0.76 & 0.23 \\
    \cat{budget}                 & 153 & 107 &  0.7 &  1.6 & 0.58 & 0.46 \\
    \cat{email}                  & 124 & 118 &  0.0 &  0.2 & 0.11 & 0.85 \\
    \cat{file\_folder}           &   4 &   6 & 54.3 & 22.6 & 0.86 & 0.17 \\
    \cat{form}                   & 164 & 139 &  0.3 &  0.8 & 0.35 & 0.61 \\
    \cat{handwritten}            &  95 &  61 &  1.6 & 10.7 & 0.73 & 0.09 \\
    \cat{invoice}                & 132 & 100 &  0.4 &  1.0 & 0.52 & 0.45 \\
    \cat{letter}                 & 205 & 191 &  0.2 &  0.6 & 0.10 & 0.87 \\
    \cat{memo}                   & 196 & 183 &  0.2 &  0.9 & 0.10 & 0.88 \\
    \cat{news\_article}          & 484 & 302 &  0.3 &  0.7 & 0.71 & 0.49 \\
    \cat{presentation}           & 145 & 139 &  1.4 &  0.9 & 0.16 & 0.84 \\
    \cat{questionnaire}          & 237 & 197 &  0.5 &  1.2 & 0.27 & 0.70 \\
    \cat{resume}                 & 327 & 314 &  0.1 &  0.1 & 0.08 & 0.89 \\
    \cat{scientific\_publication}& 628 & 524 &  0.3 &  0.5 & 0.64 & 0.71 \\
    \cat{scientific\_report}     & 174 & 155 &  0.5 &  1.0 & 0.17 & 0.80 \\
    \cat{specification}          & 219 & 183 &  0.1 &  0.4 & 0.38 & 0.68 \\
    \midrule
    All                             & 211 & 173 &  4.0 &  3.5 & 0.34 & 0.69 \\
    \bottomrule
  \end{tabular}}
\end{table}

Textract produces 84.6 million words across the corpus and Tesseract 69.3 million, 18\% fewer, and although the two engines produce no text on similar numbers of pages (15,979 and 13,833), only 4,253 pages are empty for both, so on thousands of pages one engine returns text and the other returns none.
Our comparison does not show which engine is right on these pages, since the text returned may be text that the other engine missed or spurious output from marks on a page without text.
File folders account for most of this, with Textract producing no text on 54\% of them and Tesseract on 23\%, while on handwritten pages and advertisements the pattern is the reverse and Tesseract produces no text on 11 to 12\% against 2 to 3\% for Textract.
Where both engines produce text, the disagreement depends strongly on the category (Table~\ref{tab:ocr-agreement}).
On letters, memos, email and resumes the engines differ on a median 8 to 11\% of characters and share 85 to 89\% of their words, whereas on news articles and scientific publications the character distance is 64 to 71\% while the word overlap is 49 to 71\%, so that part of the difference there is due to reading order.
On handwritten pages the two engines share only 9\% of their words and so produce what are effectively different transcriptions, and advertisements and file folders are nearly as divergent.

The magnitude of the effect is measured in Section~\ref{sec:joint-ocr}, and independently of it, a benchmark that is used for text-based models but does not provide text leaves a necessary input to the discretion of each user, which makes comparisons between users unsound to that extent.
We release both text layers over the full corpus, with bounding boxes and with the engine version recorded per document, so that later work can hold OCR fixed by using a published one or measure its own sensitivity by running against both.
The general form of this requirement is discussed in Section~\ref{sec:discussion}.

\section{Re-Benchmarking RVL-CDIP}\label{sec:joint}

In the preceding sections we examined the defects of RVL-CDIP one at a time.
In this section, we measure how the accuracy of \nModels\ models on the RVL-CDIP test split and on RVL-CDIP-\emph{N} changes when the defects are removed, separately and together, how well these models abstain on RVL-CDIP-\emph{O}, and how their accuracy depends on the OCR engine.

\subsection{Models}\label{sec:joint-models}

We evaluate twenty-five models in seven groups, which are the row groups of the tables in this section.
Seven image models are fine-tuned on the page image, of which AlexNet \citep{alexnet}, GoogLeNet \citep{googlenet}, SqueezeNet \citep{squeezenet}, VGG-16 \citep{vgg-16}, ResNet-50 \citep{res-net-50} and ResNeXt-50 \citep{resnext-Xie2016} are initialized with ImageNet weights and DiT-base \citep{li-2022-dit-image-transformer} with weights from self-supervised pretraining on IIT-CDIP \citep{iit-cdip-2006}, the collection from which RVL-CDIP was sampled.
Two text models, BERT-base \citep{devlin-etal-2019-bert} and RoBERTa-base \citep{roberta}, take the first 512 tokens of the OCR text of a page as input, and two multi-modal models take the OCR words together with their positions on the page, LayoutLM \citep{xu-2020-layoutlm} without the page image and LayoutLMv3 \citep{layoutlmv3-huang-2022} with it; both were also pretrained on IIT-CDIP, as is usual for the document models evaluated on RVL-CDIP.
Each of these eleven models is trained once for every combination of label set and identification-code condition in Section~\ref{sec:joint-variants}, and the tables report the mean over three random seeds.
All text and word positions are produced by Amazon Textract from the pages of each condition.

Two linear baselines, each a single run, show the accuracy of a pipeline of low engineering complexity, in which no neural network is trained: a linear SVM \citep{liblinear} on TF-IDF features \citep{salton-buckley-1988} of the OCR text, and the same SVM on these features concatenated with frozen SigLIP image embeddings \citep{siglip-2023}, which is the most accurate of the linear SVMs that we trained (Tables~\ref{tab:svm-baselines} and~\ref{tab:svm-baselines-n} in Appendix~\ref{app:joint} show all of them).
The remaining twelve models are evaluated with their published weights.
DocXClassifier \citep{Saifullah2024-docxclassifier} and Donut \citep{kim2022donut} are public checkpoints that their authors fine-tuned on the original RVL-CDIP training split; we use them as released and did not retrain them, because of the computational cost of doing so for every condition.

The other ten of these twelve models are zero-shot classifiers, which are not trained on RVL-CDIP and choose among the sixteen category names.
Four of them are vision-language models that classify the page image, CLIP \citep{clip-2021}, ALIGN \citep{align}, SigLIP \citep{siglip-2023} and SigLIP~2 \citep{tschannen2025siglip}.
The other five classify the OCR text of the page: Jev~1.13 \citep{typesafe-jev-2026}, a hosted model designed to choose among a given set of options, and four large language models, Llama~4 Maverick \citep{meta2025llama4}, gpt-oss-20b \citep{openai2025gptoss120bgptoss20bmodel}, Qwen~3~32B \citep{yang2025qwen3technicalreport} and Mistral~7B \citep{mistral-7b}, which are given a prompt that lists the category names (Prompt~1 of Appendix~\ref{app:prompts}).
We also give Jev the page image in place of the OCR text, an input that its API offers as a preview, and list this run with the zero-shot image models as \emph{Jev~1.13, image}; elsewhere in the text, Jev without a qualifier refers to the run on the OCR text.
Appendix~\ref{app:promptstats} compares the accuracy of the four language models under the three prompts of that appendix.
The four language models are scored on the pages that have OCR text (38,377 of the 40,000 test pages and 1,001 of the 1,002 pages of RVL-CDIP-\emph{N}); we did not run them on the pages without identification codes because of the cost, so that they do not have entries in the columns in which the codes are removed.
Since these ten models were never trained on RVL-CDIP, they cannot have learned its label errors, its duplicated test pages or its identification codes, and their accuracy changes across the conditions only through the test pages and labels, which makes them a control for the retrained models.
Appendix~\ref{app:training} lists the pretrained weights, the input and the hyperparameters of every model.

\subsection{Corpus and Protocol Variants}\label{sec:joint-variants}

We evaluate every condition on the de-identified corpus of Section~\ref{sec:deid}.
In the \emph{original} condition, we keep the labels, duplicates and identification codes of the published corpus, so that this condition differs from it only in the replaced sensitive data.
In the other conditions, we remove the defects of Sections~\ref{sec:axis1} to~\ref{sec:axis3}, singly and together.
Of the 40,000 test pages, the review of Section~\ref{sec:axis1} found a label error on 4,317 (unknown, wrong or mixed) and marked 402 as \emph{unsure}.
In the \emph{dropped} conditions, we remove all 4,719 of these pages.
In the \emph{fixed} conditions, we give the corrected category to the 1,166 pages for which the review assigned one and remove the other 3,553 pages, which are 2,511 unknown, 640 mixed and 402 unsure pages.
In the de-duplicated conditions, we keep only the 25,923 test pages that have no detected duplicate in the training split (Section~\ref{sec:axis2}).
In the code-removal conditions, we replace the codes with white pixels on the page and delete them from the OCR text (Section~\ref{sec:axis3}).

We apply a remediation either to the test pages only or to the training split as well.
In the test-side conditions, we score a model trained on the original training split on the changed test pages or labels, to measure how much of its accuracy depends on the defect.
In the retrained conditions (\emph{+RT}), we train the model and select its checkpoint on training and validation splits with the same remediation, to measure the accuracy of a model that never saw the defect.
After the label correction, 296,605 of the 320,000 training pages and 36,380 of the 40,000 validation pages remain.
We apply de-duplication to the test pages only, since we did not retrain any model on a de-duplicated training split, and the validation split still overlaps with the training split.
In the fully remediated condition, we combine the corrected labels, code removal in training and testing, and the de-duplicated test pages; for the models that we do not retrain, we apply only the test-side changes.
We list the eleven conditions in Table~\ref{tab:conditions} with their numbers of test pages, which differ because we remove pages when we correct the labels and when we de-duplicate.

\begin{table}[tbp]
  \centering\footnotesize
  \caption{The eleven conditions of Table~\ref{tab:joint-remediation}, under
           the column names used there. \emph{Trained on} lists the labels and
           pages of the training and validation splits, and $n$ the number of
           test pages.}
  \label{tab:conditions}
  \setlength{\tabcolsep}{4pt}
  \begin{tabular}{llllr}
    \toprule
    \textbf{Condition} & \textbf{Test pages} & \textbf{Test labels} & \textbf{Trained on} & $n$ \\
    \midrule
    \multicolumn{5}{l}{\emph{All test pages}} \\
    Orig.     & all                         & original  & original labels, with codes  & 40,000 \\
    Drop.     & wrong labels removed        & original  & original labels, with codes  & 35,280 \\
    Fixed     & uncorrectable removed       & corrected & original labels, with codes  & 36,446 \\
    Fixed+RT  & uncorrectable removed       & corrected & corrected labels, with codes & 36,446 \\
    \midrule
    \multicolumn{5}{l}{\emph{De-duplicated test pages}} \\
    Orig.     & de-duplicated               & original  & original labels, with codes  & 25,923 \\
    Drop.     & wrong labels removed, de-dup. & original  & original labels, with codes  & 22,530 \\
    Fixed     & uncorrectable removed, de-dup. & corrected & original labels, with codes  & 23,132 \\
    Fixed+RT  & uncorrectable removed, de-dup. & corrected & corrected labels, with codes & 23,132 \\
    \midrule
    \multicolumn{5}{l}{\emph{Codes removed from all test pages}} \\
    Test      & all                         & original  & original labels, with codes    & 40,000 \\
    +RT       & all                         & original  & original labels, without codes & 40,000 \\
    \midrule
    Full      & uncorrectable removed, de-dup. & corrected & corrected labels, without codes & 23,132 \\
    \bottomrule
  \end{tabular}
\end{table}

We report accuracy micro-averaged over the test pages of a condition.
Besides the test split, we evaluate the four training variants (original or corrected labels, each with or without codes) on the 1,002 pages of RVL-CDIP-\emph{N}, which does not contain identification codes or duplicates of training pages, and we measure how well they separate in-scope pages from the 2,915 out-of-scope pages of RVL-CDIP-\emph{O} (Section~\ref{sec:axis4}).
We give the text and multi-modal models the Textract OCR of the pages of each condition throughout, and we vary the OCR engine only in Section~\ref{sec:joint-ocr}.
The public checkpoints, the zero-shot image models, Jev and the four large language models have a single set of weights and therefore appear only in the test-side conditions.

\subsection{Joint Remediation}
\begin{table}[tbp]
  \centering
  \caption{Accuracy (\%) on the RVL-CDIP test split under the eleven
           conditions of Table~\ref{tab:conditions}, all on the
           de-identified corpus of Section~\ref{sec:deid}. Retrained
           models are means over three seeds (one run for the linear
           baselines), and values marked $\dagger$ apply the test-side
           changes only. Models marked $\ddagger$ are scored on the 38,377
           test pages that have OCR text.}
  \label{tab:joint-remediation}
  \resizebox{\textwidth}{!}{
  \begin{tabular}{lccccccccccc}
    \toprule
    & \multicolumn{4}{c}{\textbf{All test pages}} & \multicolumn{4}{c}{\textbf{De-duplicated}} & \multicolumn{2}{c}{\textbf{Codes removed}} & \\
    \cmidrule(lr){2-5}\cmidrule(lr){6-9}\cmidrule(lr){10-11}
    \textbf{Model} & \textbf{Orig.} & \textbf{Drop.} & \textbf{Fixed} & \textbf{Fixed+RT} & \textbf{Orig.} & \textbf{Drop.} & \textbf{Fixed} & \textbf{Fixed+RT} & \textbf{Test} & \textbf{+RT} & \textbf{Full} \\
    \midrule
    $n$ & 40,000 & 35,280 & 36,446 & 36,446 & 25,923 & 22,530 & 23,132 & 23,132 & 40,000 & 40,000 & 23,132 \\
    \midrule
    \multicolumn{12}{l}{\emph{Image}} \\
    AlexNet           & 89.3 & 92.5 & 90.2 & 92.8 & 85.8 & 90.0 & 88.5 & 90.4 & 84.9 & 88.5 & 90.2 \\
    GoogLeNet         & 89.2 & 92.6 & 90.3 & 93.2 & 86.0 & 90.3 & 88.7 & 91.0 & 85.7 & 88.5 & 90.8 \\
    SqueezeNet        & 88.7 & 92.3 & 90.1 & 92.6 & 85.8 & 90.3 & 88.9 & 90.8 & 83.8 & 87.7 & 90.5 \\
    VGG-16            & 91.3 & 94.3 & 91.9 & 94.6 & 88.1 & 92.1 & 90.5 & 92.6 & 87.5 & 90.8 & 92.3 \\
    ResNet-50         & 90.9 & 94.1 & 91.7 & 94.3 & 87.7 & 91.9 & 90.3 & 92.4 & 87.7 & 90.2 & 92.1 \\
    ResNeXt-50        & 91.3 & 94.4 & 92.0 & 94.7 & 88.2 & 92.3 & 90.7 & 92.7 & 88.2 & 90.6 & 92.6 \\
    DiT-base          & 92.8 & 95.6 & 93.1 & 95.9 & 90.4 & 94.1 & 92.3 & 94.5 & 89.4 & 92.2 & 94.3 \\
    \midrule
    \multicolumn{12}{l}{\emph{Text}} \\
    BERT-base         & 92.3 & 95.3 & 93.1 & 95.7 & 89.4 & 93.4 & 92.0 & 93.8 & 90.9 & 91.4 & 93.4 \\
    RoBERTa-base      & 92.5 & 95.8 & 93.6 & 96.0 & 89.7 & 94.0 & 92.7 & 94.4 & 90.7 & 91.6 & 94.1 \\
    \midrule
    \multicolumn{12}{l}{\emph{Multi-modal}} \\
    LayoutLM-base     & 93.4 & 96.4 & 94.1 & 96.7 & 91.0 & 94.9 & 93.4 & 95.4 & 91.9 & 92.6 & 95.2 \\
    LayoutLMv3-base   & 94.8 & 97.2 & 94.6 & 97.5 & 93.0 & 96.2 & 94.3 & 96.6 & 92.7 & 94.4 & 96.5 \\
    \midrule
    \multicolumn{12}{l}{\emph{Linear baselines}} \\
    TF-IDF SVM        & 88.9 & 93.0 & 91.2 & 93.4 & 84.7 & 90.0 & 89.1 & 90.3 & 88.8 & 88.8 & 90.3 \\
    TF-IDF + SigLIP SVM & 93.1 & 96.8 & 94.8 & 97.1 & 90.6 & 95.5 & 94.4 & 96.0 & 92.5 & 92.8 & 96.0 \\
    \midrule
    \multicolumn{12}{l}{\emph{Public checkpoints}} \\
    DocXClassifier    & 94.0 & 96.7 & 94.1 & -- & 91.7 & 95.3 & 93.4 & -- & 93.2 & -- & 93.4$^\dagger$ \\
    Donut             & 95.2 & 97.6 & 94.9 & -- & 93.4 & 96.6 & 94.6 & -- & 84.2 & -- & 86.0$^\dagger$ \\
    \midrule
    \multicolumn{12}{l}{\emph{Zero-shot, image}} \\
    CLIP ViT-L/14     & 43.7 & 46.8 & 46.6 & -- & 44.9 & 49.2 & 48.9 & -- & 44.1 & -- & 49.4$^\dagger$ \\
    ALIGN             & 47.0 & 51.0 & 51.2 & -- & 47.1 & 52.0 & 52.2 & -- & 47.4 & -- & 53.1$^\dagger$ \\
    SigLIP            & 56.2 & 61.4 & 61.3 & -- & 57.6 & 64.0 & 64.1 & -- & 58.1 & -- & 67.1$^\dagger$ \\
    SigLIP 2          & 54.0 & 58.8 & 58.7 & -- & 53.9 & 59.6 & 59.6 & -- & 54.6 & -- & 60.5$^\dagger$ \\
    Jev 1.13, image   & 72.1 & 79.0 & 78.5 & -- & 73.3 & 81.4 & 81.2 & -- & 71.5 & -- & 80.1$^\dagger$ \\
    \midrule
    \multicolumn{12}{l}{\emph{Zero-shot, text}} \\
    Jev 1.13          & 67.0 & 72.9 & 72.3 & -- & 66.5 & 73.0 & 72.5 & -- & 67.4 & -- & 73.3$^\dagger$ \\
    Llama 4 Maverick$^\ddagger$& 67.6 & 73.5 & 72.8 & -- & 65.3 & 72.0 & 71.6 & -- & -- & -- & -- \\
    gpt-oss-20b$^\ddagger$  & 64.6 & 70.2 & 69.7 & -- & 63.8 & 70.1 & 69.7 & -- & -- & -- & -- \\
    Qwen 3 32B$^\ddagger$   & 62.3 & 67.3 & 66.4 & -- & 62.6 & 68.9 & 68.5 & -- & -- & -- & -- \\
    Mistral 7B$^\ddagger$   & 57.0 & 61.6 & 60.9 & -- & 54.3 & 59.8 & 59.4 & -- & -- & -- & -- \\
    \bottomrule
  \end{tabular}}
\end{table}

Table~\ref{tab:joint-remediation} reports the accuracy of the models of Section~\ref{sec:joint-models} on the test split under the eleven conditions of Section~\ref{sec:joint-variants}.

The first four columns show the effect of the label errors.
Every model is more accurate when the 4,719 test pages with a label error or an unsure label are removed (\emph{Drop.}), by 2.3 to 4.1 points for the retrained models and the public checkpoints and by 3.1 to 6.8 points for the zero-shot models.
When the predictions of the same models are scored against the corrected labels (\emph{Fixed}), the gain over the original condition is smaller, 0.3 to 1.5 points for the image models, BERT, RoBERTa and LayoutLM, 0.1 points for DocXClassifier, and 2.2 and 1.7 points for the two SVMs, and LayoutLMv3 and Donut, the two most accurate models under the original labels, are 0.2 and 0.3 points less accurate than in the original condition.
The corrected test split contains the 1,166 pages whose label was changed, and a model trained on the original labels mostly predicts the original label for them: LayoutLMv3 predicts the original label for 75.8\% of these pages and the corrected label for 15.7\%, against 56.8\% and 34.6\% for the TF-IDF SVM.
The original test split therefore rewards a model for reproducing the errors of the training labels, and the two most accurate models under the original labels are the two that become less accurate when the test labels are corrected.
Retraining on the corrected labels (\emph{Fixed+RT}) adds 2.2 to 2.8 points for every retrained model, so that the retrained models are 2.7 to 4.5 points more accurate than in the original condition, with the smallest total gain for LayoutLMv3 (2.7) and the largest for the two SVMs (4.5 and 4.0).

The next four columns repeat these conditions on the 25,923 test pages for which no training duplicate was detected.
Under the original labels the retrained models and the public checkpoints are 1.8 to 4.2 points less accurate on these pages than on all test pages: the decrease is 2.9 to 3.5 points for the six CNNs, 2.3 to 3.0 points for DiT, BERT, RoBERTa, LayoutLM, DocXClassifier and the SVM on TF-IDF text and SigLIP embeddings, 1.8 points for LayoutLMv3 and Donut, and 4.2 points for the TF-IDF SVM.
The five zero-shot image models and Jev, which do not have access to the training split, change by between $-0.6$ and $+1.4$ points.
Section~\ref{sec:axis2-vs-axis1} suggested that duplicated pages are mostly standardized documents whose type is evident from their structure, which would make them easier to classify for any model; the zero-shot results do not show this, since the de-duplicated pages are not harder to classify for a model that has not seen their duplicates, and the decrease of the trained models is therefore not due to the pages that remain being harder in themselves.
The four large language models, which are scored on the pages with OCR text, change by between $-2.6$ and $+0.3$ points.
With retraining on the corrected labels the decrease is between 0.9 points for LayoutLMv3 and 3.1 points for the TF-IDF SVM.
Since the zero-shot models show that these pages are not harder to classify, this decrease of 0.9 to 3.1 points is the part of a retrained model's accuracy on the full test split that depends on having seen a duplicate of the test page during training.

\begin{figure}[tbp]
  \centering
  \includegraphics[width=\linewidth]{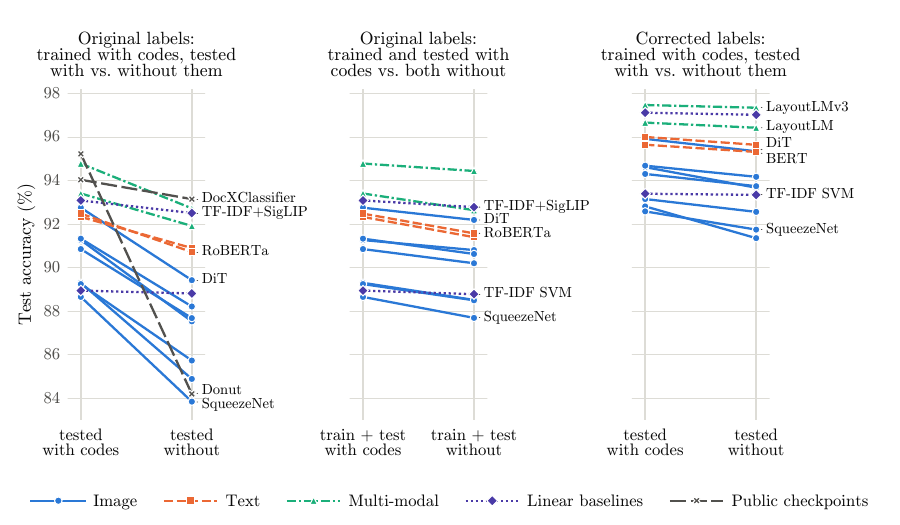}
  \caption{Test accuracy with and without identification codes, one line per
           model. \emph{Left:} models trained on pages with codes and the
           original labels, tested on pages with the codes and with the codes
           removed. \emph{Middle:} the same models against models trained and
           tested on pages without codes. \emph{Right:} the left panel for
           models trained on the corrected labels. Retrained models are means
           over three seeds (one run for the linear baselines).}
  \label{fig:before-after-codes}
\end{figure}

\begin{figure}[tbp]
  \centering
  \includegraphics[width=\linewidth]{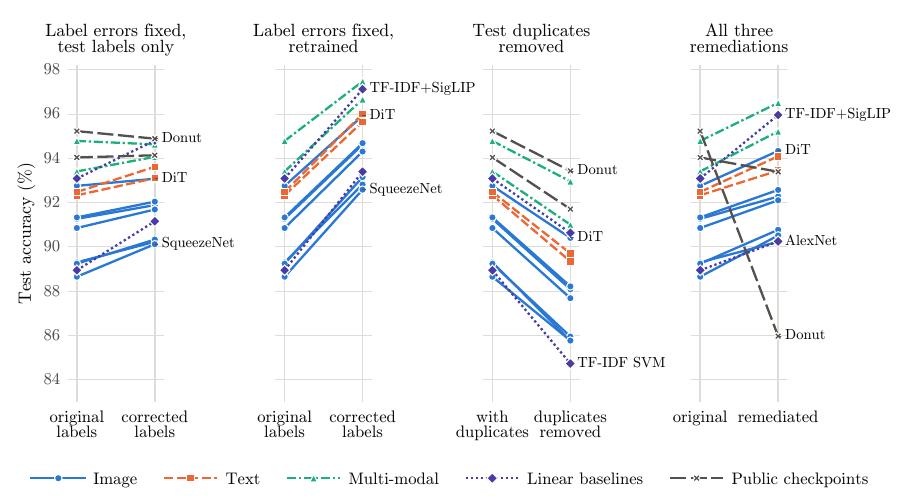}
  \caption{Test accuracy before and after the other remediations, each
           measured against the original condition (original labels, all test
           pages, codes present), one line per model: corrected test labels
           without retraining, corrected labels with retraining, test pages
           with a detected training duplicate removed, and all three
           remediations together. Values are those of
           Table~\ref{tab:joint-remediation}.}
  \label{fig:before-after-remediation}
\end{figure}

Figure~\ref{fig:before-after-codes} shows the effect of the identification codes for each model as a pair of connected points.
Under the original labels, removing the codes from the test pages lowers the accuracy of the seven image models by 3.1 to 4.8 points, BERT and RoBERTa by 1.4 and 1.8 points, LayoutLM and LayoutLMv3 by 1.5 and 2.1 points and the two SVMs by 0.1 and 0.6 points, and it lowers the accuracy of the public Donut checkpoint from 95.2 to 84.2\% and that of DocXClassifier by 0.9 points.
The four zero-shot vision-language models and Jev, which were never trained on RVL-CDIP, are 0.3 to 1.8 points more accurate on the same pages.

Jev with the page image is 0.6 points less accurate, and all of this decrease falls on the \cat{file\_folder} category, in which its accuracy decreases from 79.9 to 69.4\%, whereas its accuracy on the other fifteen categories changes by less than 0.1 points.
Many \cat{file\_folder} pages show a folder cover on which the code is one of the few printed marks, so that removing it leaves a nearly blank page, which Jev then mostly assigns to \cat{scientific\_publication} or \cat{handwritten} (Appendix~\ref{app:joint}).
The trained image models lose 3.3 to 5.2 points on the other fifteen categories and change by at most 0.5 points on \cat{file\_folder}, so that their decrease does not come from these pages.
Apart from the folder covers, the pages therefore do not become harder to classify when the codes are erased, and the lower accuracy of the trained models is due to what these models had learned from the codes.

Models that are trained and tested on pages without codes are less accurate than their counterparts with codes, by 0.5 to 1.0 points for the image models, 0.9 points for BERT and RoBERTa, 0.8 and 0.3 points for LayoutLM and LayoutLMv3 and at most 0.3 points for the SVMs, so that most of the loss at test time is recovered when the codes are also absent from training.
Under the corrected labels the loss at test time is 0.5 to 1.5 points for the image models, 0.3 and 0.4 points for BERT and RoBERTa, 0.2 and 0.1 points for LayoutLM and LayoutLMv3, and at most 0.1 points for the SVMs.
Correcting the labels therefore reduces the effect of the codes on every retrained model, to between a twentieth (LayoutLMv3) and two fifths (the TF-IDF SVM) of its size under the original labels.

Under the original labels, most of the effect falls on the 3,553 unknown, mixed and unsure test pages that we remove in the fixed conditions: on these pages, removing the codes lowers the accuracy of the image models by 11.6 to 18.3 points and that of BERT, RoBERTa, LayoutLM and LayoutLMv3 by 7.8 to 8.6 points, against 2.2 to 3.6 and 0.8 to 1.4 points on the other 36,446 test pages.
A model trained on the original labels thus uses the codes mostly to reproduce the original label of a page whose content does not support that label, which the codes allow because they record the place of the page in the source archive, and that place is correlated with the original label (Section~\ref{sec:axis3}).

Figure~\ref{fig:before-after-remediation} shows the label and duplicate comparisons of Table~\ref{tab:joint-remediation} in the same form, together with the fully remediated condition, in which the models are trained on the corrected labels and on pages without codes and are scored on the corrected, de-duplicated test pages without codes.
In this condition the retrained models are 0.9 to 2.9 points more accurate than in the original condition, because their gain from the label corrections (2.7 to 4.5 points) is larger than their combined loss from the removal of the duplicates and the codes when these are measured under the corrected labels (0.9 to 3.1 and at most 0.4 points); LayoutLM and LayoutLMv3 gain 1.8 and 1.7 points.
The codes contribute little to this difference once the labels are corrected and the models retrained, since the fully remediated accuracy of every retrained model is within 0.4 points of its accuracy on the same de-duplicated pages with codes (\emph{Fixed+RT}, de-duplicated).
Donut, for which no retrained version exists, is 9.3 points less accurate than in the original condition with the test-side changes alone (86.0 against 95.2\%), and DocXClassifier 0.6 points.

\begin{figure}[tbp]
  \centering
  \includegraphics[width=\linewidth]{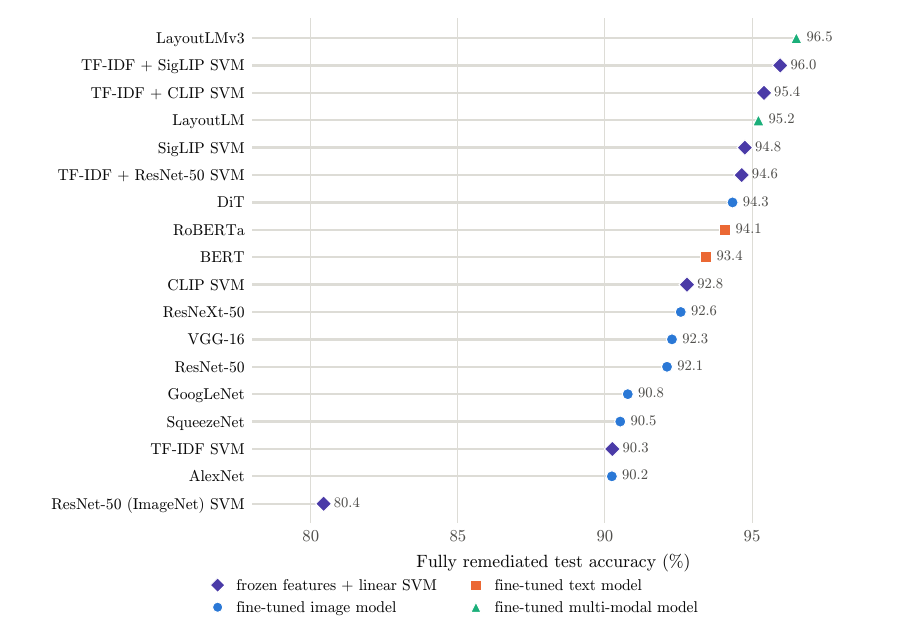}
  \caption{Fully remediated test accuracy (corrected labels, codes removed from
           the training and test pages, test pages with a detected training
           duplicate left out) of the retrained models, with the linear SVMs
           on frozen features set against the fine-tuned models. The SVMs are
           those of Table~\ref{tab:svm-baselines} without the
           handwriting-token and Tesseract variants.}
  \label{fig:remediated-ladder}
\end{figure}

Figure~\ref{fig:remediated-ladder} ranks the retrained models by their fully remediated accuracy and separates the linear SVMs on frozen features from the fine-tuned models.
The SVM on TF-IDF text concatenated with SigLIP embeddings obtains 96.0\%, between LayoutLM at 95.2\% and LayoutLMv3 at 96.5\%, and the same SVM with ImageNet ResNet-50 features in place of the SigLIP embeddings obtains 94.6\%, above DiT at 94.3\%, RoBERTa at 94.1\% and BERT at 93.4\%, although the ResNet-50 features alone yield an accuracy of 80.4\% and the TF-IDF text alone 90.3\% (Table~\ref{tab:svm-baselines}).

\subsection{Accuracy on RVL-CDIP-\emph{N}}\label{sec:joint-n}

\FloatBarrier
\begin{table}[tbp]
  \centering
  \footnotesize
  \caption{Accuracy (\%) on RVL-CDIP-\emph{N}. Left, the retrained models by
           training variant (means over three seeds, one run for the linear
           baselines); right, the public checkpoints and zero-shot models,
           which were not retrained. Models marked $\ddagger$ are scored on the
           1,001 pages that have OCR text. $\Delta$ is the difference
           between \emph{Both} and \emph{Original}.}
  \label{tab:joint-remediation-n}
  \setlength{\tabcolsep}{4pt}
  \begin{tabular}[t]{lccccc}
    \toprule
    \textbf{Model} & \textbf{Original} & \textbf{Fixed} & \textbf{No codes} & \textbf{Both} & $\Delta$ \\
    \midrule
    \multicolumn{6}{l}{\emph{Image}} \\
    AlexNet           & 61.4 & 73.1 & 59.9 & 70.2 & $+$8.8 \\
    GoogLeNet         & 64.0 & 76.1 & 61.7 & 72.9 & $+$8.8 \\
    SqueezeNet        & 66.4 & 76.4 & 63.0 & 73.7 & $+$7.3 \\
    VGG-16            & 66.8 & 79.6 & 67.0 & 77.7 & $+$10.9 \\
    ResNet-50         & 66.6 & 78.1 & 63.6 & 76.7 & $+$10.1 \\
    ResNeXt-50        & 66.8 & 79.5 & 66.6 & 76.8 & $+$10.0 \\
    DiT-base          & 71.2 & 86.9 & 71.4 & 84.2 & $+$13.0 \\
    \midrule
    \multicolumn{6}{l}{\emph{Text}} \\
    BERT-base         & 86.6 & 91.7 & 86.4 & 91.6 & $+$5.0 \\
    RoBERTa-base      & 88.8 & 92.5 & 87.8 & 93.4 & $+$4.7 \\
    \midrule
    \multicolumn{6}{l}{\emph{Multi-modal}} \\
    LayoutLM-base     & 91.0 & 96.7 & 90.0 & 96.6 & $+$5.6 \\
    LayoutLMv3-base   & 83.7 & 97.2 & 85.6 & 96.1 & $+$12.3 \\
    \midrule
    \multicolumn{6}{l}{\emph{Linear baselines}} \\
    TF-IDF SVM        & 84.6 & 85.7 & 84.5 & 86.1 & $+$1.5 \\
    TF-IDF + SigLIP SVM & 89.8 & 97.3 & 92.0 & 97.0 & $+$7.2 \\
    \bottomrule
  \end{tabular}\hspace{1.5em}%
  \begin{tabular}[t]{lc}
    \toprule
    \textbf{Model} & \textbf{Acc.} \\
    \midrule
    \multicolumn{2}{l}{\emph{Public checkpoints}} \\
    DocXClassifier    & 75.0 \\
    Donut             & 78.9 \\
    \midrule
    \multicolumn{2}{l}{\emph{Zero-shot, image}} \\
    CLIP ViT-L/14     & 82.7 \\
    ALIGN             & 87.7 \\
    SigLIP            & 92.3 \\
    SigLIP 2          & 93.1 \\
    Jev 1.13, image   & 95.4 \\
    \midrule
    \multicolumn{2}{l}{\emph{Zero-shot, text}} \\
    Jev 1.13          & 84.8 \\
    Llama 4 Maverick$^\ddagger$ & 87.8 \\
    gpt-oss-20b$^\ddagger$  & 84.3 \\
    Qwen 3 32B$^\ddagger$   & 85.4 \\
    Mistral 7B$^\ddagger$   & 78.2 \\
    \bottomrule
  \end{tabular}
\end{table}

Table~\ref{tab:joint-remediation-n} shows the accuracy of the same models on the 1,002 pages of RVL-CDIP-\emph{N}, which does not have identification codes, training duplicates or label corrections of its own, so that a model's accuracy there depends only on whether it was trained on the original labels and pages, on the corrected labels, on the code-free pages, or on both.
Training on the corrected labels raises the accuracy of every retrained model on RVL-CDIP-\emph{N}, by 10.0 to 15.7 points for the seven image models (from between 61.4 and 71.2\% to between 73.1 and 86.9\%), by 5.1 points for BERT, 3.8 for RoBERTa and 5.7 for LayoutLM, and by 7.5 points for the SVM on TF-IDF text and SigLIP embeddings, while the TF-IDF SVM gains only 1.1 points.
For the image models and LayoutLMv3 these gains are several times as large as their gains on the test split, which are at most 4.0 points, so that for these models the test split understates how much the label errors reduce accuracy on documents from other sources.

On RVL-CDIP-\emph{N}, LayoutLMv3 gains 13.4 points, from 83.7 to 97.2\%, and almost all of this gain is in one category: the three models trained on the original labels classify between 16 and 31\% of the 176 \cat{handwritten} pages of RVL-CDIP-\emph{N} correctly and assign most of the others to \cat{letter}, \cat{memo}, \cat{presentation} and \cat{scientific\_report}, whereas the models trained on the corrected labels classify 93\% of them correctly.
For the SVM on TF-IDF text, the same category gains most from a page-level signal provided by the OCR engine: when two tokens are added to its input, the number of words on the page and the share of them that Textract classifies as handwritten, the SVM trained on the corrected labels classifies 93.8\% of the \cat{handwritten} pages of RVL-CDIP-\emph{N} correctly instead of 65.9\%, and its accuracy on RVL-CDIP-\emph{N} rises from 85.7 to 92.1\%, against 1.2 points on the test split (Appendix~\ref{app:joint}).

Removing the identification codes from the training pages lowers the RVL-CDIP-\emph{N} accuracy of the corrected-label image models by 1.4 to 3.3 points and that of LayoutLMv3 by 1.1, and changes the accuracy of BERT, RoBERTa, LayoutLM and the two SVMs by at most one point.
Under the corrected labels, every retrained model that reads text except the TF-IDF SVM is more accurate on RVL-CDIP-\emph{N} than any fine-tuned image model, with the SVM on TF-IDF text and SigLIP embeddings at 97.3\%, LayoutLMv3 at 97.2\%, LayoutLM at 96.7\%, RoBERTa at 92.5\% and BERT at 91.7\%, whereas the best fine-tuned image model, DiT, obtains 86.9\% and the public DocXClassifier and Donut checkpoints 75.0 and 78.9\%.
On RVL-CDIP-\emph{N}, SigLIP~2 and SigLIP, which were never trained on RVL-CDIP, obtain 93.1 and 92.3\% and so exceed every image model fine-tuned on the corpus, Jev, a zero-shot model that reads the Textract text, obtains 84.8\%, and the four large language models are between 78.2 and 87.8\% accurate, although the two SigLIP models classify only 54.0 and 56.2\% of the original test split correctly (Table~\ref{tab:joint-remediation}).
Given the page image, Jev obtains 95.4\% on RVL-CDIP-\emph{N}, which is more than any model trained on the original labels (at most 91.0\%, for LayoutLM) and less than LayoutLM, LayoutLMv3 and the SVM on TF-IDF text and SigLIP embeddings trained on the corrected labels (96.7 to 97.3\%); most of its gain over the OCR text is on the \cat{handwritten} pages, of which it classifies 80\% correctly instead of 22\%.
A model's accuracy on the RVL-CDIP test split is therefore a poor guide to its accuracy on contemporary documents of the same categories.

\begin{figure}[tbp]
  \centering
  \includegraphics[width=\linewidth]{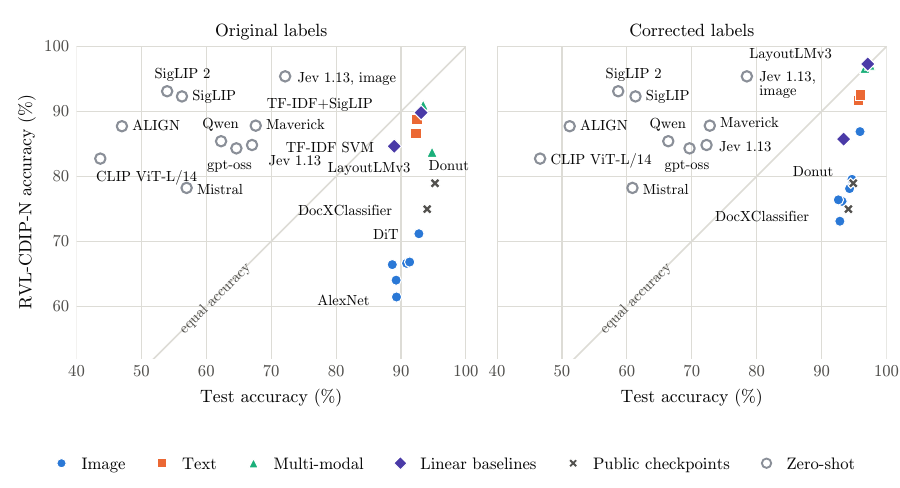}
  \caption{Test accuracy against accuracy on RVL-CDIP-\emph{N}, one point per
           model; points below the diagonal are less accurate on
           RVL-CDIP-\emph{N} than on the test split. \emph{Left:} models
           trained and tested with the original labels. \emph{Right:} models
           retrained on the corrected labels; the public checkpoints and the
           zero-shot models are scored against the corrected test labels and
           keep their single RVL-CDIP-\emph{N} value. The four large language
           models (Maverick, gpt-oss, Qwen, Mistral) are scored on the pages
           that have OCR text.}
  \label{fig:test-vs-n}
\end{figure}

Figure~\ref{fig:test-vs-n} plots the accuracy of each model on the test split against its accuracy on RVL-CDIP-\emph{N}.
Under the original labels the seven fine-tuned image models are 21.6 to 27.9 points less accurate on RVL-CDIP-\emph{N} than on the test split, the DocXClassifier and Donut checkpoints 19.1 and 16.3 points, LayoutLMv3 11.1 points, and BERT, RoBERTa and LayoutLM 5.7, 3.7 and 2.4 points, whereas the four zero-shot vision-language models are 36.1 to 40.7 points more accurate there, Jev 17.8 points with the OCR text and 23.3 points with the page image, and the four large language models 19.7 to 23.1 points.
Under the corrected labels the difference decreases to between 9.0 and 19.7 points for the image models, to 4.0 and 3.5 points for BERT and RoBERTa, and to at most 0.3 points for LayoutLM, LayoutLMv3 and the SVM on TF-IDF text and SigLIP embeddings.
For these three models the difference between the test split and RVL-CDIP-\emph{N} almost disappears when they are trained on the corrected labels, which suggests that much of their lower accuracy on RVL-CDIP-\emph{N} under the original labels is related to the label errors, whereas the image models are still 9.0 to 19.7 points less accurate on RVL-CDIP-\emph{N} when they are trained on the corrected labels.

\subsection{Abstention on RVL-CDIP-\emph{O}}\label{sec:joint-abstention}

This subsection reports the two abstention settings of Section~\ref{sec:axis4}, \emph{T-O} and \emph{N-O}, as AUC under maximum softmax probability and the energy score, for thirteen models trained on the original labels: the five CNNs, BERT, RoBERTa, LayoutLM and LayoutLMv3 of Section~\ref{sec:joint-models}, and the published RVL-CDIP checkpoints of DiT-base, DiT-large, DocXClassifier and Donut.
Four zero-shot vision-language models (CLIP ViT-L/14, ALIGN, SigLIP and SigLIP~2) serve as a control.
Every image of the three sets is converted to grayscale, because the released RVL-CDIP-\emph{O} images are in color and color alone would identify the out-of-scope documents.

\begin{table}[tbp]
\centering\small
\caption{Separability (AUC) of in-domain from out-of-scope documents under maximum softmax probability and the energy score, for models trained on the original labels; Section~\ref{sec:interaction} compares their corrected-label counterparts. \emph{T-O} separates the RVL-CDIP test split (40,000 documents) from RVL-CDIP-\emph{O} (2,915); \emph{N-O} separates RVL-CDIP-\emph{N} (1,002) from RVL-CDIP-\emph{O}. The zero-shot models are never trained on RVL-CDIP. Appendix~\ref{app:ood} shows FPR95 and two further scores.}\label{tab:ood-auc}
\begin{tabular}{llcccc}\toprule
 & & \multicolumn{2}{c}{\textbf{\emph{T-O}}} & \multicolumn{2}{c}{\textbf{\emph{N-O}}} \\
\cmidrule(lr){3-4}\cmidrule(lr){5-6}
\textbf{Input} & \textbf{Model} & \textbf{MSP} & \textbf{Energy} & \textbf{MSP} & \textbf{Energy} \\
\midrule
Image & AlexNet & 0.880 & 0.914 & 0.645 & 0.673 \\
 & GoogLeNet & 0.837 & 0.845 & 0.604 & 0.594 \\
 & ResNet-50 & 0.874 & 0.875 & 0.624 & 0.600 \\
 & ResNeXt-50 & 0.882 & 0.889 & 0.618 & 0.612 \\
 & VGG-16 & 0.889 & 0.930 & 0.661 & 0.668 \\
 & DiT-base & 0.895 & 0.892 & 0.733 & 0.759 \\
 & DiT-large & 0.814 & 0.825 & 0.691 & 0.731 \\
 & DocXClassifier & 0.897 & 0.808 & 0.707 & 0.626 \\
 & Donut & 0.909 & 0.960 & 0.793 & 0.840 \\
\midrule
Text & BERT & 0.859 & 0.845 & 0.841 & 0.858 \\
 & RoBERTa & 0.868 & 0.857 & 0.838 & 0.871 \\
\midrule
Multi-modal & LayoutLM & 0.861 & 0.865 & 0.834 & 0.858 \\
 & LayoutLMv3 & 0.893 & 0.907 & 0.827 & 0.838 \\
\midrule
Zero-shot VLM & CLIP ViT-L/14 & 0.623 & 0.861 & 0.806 & 0.756 \\
 & ALIGN & 0.655 & 0.612 & 0.861 & 0.666 \\
 & SigLIP & 0.710 & 0.862 & 0.890 & 0.812 \\
 & SigLIP 2 & 0.673 & 0.797 & 0.916 & 0.820 \\
\bottomrule\end{tabular}\end{table}

The results of the RVL-CDIP-\emph{O} experiments are shown in Table~\ref{tab:ood-auc}.
In the \emph{T-O} setting, every trained model separates reasonably well, with AUC between 0.808 and 0.960 across the thirteen models and both confidence functions.
In the \emph{N-O} setting, most of this separation is absent for the trained image models: the AUC of the five CNNs is lower by 0.23 to 0.28, at 0.594 to 0.673, close to the 0.5 of a confidence that does not contain information about domain membership, and that of Donut, the best model in the \emph{T-O} setting, decreases from 0.960 to 0.840.
The text models change little: BERT, RoBERTa and LayoutLM change by at most 0.03, and LayoutLMv3, which also reads the page image, by 0.07.

Our interpretation is that the image models have overfit to the appearance of RVL-CDIP pages: their confidence is high on pages that look like RVL-CDIP scans, so that in the \emph{T-O} setting they separate the cleaner, more recent and born-digital RVL-CDIP-\emph{O} documents by their appearance, and this cue is absent when the in-domain pages are also unlike RVL-CDIP.
The text models, which do not use the appearance of the page, are the control for this interpretation, and their separation is almost unchanged between the two settings.
The duplicates of Section~\ref{sec:axis2} may contribute, since a test page with a near-copy in the training split is especially familiar to the model, but we did not measure separability on the de-duplicated test pages.

Although the text models keep their separation in the \emph{N-O} setting, the separation of all thirteen models is too weak for reliable abstention: at the threshold that keeps 95\% of RVL-CDIP-\emph{N}, the best of the thirteen models still accepts 59.5\% of RVL-CDIP-\emph{O} (Appendix~\ref{app:ood}), and the \emph{N-O} setting is the one that corresponds to a model deployed on contemporary documents.

The zero-shot models are a second control.
Their confidence functions are computed in the same way as for the trained models, from the sixteen image--caption similarity scores, scaled by the model's learned temperature (and for SigLIP and SigLIP~2 offset by its learned bias), in place of logits, and a page is set aside when the maximum softmax probability or the energy score is below a threshold.
Under maximum softmax probability their pattern is the reverse of that of the trained image models: they separate the \emph{N-O} setting well (0.806 to 0.916) and the \emph{T-O} setting poorly (0.623 to 0.710), because they classify only 44 to 56\% of the RVL-CDIP test pages correctly, against 83 to 93\% of RVL-CDIP-\emph{N}, and are correspondingly unsure of them, which indicates that the \emph{T-O} separation of the trained image models results from their training on RVL-CDIP; this reversal depends in part on the confidence function (Appendix~\ref{app:ood}).

\begin{table}[tbp]
  \centering\footnotesize\setlength{\tabcolsep}{3.8pt}\renewcommand{\arraystretch}{0.9}
  \caption{Out-of-scope separability (AUC, maximum softmax probability) by
           training variant: \emph{T-O} is the RVL-CDIP test split against
           RVL-CDIP-\emph{O}, \emph{N-O} is RVL-CDIP-\emph{N} against it.
           The public checkpoints and the zero-shot models were not retrained
           and have one value per setting, placed under \emph{Original};
           Table~\ref{tab:ood-variants-energy} shows the energy score.}
  \label{tab:ood-variants}
  \begin{tabular}{lcccccccc}
    \toprule
    & \multicolumn{4}{c}{\textbf{\emph{T-O}}} & \multicolumn{4}{c}{\textbf{\emph{N-O}}} \\
    \cmidrule(lr){2-5}\cmidrule(lr){6-9}
    \textbf{Model} & \textbf{Original} & \textbf{Fixed} & \textbf{No codes} & \textbf{Both} & \textbf{Original} & \textbf{Fixed} & \textbf{No codes} & \textbf{Both} \\
    \midrule
    \multicolumn{9}{l}{\emph{Image}} \\
    AlexNet           & 0.877 & 0.905 & 0.872 & 0.902 & 0.643 & 0.731 & 0.645 & 0.717 \\
    GoogLeNet         & 0.860 & 0.884 & 0.855 & 0.876 & 0.620 & 0.702 & 0.619 & 0.688 \\
    SqueezeNet        & 0.877 & 0.899 & 0.873 & 0.901 & 0.648 & 0.744 & 0.642 & 0.738 \\
    VGG-16            & 0.899 & 0.924 & 0.887 & 0.918 & 0.671 & 0.763 & 0.656 & 0.743 \\
    ResNet-50         & 0.884 & 0.901 & 0.879 & 0.898 & 0.641 & 0.718 & 0.638 & 0.719 \\
    ResNeXt-50        & 0.890 & 0.908 & 0.882 & 0.903 & 0.635 & 0.730 & 0.637 & 0.718 \\
    DiT-base          & 0.885 & 0.914 & 0.877 & 0.917 & 0.709 & 0.787 & 0.704 & 0.782 \\
    \midrule
    \multicolumn{9}{l}{\emph{Text}} \\
    BERT-base         & 0.838 & 0.875 & 0.831 & 0.873 & 0.814 & 0.872 & 0.799 & 0.874 \\
    RoBERTa-base      & 0.845 & 0.876 & 0.844 & 0.876 & 0.822 & 0.894 & 0.806 & 0.901 \\
    \midrule
    \multicolumn{9}{l}{\emph{Multi-modal}} \\
    LayoutLM-base     & 0.846 & 0.889 & 0.847 & 0.889 & 0.821 & 0.912 & 0.810 & 0.910 \\
    LayoutLMv3-base   & 0.890 & 0.930 & 0.888 & 0.921 & 0.832 & 0.924 & 0.829 & 0.896 \\
    \midrule
    \multicolumn{9}{l}{\emph{Linear baselines}} \\
    TF-IDF SVM        & 0.936 & 0.943 & 0.935 & 0.943 & 0.857 & 0.862 & 0.857 & 0.864 \\
    TF-IDF + SigLIP SVM & 0.925 & 0.948 & 0.924 & 0.950 & 0.883 & 0.947 & 0.881 & 0.944 \\
    \midrule
    \multicolumn{9}{l}{\emph{Public checkpoints}} \\
    DocXClassifier    & 0.897 & -- & -- & -- & 0.707 & -- & -- & -- \\
    Donut             & 0.934 & -- & -- & -- & 0.806 & -- & -- & -- \\
    \midrule
    \multicolumn{9}{l}{\emph{Zero-shot, image}} \\
    CLIP ViT-L/14     & 0.623 & -- & -- & -- & 0.806 & -- & -- & -- \\
    ALIGN             & 0.655 & -- & -- & -- & 0.861 & -- & -- & -- \\
    SigLIP            & 0.710 & -- & -- & -- & 0.890 & -- & -- & -- \\
    SigLIP 2          & 0.673 & -- & -- & -- & 0.916 & -- & -- & -- \\
    Jev 1.13, image   & 0.742 & -- & -- & -- & 0.912 & -- & -- & -- \\
    \midrule
    \multicolumn{9}{l}{\emph{Zero-shot, text}} \\
    Jev 1.13          & 0.709 & -- & -- & -- & 0.860 & -- & -- & -- \\
    \bottomrule
  \end{tabular}
\end{table}

Table~\ref{tab:ood-variants} shows the AUC under maximum softmax probability for the retrained models of this section, by training variant.
Training on the corrected labels raises the AUC of every retrained model in both settings, by 0.007 to 0.043 in the \emph{T-O} setting and by 0.005 to 0.096 in the \emph{N-O} setting, where the image models gain 0.077 to 0.096 but remain between 0.702 and 0.787.
Removing the identification codes from training changes the AUC by at most 0.016 in either setting, and the variant with both changes is within 0.028 of the corrected-label variant.
With the corrected labels, the SVM on TF-IDF text and SigLIP embeddings separates the \emph{N-O} setting best (0.947), followed by LayoutLMv3 (0.924), and both exceed SigLIP~2 (0.916); Section~\ref{sec:interaction} examines why the label correction raises separability.

\subsection{OCR Engine Comparison}\label{sec:joint-ocr}

Because RVL-CDIP was released without OCR transcriptions (Section~\ref{sec:axis5}), every text or layout model depends on an OCR engine chosen by its user, and we measure the effect of that choice by training identical models on two text layers covering the full corpus: our own Tesseract 4.1.1 run, with automatic page segmentation and no preprocessing, and AWS Textract.
Both layers provide tokens and bounding boxes.
We train BERT \citep{devlin-etal-2019-bert}, RoBERTa \citep{roberta}, LayoutLM \citep{xu-2020-layoutlm} and LayoutLMv3 \citep{layoutlmv3-huang-2022}, together with TF-IDF features and a linear SVM as an unpretrained lower bound, once on the output of each OCR engine, with the same architecture, hyperparameters, schedule and seeds.
We score each model on the test split under both the original and the corrected labels of Section~\ref{sec:axis1}, and on RVL-CDIP-\emph{N} (Section~\ref{sec:axis4}), where an in-distribution gain without an out-of-distribution one would indicate that the effect is specific to the corpus and is not an improvement in document understanding.
Since the engines differ in reading order and line grouping as well as in character accuracy, the comparison for the text-only models is between whole text layers, and for LayoutLM it also covers the coordinates each engine reports.

\begin{table}[tbp]
  \centering
  \caption{Accuracy by OCR engine, holding architecture fixed, on the RVL-CDIP
           test split. Tesseract is our own 4.1.1 run. \emph{Original} uses the
           labels as released; \emph{errors fixed} uses the corrected labels of
           Section~\ref{sec:axis1}. $\Delta$ is Textract minus Tesseract, in
           accuracy points. Every $\Delta$ is significant under a paired
           McNemar test at $p < 10^{-200}$. The first three models read text
           alone; LayoutLM also reads the coordinates the engine assigns.}
  \label{tab:ocr-comparison}
  \scalebox{0.94}{
  \begin{tabular}{lcccccc}
    \toprule
    & \multicolumn{3}{c}{\textbf{Original labels}} & \multicolumn{3}{c}{\textbf{Errors fixed}} \\
    \cmidrule(lr){2-4}\cmidrule(lr){5-7}
    \textbf{Model} & \textbf{Tesseract} & \textbf{Textract} & \textbf{\boldmath$\Delta$} & \textbf{Tesseract} & \textbf{Textract} & \textbf{\boldmath$\Delta$} \\
    \midrule
    TF-IDF SVM & 81.96 & 90.23 & +8.27 & 86.21 & 94.07 & +7.86 \\
    BERT-base          & 86.66 & 92.54 & +5.88 & 90.23 & 95.97 & +5.75 \\
    RoBERTa-base       & 86.50 & 92.47 & +5.97 & 90.38 & 95.97 & +5.59 \\
    \midrule
    LayoutLM-base      & 89.39 & 93.50 & +4.11 & 92.43 & 96.99 & +4.56 \\
    \bottomrule
  \end{tabular}}
\end{table}

Textract raises the accuracy of every model in Table~\ref{tab:ocr-comparison} by 4.1 to 8.3 points, five to forty times the 0.19 to 0.82 points separating successive LayoutLM generations, while BERT and RoBERTa, despite different pretraining corpora and tokenizers, differ by at most 0.16 points in any column, so that the engine that produced the text has a larger effect than the choice between these models.
Reported architectural improvements may be real, but they are several times smaller than the variation that an unrecorded choice of engine introduces.
The gain is largest for TF-IDF (about 8 points), which cannot recover from a garbled token, smaller for the pretrained text transformers (5.6 to 6.0 points), and smallest for LayoutLM (4.1 to 4.6 points), whether because token positions are affected less by a change of engine than characters are or because its 97.0\% Textract score under the corrected labels is near ceiling; in either case its gap is still about five times the difference between two model generations.
Correcting the labels raises every score but changes each gap by less than half a point and not in a consistent direction, so the OCR effect is not an artifact of label noise and is close to independent of it (Section~\ref{sec:interaction}).

Part of Textract's advantage may come from the identification codes: the codes are short, often rotated and stamped over other content, and we expect two OCR engines to disagree more on text of this kind than on the body text of a page, so that Textract may transcribe the codes more accurately than Tesseract does.
A TF-IDF SVM trained and tested on the de-identified pages under the original labels is 9.1 points more accurate on Textract than on Tesseract text when the codes are present (88.9 against 79.8\%) and 9.1 points when they are removed from the training and test pages (88.8 against 79.7\%), and with the codes erased at test time alone it loses 0.1 points on Textract and 0.2 on Tesseract, so that for this model the gap between the engines is not due to the codes.
For the pretrained models the comparison exists on Textract only, where BERT loses 1.4 and RoBERTa 1.8 points under the original labels when the codes are removed at test time (Table~\ref{tab:joint-remediation}) and BERT, RoBERTa and LayoutLM lose between 0.2 and 0.4 points under the corrected ones, against the 4.1 to 6.0 points by which Textract exceeds Tesseract for the same architectures in Table~\ref{tab:ocr-comparison}; the codes can therefore account for at most about a third of the OCR gain, although the direct test, which would retrain these models on Tesseract text of the code-free pages, was not run.

\section{Interaction Between Defects}\label{sec:interaction}

This section examines whether the defects of Sections~\ref{sec:axis1} to~\ref{sec:axis5} are independent of one another.
If they are, their effects on measured accuracy add and each can be remediated separately, and if they are associated, the remediation of one defect changes the effect of another.
We first measure the association between pairs of defects and then compare the fully remediated accuracy of Section~\ref{sec:joint} with the sum of the effects of the single remediations.

\subsection{Pairwise Associations}

\begin{table}[tbp]
  \centering\footnotesize
  \caption{Label status of the test and validation pages against the
           detection of a training duplicate. \emph{Dupl.} is the share of
           pages with a training duplicate, \emph{OR} the odds ratio of
           having one against the pages with a correct label, and
           \emph{OR, strat.} the Mantel--Haenszel odds ratio stratified by
           original category. \emph{Error or unsure} are the pages with a
           label error (unknown, wrong or mixed) or an unsure label in the
           review of Section~\ref{sec:axis1}; the review assigned a corrected
           category to the \emph{relabeled} pages.}
  \label{tab:defect-interaction}
  \setlength{\tabcolsep}{3.5pt}
  \begin{tabular}{lrcccrccc}
    \toprule
    & \multicolumn{4}{c}{\textbf{Test}} & \multicolumn{4}{c}{\textbf{Validation}} \\
    \cmidrule(lr){2-5}\cmidrule(lr){6-9}
    \textbf{Label} & $n$ & \textbf{Dupl. (\%)} & \textbf{OR} & \textbf{OR, strat.} & $n$ & \textbf{Dupl. (\%)} & \textbf{OR} & \textbf{OR, strat.} \\
    \midrule
    Correct              & 35,280 & 36.1 & --   & --   & 35,291 & 36.2 & --   & --   \\
    Error or unsure      & 4,719  & 28.1 & 0.69 & 0.67 & 4,709  & 29.5 & 0.74 & 0.75 \\
    \quad relabeled      & 1,166  & 48.4 & 1.66 & 0.98 & 1,089  & 49.0 & 1.70 & 0.96 \\
    \quad not relabeled  & 3,553  & 21.4 & 0.48 & 0.57 & 3,620  & 23.6 & 0.55 & 0.68 \\
    \bottomrule
  \end{tabular}
\end{table}

Table~\ref{tab:defect-interaction} relates the label status of each test and validation page to whether a training duplicate was detected for it.
Pages with a label error or an unsure label have a training duplicate less often than pages with a correct label, 28.1 against 36.1\% in the test split, with an odds ratio of 0.69 (95\% interval 0.65 to 0.74) and of 0.67 when the comparison is stratified by category, and in the validation split the two odds ratios are 0.74 and 0.75.
The two groups of these pages differ: the unknown, mixed and unsure pages are duplicated at 21.4\%, whereas the relabeled pages are duplicated at 48.4\%, which is a consequence of the categories in which they occur, since their stratified odds ratio is 0.98.
The association also differs between categories: in the test split the odds ratio and its interval are below one in nine of the sixteen categories and above one in \cat{file\_folder}, \cat{handwritten} and \cat{email}, three categories in which at most 5.3\% of the correctly labeled pages have a duplicate, and across the sixteen categories the two rates are uncorrelated ($r = -0.05$), as Section~\ref{sec:axis2-vs-axis1} found for the whole corpus.

The pair of label errors and OCR engine (Axes I and V) is measured in Section~\ref{sec:joint-ocr}, which crosses the two OCR engines with the original and the corrected label sets: the Textract advantage is 5.88 points under the original labels and 5.75 under the corrected ones for BERT, 5.97 and 5.59 for RoBERTa, 8.27 and 7.86 for TF-IDF, and 4.11 and 4.56 for LayoutLM (Table~\ref{tab:ocr-comparison}).
Correcting the labels reduces the OCR effect by at most 0.41 points for the first three models, which is less than a ceiling effect alone would explain at accuracies in the mid-nineties, and increases it by 0.45 points for LayoutLM.
Axes I and V are therefore close to independent: fixing the labels does not fix the OCR confound, and a corrected corpus evaluated with an unreported engine remains as incomparable across papers as the original.

For label errors and distribution (Axes I and IV), the interaction occurs through accuracy alone (Table~\ref{tab:ood-fixed}).
Retraining on the corrected labels raises \emph{N-O} separability in eight of the nine model pairs of Section~\ref{sec:joint-abstention}, by 0.014 to 0.106 AUC depending on the model and confidence function, while \emph{T-O} separability changes by between $-0.041$ and $+0.007$.
The mean confidence that the models assign shows the source of the gain: corrected-label models are more confident on RVL-CDIP-\emph{N}, as their higher accuracy there would predict, but in all nine pairs they are also more confident on RVL-CDIP-\emph{O}.
The separation increases because the confidence on the in-domain side increases, and the models do not learn to decline what they cannot classify: correcting the labels makes a model a better classifier of in-scope documents from other sources and does not make it more likely to reject out-of-scope ones.
LayoutLM, whose accuracy on RVL-CDIP-\emph{N} gains least from the correction (1.9 points), is the one pair in which \emph{N-O} separability decreases.

\begin{table}[tbp]
\centering\footnotesize\setlength{\tabcolsep}{3.8pt}
\caption{Retraining on the corrected labels and out-of-scope separability. Each row pairs a model trained on the original labels with the same architecture retrained on the corrected labels. Accuracy on RVL-CDIP-\emph{N} (\%) and the mean maximum softmax probability on RVL-CDIP-\emph{N} and RVL-CDIP-\emph{O} are given as original~$\to$~corrected; $\Delta$AUC is corrected minus original.}\label{tab:ood-fixed}
\begin{tabular}{lccccccc}\toprule
 & & \multicolumn{2}{c}{\textbf{Mean MSP}} & \multicolumn{2}{c}{\textbf{\boldmath$\Delta$AUC \emph{N-O}}} & \multicolumn{2}{c}{\textbf{\boldmath$\Delta$AUC \emph{T-O}}} \\
\cmidrule(lr){3-4}\cmidrule(lr){5-6}\cmidrule(lr){7-8}
\textbf{Model} & \textbf{Accuracy \emph{N}} & \textbf{\emph{N}} & \textbf{\emph{O}} & \textbf{MSP} & \textbf{Energy} & \textbf{MSP} & \textbf{Energy} \\ \midrule
AlexNet & 62.1\,$\to$\,72.9 & 0.884\,$\to$\,0.930 & 0.821\,$\to$\,0.837 & $+0.096$ & $+0.078$ & $+0.007$ & $+0.005$ \\
GoogLeNet & 63.1\,$\to$\,73.5 & 0.900\,$\to$\,0.933 & 0.861\,$\to$\,0.870 & $+0.080$ & $+0.067$ & $+0.005$ & $-0.015$ \\
ResNet-50 & 62.1\,$\to$\,75.5 & 0.902\,$\to$\,0.932 & 0.858\,$\to$\,0.876 & $+0.066$ & $+0.083$ & $-0.006$ & $-0.014$ \\
ResNeXt-50 & 65.2\,$\to$\,77.9 & 0.893\,$\to$\,0.944 & 0.846\,$\to$\,0.865 & $+0.106$ & $+0.090$ & $-0.006$ & $-0.016$ \\
VGG-16 & 67.0\,$\to$\,80.3 & 0.933\,$\to$\,0.963 & 0.884\,$\to$\,0.892 & $+0.098$ & $+0.096$ & $-0.008$ & $-0.006$ \\
BERT & 88.1\,$\to$\,91.3 & 0.987\,$\to$\,0.991 & 0.929\,$\to$\,0.941 & $+0.026$ & $+0.014$ & $-0.005$ & $-0.011$ \\
RoBERTa & 89.8\,$\to$\,94.9 & 0.990\,$\to$\,0.996 & 0.927\,$\to$\,0.952 & $+0.070$ & $+0.036$ & $-0.017$ & $-0.024$ \\
LayoutLM & 90.3\,$\to$\,92.2 & 0.976\,$\to$\,0.980 & 0.850\,$\to$\,0.902 & $-0.032$ & $-0.029$ & $-0.041$ & $-0.038$ \\
LayoutLMv3 & 81.3\,$\to$\,95.1 & 0.959\,$\to$\,0.985 & 0.833\,$\to$\,0.878 & $+0.048$ & $+0.072$ & $+0.002$ & $+0.002$ \\
\bottomrule\end{tabular}\end{table}

Identification codes and OCR engine (Axes III and V) are the one case where the two defects could be the same defect.
Identification codes are text, so that an engine that reads them better gives a model a stronger shortcut and does not thereby give it a better transcription of the document; the comparisons in Section~\ref{sec:joint-ocr} indicate that this accounts for none of the OCR advantage of a TF-IDF SVM and for at most about a third of that of the pretrained text models, for which the direct comparison on Tesseract text was not run.

\subsection{Additivity of Confounds}

\begin{table}[tbp]
  \centering\footnotesize\setlength{\tabcolsep}{4pt}
  \caption{Change in test accuracy (points) of the retrained models against
           the original condition: under each remediation alone (corrected
           labels with retraining, codes removed from the training and test
           pages, de-duplicated test pages), their sum, and all three
           together, with the difference between the last two. The last three
           columns show the interaction of each pair of remediations (L
           labels, C codes, D de-duplication), which is the amount by which
           the change under the pair differs from the sum of its two single
           changes. Values are means over three seeds (one run for the linear
           baselines).}
  \label{tab:additivity}
  \begin{tabular}{lccccccccc}
    \toprule
    & \multicolumn{3}{c}{\textbf{Alone}} & & & & \multicolumn{3}{c}{\textbf{Pairs}} \\
    \cmidrule(lr){2-4}\cmidrule(lr){8-10}
    \textbf{Model} & \textbf{Labels} & \textbf{Codes} & \textbf{De-dup.} & \textbf{Sum} & \textbf{All three} & \textbf{Difference} & \textbf{L, C} & \textbf{L, D} & \textbf{C, D} \\
    \midrule
    \multicolumn{10}{l}{\emph{Image}} \\
    AlexNet             & $+$3.5 & $-$0.8 & $-$3.5 & $-$0.8 & $+$0.9 & $+$1.7 & $+$0.7 & $+$1.1 & $-$0.3 \\
    GoogLeNet           & $+$3.9 & $-$0.7 & $-$3.3 & $-$0.1 & $+$1.5 & $+$1.7 & $+$0.6 & $+$1.1 & $-$0.3 \\
    SqueezeNet          & $+$3.9 & $-$1.0 & $-$2.9 & $+$0.1 & $+$1.9 & $+$1.8 & $+$0.8 & $+$1.1 & $-$0.3 \\
    VGG-16              & $+$3.4 & $-$0.5 & $-$3.2 & $-$0.3 & $+$1.0 & $+$1.3 & $+$0.3 & $+$1.1 & $-$0.2 \\
    ResNet-50           & $+$3.5 & $-$0.7 & $-$3.2 & $-$0.4 & $+$1.3 & $+$1.6 & $+$0.5 & $+$1.2 & $-$0.2 \\
    ResNeXt-50          & $+$3.4 & $-$0.7 & $-$3.1 & $-$0.5 & $+$1.2 & $+$1.7 & $+$0.5 & $+$1.2 & $-$0.3 \\
    DiT-base            & $+$3.2 & $-$0.6 & $-$2.3 & $+$0.3 & $+$1.6 & $+$1.3 & $+$0.4 & $+$1.0 & $-$0.2 \\
    \midrule
    \multicolumn{10}{l}{\emph{Text}} \\
    BERT-base           & $+$3.3 & $-$0.9 & $-$3.0 & $-$0.6 & $+$1.1 & $+$1.7 & $+$0.7 & $+$1.2 & $-$0.4 \\
    RoBERTa-base        & $+$3.5 & $-$0.9 & $-$2.8 & $-$0.2 & $+$1.6 & $+$1.8 & $+$0.7 & $+$1.2 & $-$0.4 \\
    \midrule
    \multicolumn{10}{l}{\emph{Multi-modal}} \\
    LayoutLM-base       & $+$3.3 & $-$0.8 & $-$2.4 & $+$0.1 & $+$1.8 & $+$1.7 & $+$0.7 & $+$1.1 & $-$0.3 \\
    LayoutLMv3-base     & $+$2.7 & $-$0.3 & $-$1.8 & $+$0.5 & $+$1.7 & $+$1.2 & $+$0.3 & $+$0.9 & $-$0.2 \\
    \midrule
    \multicolumn{10}{l}{\emph{Linear baselines}} \\
    TF-IDF SVM          & $+$4.5 & $-$0.2 & $-$4.2 & $+$0.1 & $+$1.3 & $+$1.2 & $+$0.1 & $+$1.1 & 0.0 \\
    TF-IDF + SigLIP SVM & $+$4.0 & $-$0.3 & $-$2.5 & $+$1.3 & $+$2.9 & $+$1.6 & $+$0.3 & $+$1.4 & $-$0.1 \\
    \bottomrule
  \end{tabular}
\end{table}

Table~\ref{tab:additivity} compares, for each retrained model, the change in test accuracy under all three remediations with the sum of the changes under each remediation alone, all measured against the original condition of Table~\ref{tab:joint-remediation}.
The fully remediated accuracy of every model is 1.2 to 1.8 points higher than the sum of the three single changes would predict.
Most of this difference is due to the pair of label correction and de-duplication (0.9 to 1.4 points): de-duplication lowers the accuracy of the models by 1.8 to 4.2 points under the original labels and by 0.9 to 3.1 points under the corrected labels, because the test pages without a training duplicate contain a larger share of wrong labels (13.1\%) than the pages with one (9.4\%, Table~\ref{tab:defect-interaction}), so that part of the accuracy that the models lose on the de-duplicated pages under the original labels is due to label errors.
The pair of label correction and codes contributes 0.1 to 0.8 points, since the removal of the codes lowers accuracy less under the corrected labels, and the pair of codes and de-duplication between $-0.4$ and $0.0$ points.
Subtracting the separately measured effect of each defect from a reported accuracy would therefore underestimate the accuracy of the model on the fully remediated benchmark by 1.2 to 1.8 points.

\section{Discussion}\label{sec:discussion}

This section discusses what the results mean for readers of published RVL-CDIP results, gives guidance to researchers who continue to use the benchmark, considers whether it should still be used and what a successor would require, and draws lessons for the auditing of other benchmarks.

\subsection{Interpreting RVL-CDIP Scores}

Our audit found five properties of RVL-CDIP that affect a published accuracy.
Approximately \errorRate\ of the labels are wrong, and a model is scored as correct when it reproduces a wrong label and as wrong when it predicts the correct category.
Approximately \dupRate\ of the test pages have a near-duplicate in the training split, and a model can classify these pages by recalling the training page.
An identification code added during the litigation appears on 92\% of the documents and is associated with the label, and a classifier that uses only the codes is 69\% accurate (Section~\ref{sec:axis3}).
All documents come from one industry, one medium, and one era, and an accuracy on them is not evidence about documents from other settings.
Text and layout models require OCR text, which the benchmark does not distribute; the authors of each paper produce it with their own OCR engine, and the choice of engine changes accuracy by 4.1 to 8.3 points in our experiments (Section~\ref{sec:axis5}), so the accuracies of text and layout models from different papers are not comparable.

Correcting the labels raises the measured accuracy, removing the duplicates and removing the identification codes lower it, and changing the OCR engine can raise or lower it.
Because these effects differ in sign and are not additive (Section~\ref{sec:interaction}), a published accuracy cannot be corrected by adjusting it for each defect separately, and Section~\ref{sec:joint} therefore measures the accuracy with all of them remediated at once.

We draw two further conclusions about the published results, which do not depend on the measurements of Section~\ref{sec:joint}.
First, the models at the top of the leaderboard differ by fractions of a point (0.19 to 0.82 points between successive LayoutLM generations), whereas each of the defects that we quantify changes accuracy by several points.
RVL-CDIP in its present form therefore does not show which of these models is more accurate, although they may differ in ability.

Second, a result of 98.8\% indicates that the task is nearly solved only if the remaining 1.2\% of errors reflect the difficulty of the task.
Since approximately \errorRate\ of the corpus is mislabeled, nearly nine times the remaining 1.2\%, a model that scores 98.8\% reproduces most of the wrong labels in the test set, and the score shows how closely the model matches the original labeling rather than how close the task is to being solved.
The out-of-distribution results of Section~\ref{sec:joint} show that the classification of contemporary documents is not close to being solved.

\subsection{Guidance for Users of RVL-CDIP}

We offer the following guidance to researchers who will continue to use this benchmark, which, for current lack of an alternative, will be most of them.

\paragraph{Heterogeneous categories.}
Several categories comprise multiple distinct document types: \cat{resume} contains curricula vitae, biographical sketches, executive biographies, and business cards, and \cat{advertisement} contains print advertisements and broadcast scripts.
This affects any method that assumes that a category name describes its members: zero-shot classification that represents each category by its name and clustering analyses that expect categories to be coherent are both affected, and weak results in those settings may be due to the label set.

\paragraph{Label errors.}
Approximately \errorRate\ of the labels are wrong, a rate that exceeds the error rate of every model at the top of the leaderboard.
It also varies by more than a factor of twenty across categories, from 1.48\% for \cat{resume} to 31.26\% for \cat{letter}, so that per-category results are not comparable with one another.

\paragraph{Documents with several valid labels.}
Approximately 1.5\% of documents have more than one valid label.
A single-label benchmark accepts only one of the valid labels, so a model that predicts another valid label for such a document is scored as wrong.

\paragraph{Test--train duplication.}
Approximately \dupRate\ of the test set is duplicated in training, with the highest rate of any category for \cat{specification}, at 82.2\%.
Because the validation split has the same rate, a random re-partitioning does not reduce the duplication; the duplicates have to be identified first, and then either removed or assigned to the same split as the pages that they duplicate.

\paragraph{Identification codes.}
Identification codes are a usable shortcut: a classifier that uses only features of the codes is 69\% accurate, and the public Donut checkpoint is 11 points less accurate when the codes are removed (95.2 against 84.2\%).
Most of the effect of the identification codes concerns the wrong labels: a model trained on the original labels uses the codes mostly on pages whose label is wrong, and a model trained on the corrected labels loses at most 1.5 points when the codes are removed (Section~\ref{sec:joint}).
Work on explainability, robustness, or shortcut learning that uses this corpus should account for the identification codes explicitly; our code-free version is available for that purpose.

\paragraph{Scope of the corpus.}
The corpus is limited in era, medium, industry, and language:
a score on RVL-CDIP is evidence only about English-language scanned tobacco-industry documents produced before 2006.
Claims of general document classification ability require out-of-distribution evaluation on data such as RVL-CDIP-\emph{N} and RVL-CDIP-\emph{O}.

\paragraph{OCR engine.}
For text and layout models the OCR engine is part of the experimental setup and materially affects the result.
Papers should at least name the engine, and should preferably use a published text layer, so that others can reproduce the input.

\paragraph{Sensitive personal data.}
The corpus contains sensitive personal data.
Care should be taken when redistributing the corpus, and users of RVL-CDIP should prefer de-identified versions, which we show do not measurably change classification accuracy.
Models can also memorize and leak training data \citep{carlini21extracting}, so that the deployment of models trained on this corpus is a privacy question as well as a question of curation.

\paragraph{Release of fine-tuned classifiers.}
Corrected labels are of limited use if the models scored against the old ones cannot be re-scored.
We checked all 34 papers of Table~\ref{tab:survey-datasets} for released weights (Table~\ref{tab:weights}).
Five make an RVL-CDIP classifier available from the authors, and three more exist only as community uploads whose numbers do not match the published ones.
For the remaining twenty-six the reported accuracy cannot be reproduced without repeating the fine-tuning, and in most cases the pre-training as well; eighteen of the 34 do not release any code of their own.
The three highest scores in Table~\ref{tab:model-scores} (LayoutLLM, EAML, and Bi-VLDoc) are all in that group.
Some of these papers do release something: five publish a pre-trained backbone, and two of these, UDOP \citep{tang-2023-udop} and Dessurt \citep{dessurt-2022}, also release RVL-CDIP fine-tuning scripts and dataset loaders but not the resulting checkpoint.
We recommend that authors release both the pre-trained backbone and the fine-tuned classifier, since a reader can reproduce the reported accuracy with the classifier but not with the backbone alone, and that the checkpoints be kept available so that they can be scored on the corrected evaluation data that we provide.
Authors who cannot release a checkpoint should at least release the model's prediction for each test document, preferably with its confidence scores, which is enough to re-score the model against the corrected labels.

\begin{table}[tbp]
  \centering
  \footnotesize
  \caption{Availability of model weights for the 34 papers of
           Table~\ref{tab:survey-datasets}, as of September 2026. Only the
           fourteen papers that release weights of any kind are listed, named
           by the model each reports on RVL-CDIP; the
           other twenty release neither a backbone nor an RVL-CDIP classifier.
           ``Backbone'' is the pre-trained model before task fine-tuning;
           ``classifier'' is a checkpoint that performs the 16-way RVL-CDIP
           task and so permits re-scoring against corrected labels without
           retraining. Papers below the rule release a backbone, and a
           classifier for them is not available from any source.}
  \label{tab:weights}
  \setlength{\tabcolsep}{4pt}
  \begin{tabular}{@{}llll@{}}
  \toprule
  \textbf{Model} & \textbf{Backbone} & \textbf{RVL-CDIP classifier} & \textbf{Availability} \\
  \midrule
  DiT \citep{li-2022-dit-image-transformer}           & authors & authors     & Hugging Face, live \\
  Donut \citep{kim2022donut}                          & authors & authors     & Hugging Face, live \\
  DocXClassifier \citep{Saifullah2024-docxclassifier} & authors & authors     & institutional host, live \\
  StrucTexTv2 \citep{yu2023structextv}                & authors & authors     & host refuses access \\
  Stacked CNN \citep{das-2018}                        & authors & authors     & 2018 cloud link, unverified \\
  LayoutLMv1 \citep{xu-2020-layoutlm}                 & authors & third party & fidelity unverified \\
  LayoutLMv3 \citep{layoutlmv3-huang-2022}            & authors & third party & reports 95.2 vs.\ 95.44 \\
  LiLT \citep{wang-etal-2022-lilt}                    & authors & third party & separate OCR pipeline \\
  ERNIE-Layout \citep{ernie-layout-2022}              & authors & not found   & model zoo path moved \\
  EfficientNet \citep{ferrando-2020}                  & ---     & on request  & no public download \\
  \midrule
  UDOP \citep{tang-2023-udop}                         & authors & none        & releases a fine-tuning script \\
  Dessurt \citep{dessurt-2022}                        & authors & none        & releases a dataset loader \\
  StructuralLM \citep{li-etal-2021-structurallm}      & authors & none        & backbone only \\
  LayoutLMv2 \citep{xu-etal-2021-layoutlmv2}          & authors & none        & backbone only \\
  \bottomrule
  \end{tabular}
\end{table}

\subsection{Should RVL-CDIP Still Be Used?}

A natural response to an audit of this kind is that the benchmark should be abandoned, which we do not recommend, for the following reason.
RVL-CDIP is effectively the only standardized document classification benchmark at its scale, so that retiring it without a replacement would remove the one benchmark that papers share, and comparison across papers would become impossible where it is now imprecise.
A common measure with known flaws is more useful than no common measure.

We recommend instead that researchers use the remediated variants that we release, which are de-identified by default, address four of the five axes directly, and provide both text layers for the fifth, and that they report out-of-distribution results alongside in-distribution ones, using RVL-CDIP-\emph{N} and RVL-CDIP-\emph{O} or comparable data, so that a claim of general capability is supported by evidence.
We also recommend treating the per-category defect profiles in this paper as a guide to which parts of the corpus remain trustworthy: a study confined to \cat{email}, \cat{handwritten}, and \cat{file\_folder} has very little duplication, whereas one centered on \cat{letter} uses a category whose labels are wrong nearly a third of the time.
In parallel, the field should build a successor benchmark, for which the next subsection lists requirements.

\subsection{Requirements for a Successor Benchmark}

Our findings suggest the following requirements, of which the first several restate and update recommendations from our earlier work and the remainder are new.

\begin{enumerate}
\itemsep0em
  \item \textbf{Published annotation guidelines}, released together with the dataset.
  Without them, a label review has to define the categories first, as we did in Section~\ref{sec:guidelines} before the review of Section~\ref{sec:axis1}.
  \item \textbf{Low label error, measured and reported.} A stated error rate, with the method by which it was obtained, is more useful than an unstated assumption of correctness.
  \item \textbf{Multi-label annotation} for documents that have more than one type.
  \item \textbf{Verified split integrity.} Duplication should be checked at construction, when it is cheap to prevent; an audit afterwards is expensive.
  \item \textbf{No sensitive personal data}, verified before release.
  \item \textbf{Documented provenance artifacts.} Anything stamped, watermarked, or otherwise added to documents by the collection process should be identified and reported, since such features are candidate shortcuts by construction (Section~\ref{sec:axis3}).
  \item \textbf{A canonical text layer.} A benchmark that expects text-based models should distribute the text, as it distributes labels and splits.
  Otherwise a necessary input is left to each user and cross-paper comparison is unsound (Section~\ref{sec:axis5}).
  \item \textbf{Paired out-of-distribution evaluation data}, provided from the outset, so that generalization is measured from the first use of the benchmark.
  \item \textbf{Contemporary and born-digital documents} in addition to historical scans, since deployed systems receive such documents.
  \item \textbf{Breadth and multilinguality:} substantially more than 16 categories, and more than one language, to test capabilities that RVL-CDIP cannot test.
\end{enumerate}

This list describes a considerable undertaking, and RVL-CDIP has not been replaced for a decade because assembling 400,000 labeled documents is difficult and assembling them well is more difficult.
A successor does not need to be as large as RVL-CDIP, however: its test set has to be large enough to separate models reliably, and pre-trained models need much less labeled training data than the 320,000 pages of the RVL-CDIP training split, so a much smaller corpus with reviewed labels would serve the purposes listed above.
We consider this a reason to begin the work now, since the alternative is to continue to rely on discovery material that was never intended for this purpose.

\subsection{Lessons for Benchmark Auditing}

In this subsection we describe the findings of the audit that also apply to other benchmarks.

\paragraph{Provenance.}\label{sec:common-cause}
The five problems of RVL-CDIP are consequences of a single fact: the corpus was assembled for legal discovery and subsequently repurposed for machine learning without curation.
Label errors follow from category tags applied retrospectively, by non-specialists, to material that was indexed for litigation (Section~\ref{sec:axis1}).
Test--train duplication follows from repeated scanning campaigns that entered the same physical page into the collection more than once (Section~\ref{sec:axis2}).
Identification codes follow from the Bates stamping that discovery requires (Section~\ref{sec:axis3}).
Distributional narrowness follows from the scope of the litigation, which determined what was disclosed and when (Section~\ref{sec:axis4}).
The presence of sensitive personal data follows from the fact that these documents were never intended to be public (Section~\ref{sec:deid}): they are internal records, such as personnel files and tax forms, that entered the public domain through compulsory disclosure,
and the personal data that such records ordinarily contain was not removed before the disclosure.
The general lesson is that an audit should examine how a corpus was assembled, because this tells an auditor what to look for: corpora from opportunistic sources, such as web scrapes, administrative records, and litigation releases, each have characteristic defects that an auditor can anticipate.

\paragraph{Automated label-error detection.}
Confident learning assumes that errors are the exception, since the confidence scores that it uses come from a model trained on the labels under audit.
Where errors are pervasive and directionally consistent, the model learns the erroneous convention and assigns high confidence to it.
On the RVL-CDIP test split it estimates an error rate of 2.0 to 6.0\% where the review finds a label error or an unsure label on 11.8\% of the pages, and it finds 10.5 to 22.8\% of the wrong labels (Appendix~\ref{app:cleanlab}).
Such tools are appropriate for finding sporadic mistakes, and they fail on a corpus whose category boundaries were never specified.

\paragraph{Defect rates across splits.}
Duplication in RVL-CDIP's validation split matches its test split to within two tenths of a point (Section~\ref{sec:axis2}).
That agreement distinguishes a defect introduced during corpus assembly from one introduced during partitioning, and the distinction determines whether re-splitting is a remedy.
The check requires almost no additional work, and we recommend it for other benchmarks.

\paragraph{Zero-shot models as a control.}
Throughout this paper, models never trained on RVL-CDIP behave in ways that isolate the confounds: the four zero-shot vision-language models and Jev are 0.3 to 1.8 points more accurate when the identification codes are removed from the test pages, and Jev with the page image is less accurate only on folder covers on which the code is one of the few printed marks, whereas every model trained on RVL-CDIP is less accurate (Table~\ref{tab:joint-remediation}).
A model that has not seen the training distribution provides a reference for separating dataset-specific shortcuts from the difficulty of the task, and we recommend including one in an audit for that reason.

\paragraph{Preprocessing left to the user.}
The OCR problem of Section~\ref{sec:axis5} arises in any benchmark that distributes data in a form that models cannot use directly and does not provide the converted form.
Other examples are audio that requires transcription, raw text that requires a tokenizer, and structured records that require feature extraction.
In each case the authors of each paper produce their own conversion, and the resulting numbers are not strictly comparable.
The remedy is for benchmarks to distribute the converted representation as a canonical artifact.
Leaving the conversion to each user has a second disadvantage: the tools are not equally available.
The OCR engines reported in the RVL-CDIP literature range from open-source Tesseract to proprietary cloud services, so that to the extent that conversion quality affects accuracy, the arrangement gives an advantage to groups that can pay for the better tools, and the reported number does not show this.

\paragraph{Cost of exhaustive auditing.}
The label review reported here occupied seven annotators for approximately two years.
We report this because it determines whether audits of this kind can be repeated, and for most benchmarks they cannot be repeated routinely.
This supports the recommendations above: quality assurance requires far less effort at construction time than afterwards, and when a benchmark is released without it, the work is left to the community, which will generally not do it.

\section{Limitations}\label{sec:limitations}

In this section, we describe the limitations of our audit.

\paragraph{Label review.}
Since RVL-CDIP was published without label criteria, we measured our error rates against criteria that we wrote ourselves (Section~\ref{sec:guidelines}).
A different team would draw some category boundaries differently, most of all the boundary between \cat{letter} and \cat{memo}, which contributes more to our error rate than any other boundary.
The 10.5\% figure should therefore be read as ``erroneous with respect to these stated criteria'', and we publish the criteria so that others can see where their boundaries differ from ours.
The review took about two years with a changing group of annotators, and agreement was substantial, with a Fleiss's kappa of 0.74 for category judgments and 0.918 for the presence of sensitive data.
Because the project lead re-examined every judgment that a label was wrong or uncertain and only a sample of the judgments that it was correct, a remaining mistake is more likely to be a missed label error than a spurious one, and our error rate is therefore more likely to be understated than overstated.

\paragraph{Lower bounds and definitions.}
Three of our main results are lower bounds: the duplication rate, the decrease in accuracy when the identification codes are removed, and the label error rate.
Duplicate detection has 100\% precision and 82.9\% recall, and the \dupRate\ duplication rate therefore omits roughly one pair in six.
Removing an identification code with white pixels erases its characters but preserves the position and dimensions of the region, which our analysis shows to be predictive (Section~\ref{sec:axis3-signal}).
The reported decreases in accuracy therefore understate the contribution of the codes by an amount that we have not measured.
The label error rate excludes the \emph{unsure} cases, and counting the \emph{unsure} cases as errors would raise the label error rate from 10.51\% to 11.48\%.
For scanned documents, the term ``duplicate'' does not have an exact definition.
A page scanned twice, a form template filled in with different content, and two documents that follow a common layout convention are not sharply separated.
We state our criterion for a duplicate and publish the adjudicated pairs, so that others can apply a different criterion.

\paragraph{Models and evaluation data.}
We did not re-benchmark every model reported on RVL-CDIP, because many lack public implementations or trained weights, and we used DocXClassifier and Donut, and in the abstention comparison also DiT-base and DiT-large, as published checkpoints without retraining.
DiT, LayoutLM and LayoutLMv3 are initialized with weights pretrained on IIT-CDIP, which contains the RVL-CDIP test pages with their identification codes.
Our remediation of duplicates and codes therefore applies to fine-tuning and evaluation only, and we did not measure the contribution of this pretraining.
The OCR comparison covers Tesseract, the most widely used open-source engine, and the commercial Amazon Textract, and with a third engine a model might be less accurate than with Tesseract or more accurate than with Textract.
RVL-CDIP-\emph{N} and RVL-CDIP-\emph{O} were collected by us following our understanding of the category criteria; to our knowledge they are the only out-of-distribution data for this benchmark that consist of new documents; the other out-of-distribution sets for RVL-CDIP are augmentations of its existing pages.
RVL-CDIP-\emph{N} covers twelve of the sixteen categories, and both sets are small (\rvlcdipnSize\ and \rvlcdipoSize\ documents), and per-category estimates from these two sets are therefore uncertain.

\paragraph{De-identification.}
Our de-identification addresses the practical risk that sensitive personal data can be retrieved from widely mirrored corpora, and it does not provide a formal privacy guarantee such as differential privacy.
An entity that our annotation missed remains in the released data, where it is only harder to locate among realistic synthetic entities, and the original corpora remain available from their existing sources.
Format matching discloses the format and approximate width of each replaced value: a date replaced as \texttt{7/31/84}~$\rightarrow$~\texttt{5/31/55} discloses that the original was a slash-delimited five-digit date, and a replaced address discloses how many lines it occupied.
This is acceptable for the corpora that we release, since the format of a field is rarely identifying on its own, but users of the pipeline should disable format matching where the format itself is sensitive.
The 3.8\% of annotated regions in the categories without replacement text are left blank (Section~\ref{sec:deid}).
On preprinted forms, an annotation box can also span part of the form as well as the value, as in a voucher reading \texttt{April 10 \_\_\_ 19 87}, where the \texttt{19} is preprinted and the \texttt{87} is the value, and the replacement can then be a string in which a date would not be written; we have not quantified how many regions are affected.

\paragraph{Scope of the conclusions.}
The methodological claims of Section~\ref{sec:discussion}, that confident learning fails under systematic error and that provenance predicts defects, are argued from a single corpus and should be read as hypotheses that are well supported in one case.
The sampled and the exhaustive error rates were also obtained by the same group, and part of the difference between the two rates may be due to our own growing familiarity with the corpus.
The category-level comparison of Section~\ref{sec:axis2-vs-axis1} is based on sixteen aggregates and is ecological, and the document-level analysis of Section~\ref{sec:interaction} should be preferred where the two address the same question.

\section{Conclusion}\label{sec:conclusion}

In this paper, we audited RVL-CDIP and found that approximately \errorRate\ of its labels are wrong, that \dupRate\ of the test pages have a duplicate or near-duplicate in the training split, and that a classifier that uses only the identification codes added during the litigation is 69\% accurate.
Image-only models trained on RVL-CDIP are 16 to 28 points less accurate on documents from other sources, the choice of OCR engine changes the accuracy of text and layout models by 4.1 to 8.3 points, and nearly 16,000 documents contain sensitive personal data.
With all defects remediated, the 13 models that we retrain are 0.9 to 2.9 points more accurate than in the original condition, and the effects of the defects on accuracy are not additive.
The differences between the leading published models are smaller than these effects, so the published results on RVL-CDIP do not show which of those models is more accurate.
We release de-identified versions of the corpus with and without identification codes, the corrected labels, the duplicate pairs, the output of two OCR engines, and the out-of-distribution sets RVL-CDIP-\emph{N} and RVL-CDIP-\emph{O}, and we recommend that researchers who continue to use RVL-CDIP evaluate on the corrected data and report out-of-distribution results alongside in-distribution ones.

\impact{%
The corpora that we audit contain personal information about employees, job applicants, and correspondents of tobacco companies, including Social Security numbers, birth dates, and home addresses.
These records became public through litigation to which these people were not parties, and the corpora have since been downloaded and mirrored widely as machine learning training data.
The people concerned did not consent to this use, and their consent can no longer be obtained.
We consider the reduction of this exposure the most important outcome of this work, and the de-identified corpora are therefore the version that we distribute by default.

Documenting where sensitive data occurs in a public corpus is necessary to justify and reproduce its removal, but the same documentation could direct a malicious reader toward the un-redacted originals, which remain available from their existing sources and which our work does not withdraw.
We therefore release the de-identified images and their OCR text but neither the locations nor the values of the sensitive entities, we report statistics only in aggregate, and we do not name any individual or reproduce any document containing sensitive data beyond the examples in Figure~\ref{fig:pii-examples}, whose sensitive fields we have obscured.
Appendix~\ref{sec:artifacts} lists the released artifacts and explains why the de-identified versions are the default.
We judge that the reduction in risk from removing thousands of Social Security numbers from actively used datasets outweighs the residual risk from describing the removal, and we hold that judgment with less confidence than we would like.

The label errors, test--train duplicates, and identification-code shortcut that we measure matter for a different reason: document classifiers built on this benchmark are deployed in settings such as records processing, claims handling, and compliance review, where misclassification has material effects on individuals.
A benchmark that overstates accuracy, and that cannot measure generalization beyond pre-2006 tobacco-industry paperwork, encourages deployment on the basis of evidence that does not support it.
The corrected corpus and the out-of-distribution data we release allow that gap to be measured.

We are conscious that this paper criticizes a resource the field has depended on, and that its authors performed a service in creating and releasing it that far outweighs the defects we document.
Almost none of what we report was reasonably discoverable in 2015: the confounds became consequential only as accuracy approached the corpus's noise floor.
We do not argue that RVL-CDIP should not have been built or used.
We think that a benchmark as widely used as RVL-CDIP should be re-examined periodically. }

\acks{We thank the co-authors of the five earlier papers that this article consolidates \citep{larson-etal-2022-rvlcdipn, larson-etal-2023-evaluation, larson-etal-2024-de, larson-etal-2025-spurious, larson-etal-2026-revising}: Yutong Ai, Ramla Alakraa, Siddharth Betala, Gayatri Chakkithara, Cyrus Desai, Sam Desai, Santiago Pedroza Diaz, Sharad Duwal, Amogh Manoj Joshi, David Kuang, Gordon Lim, Nicole Cornehl Lima, Yash Mathur, Attila Nagy, Temi Okotore, Vedant Padwal, Kaushal Kumar Prajapati, Junjie Shen, Jamiu Tunde Suleiman, Brian Vilnrotter and Yixin Yuan.
This material is based upon work supported by a National Science Foundation Graduate Research Fellowship held by Stefan Larson, by gift funding from Amazon, and by the Undergraduate Research Opportunity Program (UROP) of the University of Michigan.
Any opinions, findings, and conclusions or recommendations expressed in this material are those of the authors and do not necessarily reflect the views of the National Science Foundation or of Amazon.
Amazon had no role in the design of the study, in the collection, analysis, or interpretation of the data, or in the writing of the paper; the authors have no competing interests to declare.
}

\vskip 0.2in
\bibliography{refs}

\appendix
\counterwithin{figure}{section}
\counterwithin{table}{section}
\section{Released Artifacts}\label{sec:artifacts}

In this section, we describe the annotations, corpora, models and code that we release with this paper, and we state where each of them is hosted.
In addition, the website \url{https://rvlcdip-errors.com/} is a web interface to a database of our annotations, in which the reader can browse the documents whose labels we judged to be wrong and the test--train duplicates that we found.

\subsection{Inventory}

\begin{table}[tbp]
  \centering
  \caption{Artifacts released with this paper.}
  \label{tab:artifacts}
  \small
  \begin{tabular}{lll}
  \toprule
  \textbf{Artifact} & \textbf{Contents} & \textbf{Section} \\
  \midrule
  Category criteria           & 16 operational definitions                        & Section~\ref{sec:guidelines} \\
  Label annotations           & 400,000 judgments; error type; corrected         & Section~\ref{sec:axis1} \\
                              & label where determined                            & \\
  Corrected corpus variants   & the conditions of Section~\ref{sec:joint}         & Section~\ref{sec:joint} \\
  Duplicate pairs             & test--train and validation--train pairs           & Section~\ref{sec:axis2} \\
  Duplicate calibration set   & 1,528 pairs judged by hand, 403 matches             & Section~\ref{sec:axis2} \\
  ID code annotations         & 11,102 boxes (RVL-CDIP), 3,774 (Tobacco3482)      & Section~\ref{sec:axis3} \\
  ID code detections          & 446,332 boxes with transcriptions, all of RVL-CDIP & Section~\ref{sec:axis3} \\
  ID code detector            & trained YOLOv8 weights                            & Section~\ref{sec:axis3} \\
  ID code classifiers         & code                                              & Section~\ref{sec:axis3} \\
  ID-code-free corpora        & RVL-CDIP and Tobacco3482                          & Section~\ref{sec:axis3} \\
  RVL-CDIP-\emph{N}           & \rvlcdipnSize\ documents, 12 categories           & Section~\ref{sec:axis4} \\
  RVL-CDIP-\emph{O}           & \rvlcdipoSize\ documents                          & Section~\ref{sec:axis4} \\
  Trained classifiers         & 7 image, 4 text and layout, 12 linear models      & Section~\ref{sec:joint} \\
  De-identified corpora       & five IIT-CDIP-derived datasets                    & Section~\ref{sec:deid} \\
  De-identification pipeline  & code                                              & Section~\ref{sec:deid} \\
  OCR text layers             & Tesseract 4.1.1 and Textract, tokens and boxes    & Section~\ref{sec:axis5} \\
  \bottomrule
  \end{tabular}
\end{table}

Table~\ref{tab:artifacts} lists the artifacts, several of which are described further below.

The \textbf{de-identification pipeline} is publicly available on GitHub.\footnote{\url{https://github.com/gxlarson/doc-redact}}

The \textbf{label annotations} contain our judgment for each of the 400,000 documents.
Where a document was judged \emph{wrong} and a correct label could be determined, that label is included; where it was judged \emph{mixed}, all applicable labels are recorded.
The annotations are hosted on the Hugging Face Hub\footnote{\url{https://huggingface.co/datasets/stefan-hf/rvlcdip-label-fixes}} as one record per document in the order of the RVL-CDIP label files, giving the original label, the judgment, the corrected label, the second category of a mixed document, and the annotators' free-text note, with the exception of the notes that marked a page as containing personal data, which were removed before release.

The \textbf{duplicate pairs} are hosted on the Hugging Face Hub.\footnote{\url{https://huggingface.co/datasets/stefan-hf/rvlcdip-duplicates}}

The \textbf{duplicate calibration set}, which consists of the 1,528 pairs that we judged by hand and our judgments, is available at \url{https://rvlcdip-errors.com/}.

The \textbf{identification-code annotations and detections} and the \textbf{detector} are hosted on the Hugging Face Hub.\footnote{\url{https://huggingface.co/datasets/stefan-hf/rvlcdip-id-codes} and \url{https://huggingface.co/stefan-hf/yolov8n-rvlcdip-idcodes}}
The dataset contains the bounding box and the transcription of every code, for the hand-annotated RVL-CDIP documents and Tobacco3482 and for the 446,332 regions that the detector found on the full RVL-CDIP corpus, each transcribed by OCR of the cropped region, together with the path of the page in the original release so that the records can be joined against the corpus as distributed.
About 1\% of the detected regions are not identification codes, and the transcriptions of eight of them that had the form of a social security number were removed after manual review.

The code for the \textbf{identification-code classifiers} of Section~\ref{sec:axis3-signal} and Appendix~\ref{app:idcode-analysis} is publicly available on GitHub.\footnote{\url{https://github.com/gxlarson/rvlcdip-id-codes}}

\textbf{RVL-CDIP-\emph{N}} and \textbf{RVL-CDIP-\emph{O}} are hosted on the Hugging Face Hub,\footnote{\url{https://huggingface.co/datasets/stefan-hf/rvlcdip-n} and \url{https://huggingface.co/datasets/stefan-hf/rvlcdip-o}} each as the page images together with the Textract text of every page and the raw Textract responses, which contain the word and line boxes.

The \textbf{trained classifiers} are released as weights with the code that loads them.\footnote{\url{https://huggingface.co/stefan-hf/rvlcdip-classifiers}}
The repository contains seven image classifiers (AlexNet, GoogLeNet, ResNet-50, ResNeXt-50, SqueezeNet, VGG-16, and DiT-base), four models that read the OCR output (BERT, RoBERTa, LayoutLM, and LayoutLMv3), and twelve linear support vector machines: three on TF-IDF features of the OCR text (the Textract text with and without tokens describing the amount of handwriting, and the Tesseract text), three on frozen image features (CLIP embeddings, SigLIP embeddings, and the pooled features of a ResNet-50 trained on ImageNet), and six on the TF-IDF features concatenated with each of the three kinds of image features, with and without the handwriting tokens.
Each model is released in four variants that cross the original and the corrected labels with training pages in which the identification codes are kept or removed.
All were trained on the de-identified corpus; for the neural models one of the three training seeds is released, which was fixed in advance and not selected by accuracy, and the support vector machines are deterministic and have a single run.

The Tesseract and Textract \textbf{OCR text layers} are hosted on the Hugging Face Hub together with the page images of the de-identified corpus and of its counterpart without identification codes.\footnote{\url{https://huggingface.co/datasets/stefan-hf/rvlcdip-redact}}
We produced both layers by running the two engines on the de-identified pages.

\subsection{Release Policy for Sensitive Data}

We distribute the de-identified versions of the corpora as the default.
Section~\ref{sec:deid} shows that de-identification has almost no effect on classification: none of the documents in that evaluation changed its predicted label, and the mean changes in confidence were of order $10^{-3}$.
We therefore see no reason for research to prefer the original data.

We do not release the locations, types, or values of the sensitive entities that our annotation found.
Combined with the original corpora, which remain publicly downloadable from their existing sources, the locations alone would show where each Social Security number appears, so we report the annotation only through the aggregate statistics of Section~\ref{sec:deid}.
Our obligations conflict at this point, and we discuss the conflict in the Broader Impact Statement.

\subsection{Provenance and Licensing}

All artifacts are keyed to the original RVL-CDIP file identifiers, so that they can be joined against the corpus as distributed.
Annotations are released in plain-text tabular formats, and each is versioned, since several supersede figures reported in our earlier papers and readers will need to know which version a given number came from.

\section{Datasets Used by the Surveyed Papers}\label{app:survey-datasets}

Table~\ref{tab:survey-datasets} lists, for each of the 34 papers of Table~\ref{tab:model-scores}, the datasets on which it evaluates document classification and other tasks and the document corpus on which it pre-trains; Section~\ref{sec:background-leaderboard} summarizes it.

\begin{table}[tbp]
  \centering
  \footnotesize
  \caption{Datasets used by the 34 papers of Table~\ref{tab:model-scores}, all
           of which evaluate document classification on RVL-CDIP.
           ``Other tasks'' counts datasets used for tasks other than
           classification (IE: information extraction, VQA: question answering,
           LA: layout analysis, TSR: table structure, OCR: text detection and
           recognition, NLI: natural language inference).
           $\dagger$: not publicly available.
           ``---'' in the last column: no document-corpus pre-training.}
  \label{tab:survey-datasets}
  \setlength{\tabcolsep}{3pt}
  \begin{tabular}{@{}lcp{0.95in}p{1.15in}p{0.9in}@{}}
  \toprule
  \textbf{Paper} & \textbf{Tobacco3482} & \textbf{Other classif.} &
  \textbf{Other tasks} & \textbf{Pre-training} \\
  \midrule
  \citet{csurka-2016}                     & --- & NIST-2, MARG, CLEF-IP, IH1--3$\dagger$ & ---                        & ---            \\
  \citet{tensmeyer-2017}                  & --- & ANDOC$\dagger$          & ---                                       & ---            \\
  \citet{afzal-2017}                      & $\bullet$ & ---               & ---                                       & RVL-CDIP       \\
  \citet{das-2018}                        & --- & ---                     & ---                                       & ---            \\
  \citet{laddernet-2019}                  & $\bullet$ & NIST-6, MARG      & ---                                       & ---            \\
  \citet{dauphinee-2019}                  & --- & ---                     & ---                                       & ---            \\
  \citet{audebert-2019}                   & $\bullet$ & ---               & ---                                       & ---            \\
  \citet{xu-2020-layoutlm}                & --- & ---                     & 2 (IE)                                    & IIT-CDIP       \\
  \citet{ferrando-2020}                   & $\bullet$ & ---               & ---                                       & ---            \\
  \citet{Zingaro2021}                     & $\bullet$ & ---               & ---                                       & ---            \\
  \citet{li-etal-2021-structurallm}       & --- & ---                     & 2 (IE, VQA)                               & IIT-CDIP       \\
  \citet{selfdoc-2021}                    & --- & ---                     & 1 (IE)                                    & RVL-CDIP       \\
  \citet{xu-etal-2021-layoutlmv2}         & --- & ---                     & 5 (IE, VQA)                               & IIT-CDIP       \\
  \citet{Appalaraju_2021_ICCV-docformer}  & --- & ---                     & 3 (IE)                                    & IIT-CDIP       \\
  \citet{TILT-2021}                       & --- & ---                     & 3 (IE, VQA)                               & RVL-CDIP, IDL  \\
  \citet{eaml-bakkali-2021}               & $\bullet$ & ---               & ---                                       & ---            \\
  \citet{udoc-2021}                       & --- & ---                     & 3 (IE, LA)                                & IIT-CDIP       \\
  \citet{layoutlmv3-huang-2022}           & --- & ---                     & 4 (IE, VQA, LA)                           & IIT-CDIP       \\
  \citet{li-2022-dit-image-transformer}   & --- & ---                     & 3 (LA, TSR, OCR)                          & IIT-CDIP       \\
  \citet{wang-etal-2022-lilt}             & --- & ---                     & 3 (IE)                                    & IIT-CDIP       \\
  \citet{mgdoc-2022-emnlp}                & --- & ---                     & 2 (IE)                                    & RVL-CDIP       \\
  \citet{kim2022donut}                    & --- & ---                     & 4 (IE, VQA)                               & IIT-CDIP       \\
  \citet{dessurt-2022}                    & --- & ---                     & 6 (IE, VQA, OCR)                          & IIT-CDIP       \\
  \citet{ernie-layout-2022}               & --- & ---                     & 5 (IE, VQA)                               & IDL crawl      \\
  \citet{pham-2022-long}                  & --- & ---                     & 2 (IE)                                    & IIT-CDIP       \\
  \citet{matrix-2022}                     & --- & ---                     & 2 (IE)                                    & IIT-CDIP       \\
  \citet{vlcdoc}                          & $\bullet$ & NIST-6            & ---                                       & ---            \\
  \citet{Saifullah2024-docxclassifier}    & $\bullet$ & ---               & ---                                       & RVL-CDIP       \\
  \citet{pramanik-2022}                   & --- & arXiv$\dagger$          & 2 (IE, TSR)                               & arXiv$\dagger$ \\
  \citet{bi-vldoc-2025-luo}               & --- & ---                     & 3 (IE, VQA)                               & IIT-CDIP       \\
  \citet{wukong-reader-2022}              & --- & ---                     & 4 (IE, VQA)                               & IIT-CDIP       \\
  \citet{yu2023structextv}                & --- & ---                     & 4 (IE, LA, TSR, OCR)                      & IIT-CDIP       \\
  \citet{tang-2023-udop}                  & --- & ---                     & 13 (IE, VQA, LA, NLI)                     & IIT-CDIP       \\
  \citet{fujitake-2024-layoutllm}         & --- & ---                     & 3 (IE, VQA)                               & inherited      \\
  \bottomrule
  \end{tabular}
\end{table}

\section{Sensitive Entity Types and Automated Detection}\label{app:pii}

Table~\ref{tab:pii-definitions} describes what each entity type of Table~\ref{tab:pii-entities} covers.
Besides the eleven types, the annotators marked signatures, a residual \emph{other} category, and in a few regions person names, which were not annotated systematically (Appendix~\ref{app:redact-values}).

\begin{table}[tbp]
  \centering
  \footnotesize
  \caption{The sensitive entity types recorded during annotation.}
  \label{tab:pii-definitions}
  \begin{tabular}{@{}lp{0.62\linewidth}@{}}
    \toprule
    \textbf{Entity type} & \textbf{Covers} \\
    \midrule
    US Social Security number & A full Social Security number, or a part of one such as the last four digits. \\
    Date of birth & A person's date of birth, in any format. \\
    Age & A person's age, stated as a number. \\
    Place of birth & The city, state or country in which a person was born. \\
    Marital/parental/spousal status & A person's marital status and the number of their children, which are often in separate fields. \\
    Home address & A residential address, in full or in its parts (street, city, state, ZIP code). The address of an employer or other organization is not included. \\
    Home telephone number & A residential telephone number. Business numbers are not included. \\
    Citizenship/nationality & A person's citizenship or nationality. \\
    Religious affiliation & A person's religion or denomination. \\
    Sex/gender & A person's sex or gender, as entered in a field. \\
    Health status & A statement about a person's health. \\
    Other & Sensitive content outside these types, such as the photograph on a scanned driver's license. \\
    \bottomrule
  \end{tabular}
\end{table}

We evaluated four PII detectors against the annotations: Presidio 2.2.33, Google DLP, the language service of Microsoft Azure, and Amazon Comprehend.
All four read text, so we gave them the Textract OCR of each page, and we measured recall at the document level on 1,773 annotated documents.
Table~\ref{tab:pii-tools} shows which entity types each detector supports and its recall on them.
Seven of the eleven types are supported by none of the detectors, and Google DLP is the only one that detects dates of birth and sex or gender.
All four find every annotated age, but recall on Social Security numbers ranges from 0.70 for Azure to 0.97 for Presidio.

\begin{table}[tbp]
  \centering
  \footnotesize
  \caption{Document-level recall of four PII detectors on 1,773 annotated
           documents, by entity type. A dash marks a type that the detector
           does not support.}
  \label{tab:pii-tools}
  \begin{tabular}{@{}lcccc@{}}
    \toprule
    \textbf{Entity type} & \textbf{Presidio} & \textbf{Google DLP} & \textbf{Azure} & \textbf{Amazon} \\
    \midrule
    US Social Security number       & 0.97 & 0.93 & 0.70 & 0.77 \\
    Date of birth                   & --   & 0.82 & --   & --   \\
    Age                             & 1.00 & 1.00 & 1.00 & 1.00 \\
    Sex/gender                      & --   & 1.00 & --   & --   \\
    Place of birth                  & --   & --   & --   & --   \\
    Marital/parental/spousal status & --   & --   & --   & --   \\
    Home address                    & --   & --   & --   & --   \\
    Home telephone number           & --   & --   & --   & --   \\
    Citizenship/nationality         & --   & --   & --   & --   \\
    Religious affiliation           & --   & --   & --   & --   \\
    Health status                   & --   & --   & --   & --   \\
    \bottomrule
  \end{tabular}
\end{table}

\begin{figure}[tbp]
  \centering
  \includegraphics[width=0.7\linewidth]{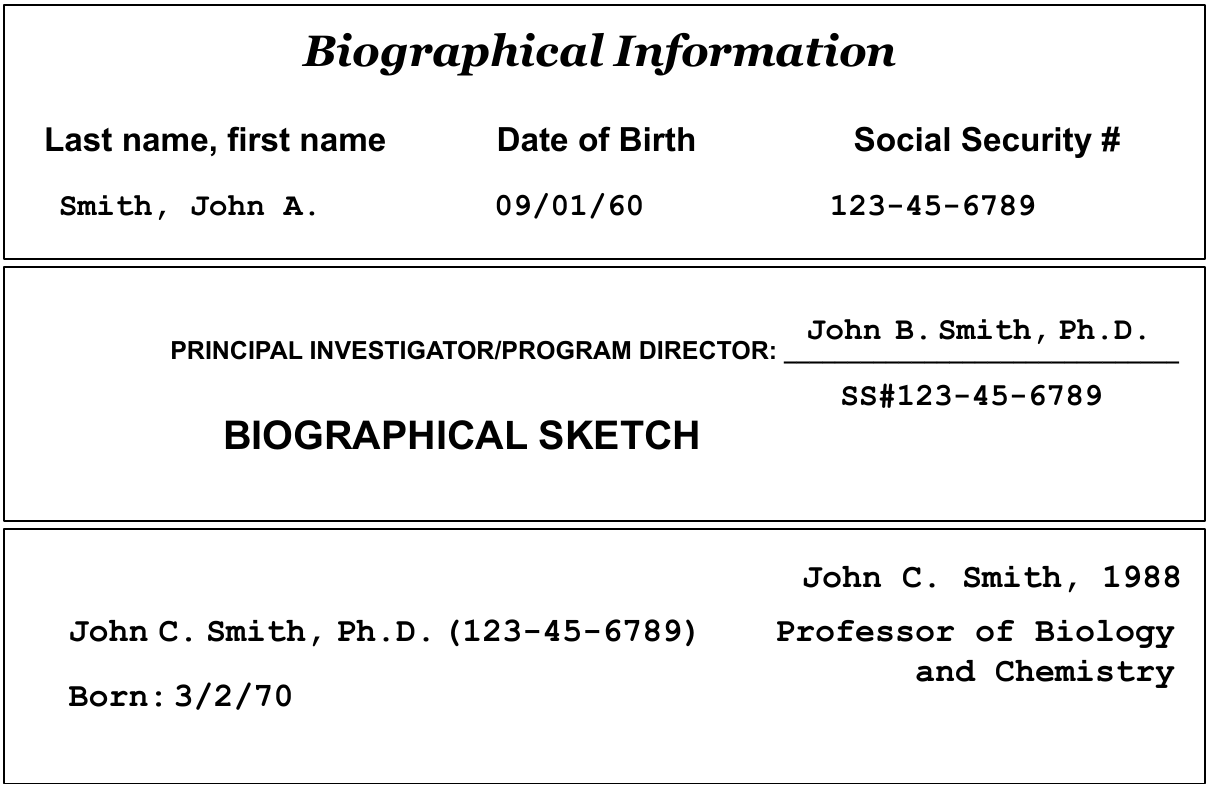}
  \caption{Reconstructed examples, with invented values, of the layouts in
           which detectors missed Social Security numbers. In the top example
           the label is in a column heading above the number, and in the
           bottom example the number has no label.}
  \label{fig:pii-tool-misses}
\end{figure}

\begin{figure}[tbp]
  \centering
  \includegraphics[width=0.6\linewidth]{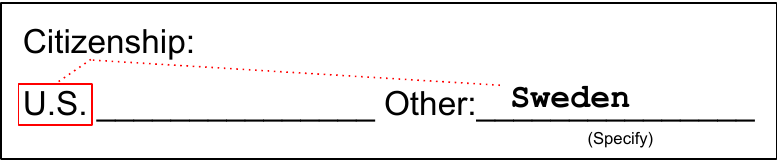}
  \caption{A reconstructed citizenship field. The text next to the label is
           the printed option \emph{U.S.}, and the entered value is
           \emph{Sweden}, after \emph{Other}.}
  \label{fig:pii-redact-failure}
\end{figure}

The missed Social Security numbers follow from how the detectors use context (Figure~\ref{fig:pii-tool-misses}).
A detector that looks for a keyword such as \emph{SSN} near a nine-digit number misses a number whose label is in a column heading far above it, and misses a number that is written without any label, as in a biographical sketch that gives it in parentheses after the name.
Placing a redaction by its position relative to a field label is also unreliable (Figure~\ref{fig:pii-redact-failure}): on a form that prints the options of a field, the text next to the label is a printed option, and the value that the person entered is elsewhere on the line.

\section{Sensitive Data in Datasets Derived from IIT-CDIP}\label{app:pii-counts}

Section~\ref{sec:deid} reports the RVL-CDIP counts. Table~\ref{tab:pii-counts-full}
shows them for all five datasets derived from IIT-CDIP, since a sensitive entity on
an RVL-CDIP page recurs in every other dataset built from the same source page.

\begin{table}[tbp]
  \centering
  \caption{Sensitive entity counts across the five IIT-CDIP-derived datasets.
           ``Documents affected'' counts documents containing at least one
           sensitive entity, so it is not the column sum.}
  \label{tab:pii-counts-full}
  \scalebox{0.82}{
  \begin{tabular}{lrrrrr|r}
    \toprule
    & \textbf{RVL-CDIP} & \textbf{Tob.3482} & \textbf{Tob.800} & \textbf{FUNSD} & \textbf{DocVQA} & \textbf{Total} \\
    \midrule
    Dataset size            & 400,000 & 3,482 & 1,290 & 199 & 12,767 & 417,738 \\
    \midrule
    US Social Security no.  & 2,342  & 9  & 5 & 2 & 70  & 2,428 \\
    Date of birth           & 12,800 & 62 & 7 & --- & 232 & 13,101 \\
    Place of birth          & 6,125  & 29 & 7 & --- & 123 & 6,284 \\
    Age                     & 219    & --- & --- & --- & 44  & 263 \\
    Home address            & 3,908  & 6  & 1 & 1 & 276 & 4,192 \\
    Home telephone          & 1,801  & 4  & --- & --- & 143 & 1,948 \\
    Marital/parental status & 4,228  & 18 & 7 & --- & 116 & 4,369 \\
    Citizenship/nationality & 2,647  & 19 & --- & --- & 49  & 2,715 \\
    Sex/gender              & 602    & 3  & --- & --- & 88  & 693 \\
    Health status           & 60     & --- & --- & --- & 2   & 62 \\
    Religious affiliation   & 43     & --- & --- & --- & 4   & 47 \\
    \midrule
    Documents affected      & 15,956 & 66 & 12 & 2 & 360 & 16,396 \\
    \bottomrule
  \end{tabular}}
\end{table}

\section{The De-Identification Pipeline}\label{app:redact}

The pipeline of Section~\ref{sec:deid} is released as \texttt{doc-redact}.\footnote{\url{https://github.com/gxlarson/doc-redact}}
In this appendix, we describe how the pipeline generates the replacement values and chooses the typeface, and how we checked the annotations before any page was rendered.
The encoding of the output pages is described in Appendix~\ref{app:redact-encoding}.

\subsection{Replacement Values}\label{app:redact-values}

The pipeline has a value generator for fifteen annotation categories, listed in Table~\ref{tab:redact-generators}.
Four of the categories are the parts of a home address (street, city, state and ZIP code), for annotations that mark them separately, and two, person names and other identification numbers, are not among the eleven entity types of Table~\ref{tab:pii-entities}; names were not annotated systematically, and the generator covers the few regions in which one was marked.
Sex and gender, health status, signatures and the residual \emph{other} category have no generator, and their regions are erased and refilled but not written back (Section~\ref{sec:deid}).

\begin{table}[tbp]
  \centering
  \footnotesize
  \caption{Value generators of the de-identification pipeline. Each replacement is written in the format of the original where the format can be detected, and in a format chosen at random otherwise.}
  \label{tab:redact-generators}
  \begin{tabular}{@{}lp{0.62\linewidth}@{}}
    \toprule
    \textbf{Category} & \textbf{Replacement} \\
    \midrule
    Social Security number & Nine digits in the form the Social Security Administration issues: area 001 to 899 other than 666, group 01 to 99, serial 0001 to 9999. A partial number, such as the last four digits, is filled digit by digit in the layout of the original. \\
    Date of birth & A date between 1930 and 1970, in the original's format (numeric or with the month written out, its separators, and a two- or four-digit year). \\
    Age & A number between 29 and 59. \\
    Place of birth & A real US city, with the state written as in the original (postal code, traditional abbreviation, full name, or omitted). \\
    Marital/parental/spousal status & One status (married, divorced, widowed or single) and one number of children (none to four) per document, written as in the original (a word, an initial, a coded entry such as ``c.~2'', or a count of children). \\
    Home address & A street line in the usual US form, named after a tree, a president, a surname, a US place or a numbered street, with a real US city and a ZIP code from the range of the state it is written beside. \\
    Street, city, state, ZIP & As for a home address, for annotations that mark the parts separately. \\
    Home telephone number & Ten digits that follow the North American Numbering Plan, so that the area code and the exchange both begin with 2 to 9, in the original's layout. \\
    Citizenship/nationality & One of 25 forms, from \emph{American} and \emph{U.S.A.} to \emph{Canadian} and \emph{Japanese}. \\
    Religious affiliation & One of seven affiliations, such as \emph{Catholic} or \emph{Jewish}. \\
    Other identification number & Seven to twelve digits, in the layout of the original where it consists of digits and separators. \\
    Person name & A given name and a surname, with any title of the original kept and the given name chosen to agree with it. \\
    \bottomrule
  \end{tabular}
\end{table}

Replacements are consistent within a page.
When the same value appears in more than one region, such as a Social Security number entered on two forms or a date of birth written both as 4/12/50 and as April 12, 1950, all of its regions receive one replacement, each written in its own format.
Addresses are linked by their parts, so that a street, a city and a ZIP code that recur on a page, whether in full addresses or in separately annotated parts, are each replaced in the same way everywhere they occur.
The marital status and the number of children are linked in the same way, because they are often annotated as separate regions.
The original values are compared only in memory and are not stored.

\subsection{Typeface Matching}\label{app:redact-typeface}

\begin{figure}[tbp]
  \centering
  \includegraphics[width=\linewidth]{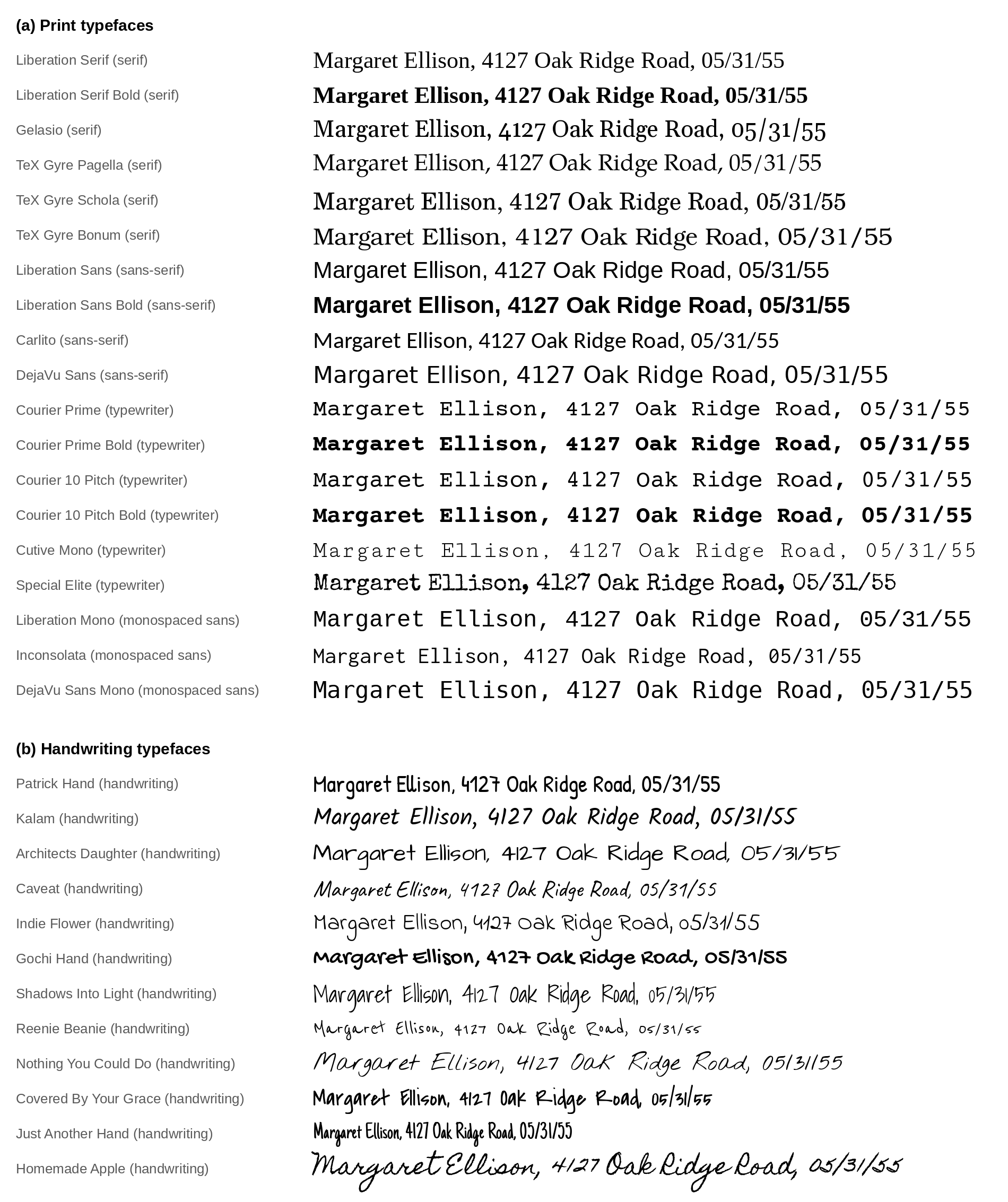}
  \caption{The typefaces bundled with \texttt{doc-redact}: (a) print, (b) handwriting, drawn here without the per-letter changes.}
  \label{fig:font-samples}
\end{figure}

The nineteen print typefaces, shown in Figure~\ref{fig:font-samples}, are of four kinds:
serif (Liberation Serif in regular and bold, Gelasio, and TeX Gyre Pagella, Schola and Bonum),
sans-serif (Liberation Sans in regular and bold, Carlito, and DejaVu Sans),
typewriter faces with serifs (Courier Prime and Courier 10 Pitch, each in regular and bold, Cutive Mono, and Special Elite),
and monospaced sans-serif (Liberation Mono, Inconsolata, and DejaVu Sans Mono).
All are distributed under licenses that allow redistribution, and they are included with the \texttt{doc-redact} code.
Regions whose words Textract labels as handwriting are matched against twelve handwriting typefaces instead, and the letters of a handwritten replacement are drawn one at a time, each with a small random change of size, height, spacing and rotation, so that a repeated letter is not drawn identically twice.

Only words of at least four letters and at least 10 pixels in height are compared, since shorter words do not distinguish between typefaces.
The words inside the region are tried first, longest first, and if none of them qualifies, the words on the same line nearest to the region are used; this is the case for a Social Security number or a telephone number, which contains no letters.
Each candidate typeface renders a word at the size at which its width matches the scan, and at 3\% smaller and larger, and the scan and the rendering are compared by chamfer distance over shifts of up to 3 pixels.
Each word ranks the typefaces, and the typeface with the lowest total rank over at most three words is chosen.
If no word can be measured, the replacement is drawn in Inconsolata, or in Patrick Hand for handwriting.
The matcher's errors are most likely where a region and its line offer only one or two measurable words, since the vote then rests on few letters, and it can then choose a serif face for text that is set in a bold sans-serif.

\subsection{Checking Annotations Before Rendering}\label{app:redact-checks}

Before rendering, the pipeline compares the OCR text inside each annotated region with the region's category, and flags the regions where the two disagree.
A region is flagged if it contains no OCR words; if its text reads as a date and its category is not a date of birth, a home address or a street; if its text has the form of a Social Security number and its category is not that; if its category consists of digits (Social Security number, telephone number, age, ZIP code or date of birth) and its text has none; or if its category is a word (place of birth, marital status, religion, nationality, city or state) and its text has only digits and punctuation.
We used these flags to find annotation errors before the pages were rendered.
A further flag records that the format of the original could not be detected, in which case the generator chooses a format at random.

\subsection{Encoding and Use of OCR}\label{app:redact-encoding}

We describe here how the pages of the released corpus are encoded and how the replacement pipeline uses OCR.
RVL-CDIP is distributed as TIFF files, which we converted to JPEG at quality 75, and the pipeline reads these copies and writes the de-identified pages as JPEG at quality 95.
Re-encoding changes the whole page and not only the replaced regions: measured against the quality-75 copies outside all annotated regions, the mean change is 0.31 gray levels, against 0.003 for a lossless encode.
The same encoding is applied to all pages and therefore to all conditions in Section~\ref{sec:joint}, so the comparisons between those conditions are made on equally encoded pages.
Our figures are therefore not bitwise comparable with previously published RVL-CDIP results, although the comparison of original and de-identified pages in Section~\ref{sec:deid} indicates that the difference is smaller than the precision at which those results are reported, and our measurements are made on the files that we distribute.

The words, and the word boxes from which the pipeline measures size, baseline and typeface, are obtained from Amazon Textract.
Since Section~\ref{sec:axis5} compares Textract against Tesseract as engines, we note that these are two different uses of it: Textract located the words a replacement had to imitate, on the 4\% of pages that contain an annotation; it did not produce any text on which a model is evaluated, and both engines are run over the replaced pages on equal terms.

\section{Confident Learning on RVL-CDIP}\label{app:cleanlab}

Confident learning \citep{northcutt2021confidentlearning} estimates which labels of a dataset are wrong from the out-of-sample class probabilities of a model trained on those labels.
We apply it to the RVL-CDIP test split, for which the review of Section~\ref{sec:axis1} gives the status of every label: 4,719 of the 40,000 test pages (11.8\%) have a label error (4,317) or an unsure label (402), and in this appendix we count all of them as wrong labels; the review assigned a corrected category to 1,166 of them.
The probabilities are those of the eleven fine-tuned models of Section~\ref{sec:joint-models}, trained on the original labels and on pages with identification codes with three seeds each, and they are out-of-sample for the test pages because the models were trained on the training split.
We use the \texttt{cleanlab} package (version 2.9.0) with its default settings, and we count a flagged page as correct when the review found its label wrong.

\begin{table}[tbp]
  \centering\footnotesize
  \caption{Confident learning on the RVL-CDIP test split, scored against the
           label review, which finds a label error or an unsure label on 11.8\% of
           the 40,000 test pages.
           \emph{Estimate} is the share of wrong labels estimated by the
           method and \emph{Flagged} the share of pages that it flags;
           precision and recall are those of the flagged pages; \emph{P at
           4,719} is the precision among the 4,719 pages with the lowest
           label-quality score. All values are in percent and are means over
           three seeds.}
  \label{tab:cleanlab}
  \begin{tabular}{lcccccc}
    \toprule
    \textbf{Model} & \textbf{Test acc.} & \textbf{Estimate} & \textbf{Flagged} & \textbf{Precision} & \textbf{Recall} & \textbf{P at 4,719} \\
    \midrule
    \multicolumn{7}{l}{\emph{Image}} \\
    AlexNet          & 89.3 & 4.5 & 5.3 & 39.7 & 17.8 & 37.2 \\
    GoogLeNet        & 89.2 & 5.9 & 6.6 & 40.4 & 22.6 & 38.1 \\
    SqueezeNet       & 88.7 & 3.4 & 4.3 & 44.9 & 16.4 & 39.9 \\
    VGG-16           & 91.3 & 6.0 & 6.3 & 42.6 & 22.8 & 38.6 \\
    ResNet-50        & 90.9 & 5.2 & 5.7 & 44.3 & 21.5 & 39.6 \\
    ResNeXt-50       & 91.3 & 4.9 & 5.3 & 44.8 & 20.3 & 39.5 \\
    DiT-base         & 92.8 & 3.6 & 4.0 & 47.1 & 15.8 & 40.0 \\
    \midrule
    \multicolumn{7}{l}{\emph{Text}} \\
    BERT-base        & 92.3 & 5.3 & 5.6 & 47.6 & 22.5 & 35.7 \\
    RoBERTa-base     & 92.5 & 4.8 & 5.1 & 51.2 & 22.1 & 38.5 \\
    \midrule
    \multicolumn{7}{l}{\emph{Multi-modal}} \\
    LayoutLM-base    & 93.4 & 4.5 & 4.7 & 52.1 & 20.6 & 34.5 \\
    LayoutLMv3-base  & 94.8 & 2.0 & 2.2 & 57.1 & 10.5 & 39.1 \\
    \bottomrule
  \end{tabular}
\end{table}

Table~\ref{tab:cleanlab} shows, for each of the eleven models, the share of test labels that confident learning estimates to be wrong, the share of pages that it flags, and the precision and recall of the flagged pages against our review.
Depending on the model, confident learning estimates that between 2.0 and 6.0\% of the test labels are wrong, where the review finds 11.8\%.
It flags 2.2 to 6.6\% of the pages, 39.7 to 57.1\% of the flagged pages have a wrong label, and the flagged pages contain 10.5 to 22.8\% of the wrong labels.
LayoutLMv3, the most accurate of the eleven models, yields the smallest estimate (2.0\%) and finds the smallest share of the wrong labels (10.5\%).
The three other confident-learning filters of the package find 8.6 to 25.4\% of the wrong labels, and the rule that flags every page whose predicted category differs from its label finds 23.5 to 38.4\% at a precision of 38.3 to 53.3\%.
When the pages are ranked by the label-quality score of the package and as many pages are inspected as there are wrong labels, 34.5 to 40.0\% of the inspected pages have a wrong label.
The wrong labels that the review could correct are found about as often as those of the pages without a corrected category (11.1 to 26.1\% against 10.3 to 22.3\%), and for a flagged page of the first kind the prediction of the model is the corrected label in 66.6 to 89.0\% of the cases.

\begin{table}[tbp]
  \centering\footnotesize
  \caption{Confident learning by original category on the RVL-CDIP test
           split: number of test pages, share of wrong labels according to
           the review, and the share of pages flagged, with the precision
           and recall of the flags. Values are in percent and are means over
           the eleven models and three seeds of
           Table~\ref{tab:cleanlab}.}
  \label{tab:cleanlab-category}
  \begin{tabular}{lrcccc}
    \toprule
    \textbf{Category} & $n$ & \textbf{Wrong labels} & \textbf{Flagged} & \textbf{Precision} & \textbf{Recall} \\
    \midrule
    \cat{letter} & 2,464 & 30.7 & 5.6 & 68.5 & 12.3 \\
    \cat{budget} & 2,505 & 21.3 & 5.2 & 63.5 & 14.9 \\
    \cat{form} & 2,506 & 16.5 & 10.6 & 47.7 & 30.1 \\
    \cat{questionnaire} & 2,435 & 15.2 & 5.9 & 55.8 & 20.1 \\
    \cat{invoice} & 2,477 & 14.8 & 4.6 & 55.6 & 16.6 \\
    \cat{scientific\_report} & 2,498 & 13.2 & 10.3 & 31.9 & 23.8 \\
    \cat{handwritten} & 2,532 & 12.7 & 3.4 & 70.3 & 16.4 \\
    \cat{email} & 2,516 & 11.1 & 0.8 & 68.9 & 5.1 \\
    \cat{advertisement} & 2,515 & 10.9 & 4.8 & 41.1 & 16.4 \\
    \cat{news\_article} & 2,463 & 9.5 & 5.6 & 51.9 & 30.3 \\
    \cat{scientific\_publication} & 2,571 & 8.3 & 4.0 & 45.1 & 21.6 \\
    \cat{memo} & 2,492 & 8.1 & 3.6 & 49.6 & 21.5 \\
    \cat{presentation} & 2,489 & 6.3 & 9.0 & 29.3 & 40.2 \\
    \cat{file\_folder} & 2,527 & 5.7 & 1.9 & 55.0 & 16.6 \\
    \cat{specification} & 2,472 & 3.4 & 3.5 & 37.2 & 36.0 \\
    \cat{resume} & 2,537 & 1.5 & 1.6 & 39.1 & 32.0 \\
    \bottomrule
  \end{tabular}
\end{table}

Table~\ref{tab:cleanlab-category} shows the same quantities for each category.
In \cat{letter}, the category with the most wrong labels, the review finds 30.7\% of the test labels wrong, and the method flags 5.6\% of the pages, which contain 12.3\% of the wrong labels; in \cat{budget} the three values are 21.3, 5.2 and 14.9\%, and \cat{email} has the lowest recall of the sixteen categories (5.1\%).
The same procedure applied to the models trained on the corrected labels, with the corrected test labels, flags 0.9 to 3.8\% of the 36,446 pages of the corrected test split.

Confident learning therefore underestimates the label error rate of the test split by a factor of two to six, and with its default settings it finds less than a quarter of the wrong labels that the review found.
We did not apply it to the training split, which would require probabilities from cross-validation.

\section{Original and Corrected Labels}\label{app:label-confusion}

Table~\ref{tab:label-confusion} cross-tabulates the original and corrected labels of the 15,366 documents across all three splits that the review found \emph{wrong} and relabeled (Section~\ref{sec:axis1}); documents found \emph{unknown}, which have no corrected label, are not included.

\begin{table}[tbp]
  \centering
  \caption{Original label (rows) against corrected label (columns) for the 15,366
           relabeled documents of RVL-CDIP, all splits. Empty cells are zero.
           adv = \cat{advertisement}, bud = \cat{budget}, ema = \cat{email},
           fol = \cat{file\_folder}, frm = \cat{form}, hw = \cat{handwritten},
           inv = \cat{invoice}, let = \cat{letter}, mem = \cat{memo},
           news = \cat{news\_article}, pres = \cat{presentation},
           qst = \cat{questionnaire}, res = \cat{resume},
           pub = \cat{scientific\_publication}, rep = \cat{scientific\_report},
           spec = \cat{specification}.}
  \label{tab:label-confusion}
  \footnotesize
  \setlength{\tabcolsep}{2.5pt}
  \begin{tabular}{l*{16}{r}r}
  \toprule
 & \textbf{adv} & \textbf{bud} & \textbf{ema} & \textbf{fol} & \textbf{frm} & \textbf{hw} & \textbf{inv} & \textbf{let} & \textbf{mem} & \textbf{news} & \textbf{pres} & \textbf{qst} & \textbf{res} & \textbf{pub} & \textbf{rep} & \textbf{spec} & \textbf{Total} \\
  \midrule
\texttt{adv} & -- & 1 & 1 & 38 & 22 & 16 &  & 14 & 17 & 39 & 1 &  &  &  & 1 & 1 & 151 \\
\texttt{bud} &  & -- &  & 156 & 53 & 370 & 1,499 & 10 & 16 & 6 & 7 &  &  &  &  & 3 & 2,120 \\
\texttt{ema} & 2 & 15 & -- & 3 & 7 & 2 &  & 52 & 138 & 32 & 49 & 1 & 2 & 15 &  &  & 318 \\
\texttt{fol} & 2 & 1 &  & -- & 12 & 10 & 1 & 2 & 2 & 1 & 4 &  &  & 3 & 1 & 1 & 40 \\
\texttt{frm} &  & 62 & 6 & 100 & -- & 74 & 188 & 82 & 90 &  & 1 & 21 & 6 &  & 20 & 255 & 905 \\
\texttt{hw} & 39 & 2 & 1 & 45 & 78 & -- & 1 & 234 & 19 & 6 &  & 106 &  & 3 &  & 2 & 536 \\
\texttt{inv} & 1 & 1,846 &  & 54 & 97 & 34 & -- & 82 & 12 & 3 &  &  &  &  &  &  & 2,129 \\
\texttt{let} & 3 & 3 & 175 & 44 & 81 & 272 & 15 & -- & 3,253 & 11 & 7 & 1 & 3 & 7 & 1 & 2 & 3,878 \\
\texttt{mem} & 1 &  & 11 & 48 & 77 & 176 & 3 & 92 & -- & 1 & 2 &  &  &  & 3 & 17 & 431 \\
\texttt{news} & 40 &  &  & 56 & 23 & 3 &  & 1 & 4 & -- & 455 &  &  & 544 &  &  & 1,126 \\
\texttt{pres} & 6 & 3 &  & 162 & 18 & 100 &  & 14 & 3 & 19 & -- &  &  & 8 & 2 &  & 335 \\
\texttt{qst} & 11 &  & 2 & 122 & 118 & 205 &  & 972 & 48 & 3 & 5 & -- & 3 & 1 & 1 & 3 & 1,494 \\
\texttt{res} &  &  &  &  &  & 1 &  &  &  &  &  &  & -- &  & 1 &  & 2 \\
\texttt{pub} &  &  &  & 38 &  & 2 &  & 1 & 1 & 548 &  &  &  & -- & 3 & 4 & 597 \\
\texttt{rep} & 1 & 28 &  & 34 & 257 & 449 &  & 8 & 6 & 1 & 3 &  & 2 & 203 & -- & 30 & 1,022 \\
\texttt{spec} &  &  & 1 & 37 & 104 & 58 &  & 9 & 65 &  & 5 &  &  &  & 3 & -- & 282 \\
  \bottomrule
  \end{tabular}
\end{table}

\section{Identification Codes: Leading-Digit Distribution}\label{app:idcodes}

\begin{figure}[tbp]
  \centering
  \includegraphics[width=\linewidth]{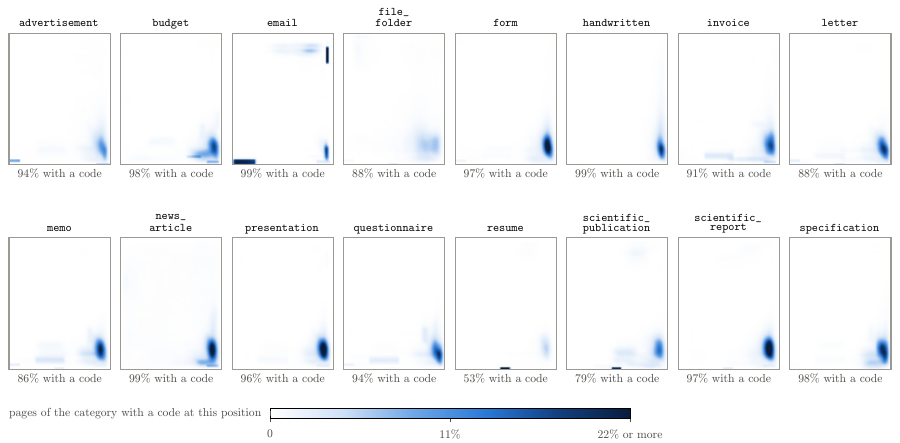}
  \caption{Position of the identification codes detected on all 400,000
           RVL-CDIP pages, by original category, as the share of the
           category's pages with a code at each position; the number under each
           panel is the share of its pages with at least one detected code.}
  \label{fig:distributions-all}
\end{figure}

Table~\ref{tab:starting-digit-full} shows the leading-character distribution for
all sixteen categories, which Section~\ref{sec:axis3} discusses.
Figure~\ref{fig:distributions-all} shows the position of the detected codes on the page for all sixteen categories.

\begin{table}[tbp]
  \centering
  \caption{Frequency of the leading character of identification codes, by
           category, for the digits 0--9. Percentages within a row; rows do not
           sum to 100 because leading characters other than digits are omitted.}
  \label{tab:starting-digit-full}
  \small
  \setlength{\tabcolsep}{4pt}
  \begin{tabular}{lcccccccccc}
  \toprule
  \textbf{Category} & \textbf{0} & \textbf{1} & \textbf{2} & \textbf{3} & \textbf{4} & \textbf{5} & \textbf{6} & \textbf{7} & \textbf{8} & \textbf{9} \\
  \midrule
  \cat{advertisement}          & \gradient{3.5}  & \gradient{4.2}  & \gradient{34.4} & \gradient{2.7} & \gradient{1.4}  & \gradient{30.4} & \gradient{8.1}  & \gradient{2.0}  & \gradient{3.8}  & \gradient{5.4} \\
  \cat{budget}                 & \gradient{1.5}  & \gradient{0.7}  & \gradient{37.6} & \gradient{0}   & \gradient{0}    & \gradient{2.2}  & \gradient{4.5}  & \gradient{0.7}  & \gradient{7.0}  & \gradient{9.1} \\
  \cat{email}                  & \gradient{0}    & \gradient{0.2}  & \gradient{54.4} & \gradient{0}   & \gradient{0.4}  & \gradient{16.4} & \gradient{0.2}  & \gradient{14.0} & \gradient{0.2}  & \gradient{0.2} \\
  \cat{file\_folder}           & \gradient{5.8}  & \gradient{0.6}  & \gradient{76.4} & \gradient{0}   & \gradient{1.1}  & \gradient{2.3}  & \gradient{1.3}  & \gradient{1.9}  & \gradient{5.9}  & \gradient{1.7} \\
  \cat{form}                   & \gradient{2.4}  & \gradient{0.7}  & \gradient{64.4} & \gradient{0.1} & \gradient{1.0}  & \gradient{18.0} & \gradient{2.6}  & \gradient{0.4}  & \gradient{4.1}  & \gradient{5.6} \\
  \cat{handwritten}            & \gradient{0}    & \gradient{0}    & \gradient{0}    & \gradient{0}   & \gradient{0.1}  & \gradient{99.9} & \gradient{0}    & \gradient{0}    & \gradient{0}    & \gradient{0} \\
  \cat{invoice}                & \gradient{4.1}  & \gradient{3.5}  & \gradient{38.4} & \gradient{0.3} & \gradient{0}    & \gradient{8.0}  & \gradient{4.6}  & \gradient{1.9}  & \gradient{17.5} & \gradient{11.6} \\
  \cat{letter}                 & \gradient{2.7}  & \gradient{3.6}  & \gradient{24.3} & \gradient{0.2} & \gradient{0.6}  & \gradient{44.1} & \gradient{8.2}  & \gradient{1.5}  & \gradient{4.2}  & \gradient{4.2} \\
  \cat{memo}                   & \gradient{3.0}  & \gradient{8.1}  & \gradient{40.2} & \gradient{1.1} & \gradient{1.1}  & \gradient{4.8}  & \gradient{14.5} & \gradient{2.7}  & \gradient{9.2}  & \gradient{6.7} \\
  \cat{news\_article}          & \gradient{6.5}  & \gradient{17.9} & \gradient{51.7} & \gradient{0.5} & \gradient{0}    & \gradient{0.3}  & \gradient{2.7}  & \gradient{0}    & \gradient{2.1}  & \gradient{1.5} \\
  \cat{presentation}           & \gradient{1.4}  & \gradient{2.0}  & \gradient{68.7} & \gradient{0.1} & \gradient{0.7}  & \gradient{1.7}  & \gradient{6.2}  & \gradient{0.7}  & \gradient{7.2}  & \gradient{6.2} \\
  \cat{questionnaire}          & \gradient{6.5}  & \gradient{7.2}  & \gradient{32.8} & \gradient{0.1} & \gradient{3.2}  & \gradient{29.0} & \gradient{8.4}  & \gradient{2.1}  & \gradient{7.1}  & \gradient{1.6} \\
  \cat{resume}                 & \gradient{4.0}  & \gradient{3.7}  & \gradient{19.1} & \gradient{0}   & \gradient{12.5} & \gradient{51.3} & \gradient{1.6}  & \gradient{0.5}  & \gradient{4.0}  & \gradient{0.5} \\
  \cat{scientific\_publication} & \gradient{12.7} & \gradient{8.9} & \gradient{35.3} & \gradient{0}   & \gradient{1.6}  & \gradient{9.2}  & \gradient{1.2}  & \gradient{0.3}  & \gradient{3.6}  & \gradient{3.1} \\
  \cat{scientific\_report}     & \gradient{5.9}  & \gradient{8.1}  & \gradient{62.4} & \gradient{0}   & \gradient{0.3}  & \gradient{3.7}  & \gradient{1.4}  & \gradient{0}    & \gradient{13.7} & \gradient{1.4} \\
  \cat{specification}          & \gradient{1.3}  & \gradient{0.6}  & \gradient{72.6} & \gradient{0.6} & \gradient{0}    & \gradient{18.2} & \gradient{5.0}  & \gradient{0.6}  & \gradient{0.7}  & \gradient{0} \\
  \midrule
  All                             & \gradient{3.7}  & \gradient{4.5}  & \gradient{45.6} & \gradient{0.4} & \gradient{1.1}  & \gradient{20.1} & \gradient{4.3}  & \gradient{2.1}  & \gradient{5.5}  & \gradient{3.6} \\
  \bottomrule
  \end{tabular}
\end{table}

\section{Identification Codes: Features of the Code-Only Classifier}\label{app:idcode-features}

The XGBoost classifier of Section~\ref{sec:axis3-signal} receives one feature vector per document, computed from the bounding boxes of the identification codes on the page and from the transcriptions of the text inside them, and Table~\ref{tab:idcode-features} lists the features by group with the number in each group for RVL-CDIP, 330 in total.
A page with several codes is described mainly through the first code that has a transcription, which we call the main code, and a page without a code keeps its row with the code features missing, so that the absence of a code is itself available to the classifier.
For RVL-CDIP the boxes are the regions that the detector of Section~\ref{sec:axis3-removal} removed, which exist on 91\% of pages (446,332 boxes on 364,168 pages), and each box was cropped with a margin of four pixels, turned upright, and transcribed with Amazon Textract, which returned text for 99.5\% of the boxes; for Tobacco3482 the boxes and the transcriptions are those of the annotation.

\begin{table}[tbp]
  \centering
  \caption{Features of the code-only classifier of
           Section~\ref{sec:axis3-signal}, by group, with the number of
           features in each group for RVL-CDIP (sixteen categories). The main
           code is the first code on the page that has a transcription.}
  \label{tab:idcode-features}
  \small
  \setlength{\tabcolsep}{5pt}
  \begin{tabular}{p{0.24\linewidth}p{0.62\linewidth}r}
  \toprule
  \textbf{Group} & \textbf{Features} & \textbf{N} \\
  \midrule
  Bounding box & position, height, width, area, and center of the main code's box, the area and the center both in pixels and as shares of the page, and whether the box is wider than it is tall & 11 \\
  All codes on the page & number of codes, the boxes and orientations of up to three codes, the box enclosing all of them, and the numbers of horizontal and vertical codes & 23 \\
  Length and characters & length of the main code in characters and in tokens, the numbers and shares of digits and of letters, and its first character, also as indicators for a digit and for a letter & 9 \\
  String structure & number of space-separated groups and the lengths of the first two, the leading run of digits as a number and its logarithm, the number of leading letters, and indicators for a trailing letter and for a hyphen & 8 \\
  Format and prefixes & for the pattern of digits and letters of the main code and for each of its first one to five characters, the distribution over categories of the training codes that share it and the logarithm of their number & 102 \\
  Nearest numbers & among training codes with the same pattern, the distribution over categories of the five nearest in numeric value, the logarithm of the distance to the nearest, and its category & 18 \\
  Second code & the string structure, format and prefix, and nearest-number features of the second transcribed code on the page, computed against all training codes & 128 \\
  OCR of the crop & for the main and the second code, the confidence of the engine, the rotation chosen for the crop and how it was determined, the confidence of the rejected rotation, and whether both rotations returned the same text; for the page, the number of transcribed codes and their mean and minimum confidence & 19 \\
  OCR of the page & for the main and the second code, whether OCR of the full page returned text inside the box, whether that text equals or is contained in the reading of the crop, their string similarity and length ratio; for the page, the number and share of codes that full-page OCR read & 12 \\
  \bottomrule
  \end{tabular}
\end{table}

The first three groups describe where the ID code is located and what kind of string it is without recording which digits or letters it contains beyond the first, and they depend on the equipment and conventions of the party that numbered the document and not on the content of the document.
The format, prefix, and nearest-number groups do record the content of the code, by replacing a pattern, a prefix, or a position in the numbering sequence with the categories observed for it among the training documents, which lets the classifier learn that a particular range of Bates numbers belongs largely to one category; these features depend on the training labels, so for a training document they are computed from the other four of five folds of the training set and for a test document from the whole training set, and no document's own label enters its features.
The two OCR groups describe how legible the ID code is, through the confidence of the engine on the crop and through whether the engine could read the code when given the whole page, which it did for 42\% of RVL-CDIP pages, and they therefore depend on the size and quality of the print and not on the content of the code.
Several features are redundant by construction, the ratios being functions of the counts and the normalized quantities of the raw ones, which is harmless for tree ensembles that split on one feature at a time.
For Tobacco3482 the same groups are computed from the annotated boxes and transcriptions, with distributions over ten categories in place of sixteen and without the rotation fields, which exist only for the RVL-CDIP crops.

The second configuration of Table~\ref{tab:idcode-classification} adds the class probabilities of a convolutional network that sees only the pixels of the code.
Each crop is scaled to a height of 32 pixels and placed on a white canvas of 32 by 256 pixels, and the network, which has five residual blocks of 32 to 256 channels followed by global average pooling, is trained from scratch for 20 epochs on the crops of the training documents with the category of the document as the target; a page with several codes receives the mean of their log-probabilities.
The probabilities given to the XGBoost classifier are out-of-fold for the training documents, from five networks each trained on four fifths of the training pages, and the mean of the five networks for the test documents.
For Tobacco3482 the network is trained in the same way within each of the ten splits, for 60 epochs on about 2,800 crops, and the network trained on RVL-CDIP is not reused because the two corpora share documents.
All XGBoost models use the default settings of the library.

\section{Identification Codes: Supplementary Results for the Code-Only Classifier}\label{app:idcode-analysis}

\begin{figure}[tbp]
  \centering
  \includegraphics[width=0.85\linewidth]{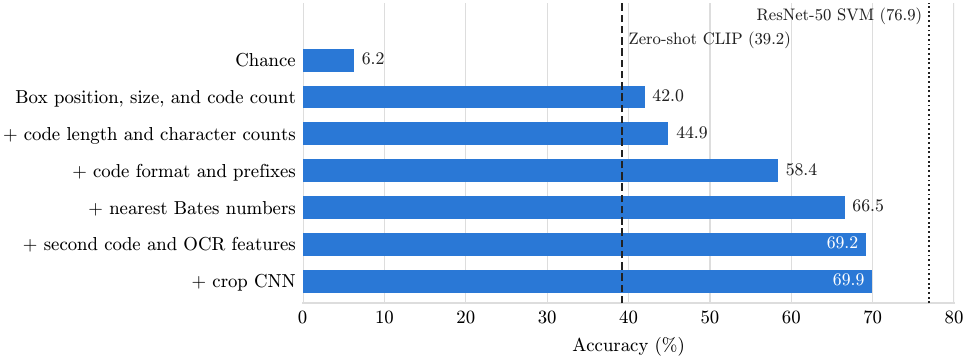}
  \caption{Accuracy of the code-only classifier on the RVL-CDIP test split as
           groups of features are added cumulatively (XGBoost, 320,000
           training documents). The dashed line marks zero-shot CLIP applied
           to the whole page \citep{siddiqui-zero-shot-2021}, and the dotted
           line a linear SVM trained on frozen ImageNet ResNet-50 features of
           the whole page (Table~\ref{tab:svm-baselines}).}
  \label{fig:idcode-feature-groups}
\end{figure}

Figure~\ref{fig:idcode-feature-groups} shows how the accuracy of the code-only classifier of Section~\ref{sec:axis3-signal} on RVL-CDIP is built up as the feature groups of Table~\ref{tab:idcode-features} are added.
The position and size of the boxes together with the number of codes on the page yield an accuracy of 42.0\% against a chance rate of 6.25\%, which is already above the 39.2\% that \citet{siddiqui-zero-shot-2021} report for zero-shot CLIP on the whole page, and adding the length of the code and its counts of digits, letters, and tokens raises this to 44.9\%.
The largest steps come from the content of the code: its format and its first one to five characters bring the accuracy to 58.4\%, and the categories of the training codes nearest in numeric value add a further 8.1 points for 66.5\%, so that the classifier gains more than twenty points from learning which party produced the document and in which batch it was numbered.
The features of the second code and the two OCR groups add 2.7 points for the 69.2\% of Table~\ref{tab:idcode-classification}; added one group at a time to the 66.5\% model, the OCR of the crop contributes 1.7 points, the OCR of the page 1.1, and the second code 0.5.
The convolutional network obtains 59.0\% from the pixels of the crops alone, without the position of the box or any transcription, and its probabilities raise the 66.5\% model to 69.0\% but the 69.2\% model only to 69.9\%, which indicates that the network and the OCR groups capture largely the same property of the ID code, namely how it was printed.
On Tobacco3482 the corresponding figures are 56.7\% before the second-code and OCR groups, 56.5\% with them, and 57.8\% with the network, whose probabilities alone yield an accuracy of 51.3\%, and none of these differences exceeds the standard deviation over the ten splits (0.7 to 1.2 points).

The file-name classifier of Section~\ref{sec:axis3-signal} uses the string-structure, format and prefix, and nearest-number groups of Table~\ref{tab:idcode-features}, computed from the first number in the file name of the page in the original release, together with the length of the name and, where the name contains a range, the number of pages in the range, for 130 features, and obtains 68.8\% on the RVL-CDIP test split; the folder name, which is the document identifier of the source archive, adds target encodings of its final characters and raises the accuracy to 70.3\%, and yields an accuracy of 41.4\% when used alone.
For Tobacco3482 the image file names yield an accuracy of 55.6\% (standard deviation 1.4 over the ten splits).
The nearest-number features, here as for the ID codes on the page, can use near-duplicate documents and multi-page documents whose pages are in different splits (Section~\ref{sec:axis2}), so part of this accuracy is due to the overlap between the splits.

\begin{figure}[tbp]
  \centering
  \includegraphics[width=0.85\linewidth]{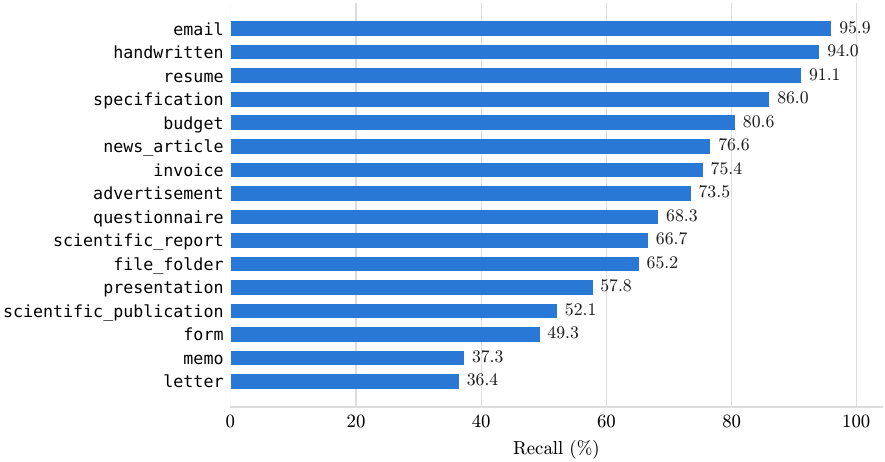}
  \caption{Recall of the code-only classifier by category on the 40,000
           documents of the RVL-CDIP test split.}
  \label{fig:idcode-recall}
\end{figure}

The signal is unevenly distributed over the categories, as Figure~\ref{fig:idcode-recall} shows for the classifier without the convolutional network: it recovers \cat{email} (95.9\%), \cat{handwritten} (94.0\%), \cat{resume} (91.1\%), and \cat{specification} (86.0\%) from the code in more than five cases out of six, whereas \cat{memo} (37.3\%) and \cat{letter} (36.4\%) are recovered in fewer than two cases out of five.
Table~\ref{tab:idcode-format}, which is computed from the annotated sample, shows part of the explanation, since the categories that are recovered most reliably are those whose codes have a format of their own: 98\% of the codes on \cat{handwritten} documents are two-part codes of five and four digits, a format that no more than 43\% of the codes in any other category share; 47\% of \cat{resume} documents do not have a code, against at most 19\% elsewhere, and the codes that they do have are vertical in only 31\% of cases where every other category is at 51\% or above; and \cat{email} has the largest share of codes containing letters (40\%).

\begin{table}[tbp]
  \centering
  \caption{Format of identification codes by category in the 11,200 annotated
           RVL-CDIP documents, in percent. \emph{No code} is the share of
           documents without a code; the remaining columns are shares of the
           documents that have one: codes whose box is taller than it is wide,
           two-part codes of five and four digits, ten-digit numbers, and
           codes containing at least one letter.}
  \label{tab:idcode-format}
  \small
  \setlength{\tabcolsep}{6pt}
  \begin{tabular}{lccccc}
  \toprule
  \textbf{Category} & \textbf{No code} & \textbf{Vertical} & \textbf{Two-part} & \textbf{Ten digits} & \textbf{Letters} \\
  \midrule
  \cat{advertisement}            & \gradient{4} & \gradient{73} & \gradient{30} & \gradient{39} & \gradient{5} \\
  \cat{budget}                   & \gradient{0} & \gradient{60} & \gradient{1} & \gradient{39} & \gradient{36} \\
  \cat{email}                    & \gradient{0} & \gradient{63} & \gradient{22} & \gradient{36} & \gradient{40} \\
  \cat{file\_folder}             & \gradient{11} & \gradient{86} & \gradient{0} & \gradient{76} & \gradient{3} \\
  \cat{form}                     & \gradient{3} & \gradient{88} & \gradient{18} & \gradient{65} & \gradient{2} \\
  \cat{handwritten}              & \gradient{0} & \gradient{94} & \gradient{98} & \gradient{0} & \gradient{1} \\
  \cat{invoice}                  & \gradient{7} & \gradient{74} & \gradient{8} & \gradient{40} & \gradient{10} \\
  \cat{letter}                   & \gradient{12} & \gradient{84} & \gradient{43} & \gradient{28} & \gradient{6} \\
  \cat{memo}                     & \gradient{13} & \gradient{70} & \gradient{2} & \gradient{50} & \gradient{9} \\
  \cat{news\_article}            & \gradient{0} & \gradient{73} & \gradient{0} & \gradient{61} & \gradient{27} \\
  \cat{presentation}             & \gradient{4} & \gradient{84} & \gradient{0} & \gradient{71} & \gradient{5} \\
  \cat{questionnaire}            & \gradient{5} & \gradient{85} & \gradient{27} & \gradient{39} & \gradient{3} \\
  \cat{resume}                   & \gradient{47} & \gradient{31} & \gradient{0} & \gradient{21} & \gradient{3} \\
  \cat{scientific\_publication}  & \gradient{19} & \gradient{51} & \gradient{0} & \gradient{47} & \gradient{27} \\
  \cat{scientific\_report}       & \gradient{2} & \gradient{86} & \gradient{0} & \gradient{71} & \gradient{4} \\
  \cat{specification}            & \gradient{1} & \gradient{76} & \gradient{18} & \gradient{73} & \gradient{1} \\
  \bottomrule
  \end{tabular}
\end{table}

In the earlier version of this experiment, a random forest that used only code features and was trained to discriminate between two categories at a time separated \cat{email} from \cat{budget} with 97\% accuracy, and \cat{email} and \cat{resume} were each distinguishable from the remaining categories at rates between 78\% and 97\%.

\section{Joint Remediation: Per-Condition Detail}\label{app:joint}

\begin{figure}[tbp]
  \centering
  \includegraphics[width=\linewidth]{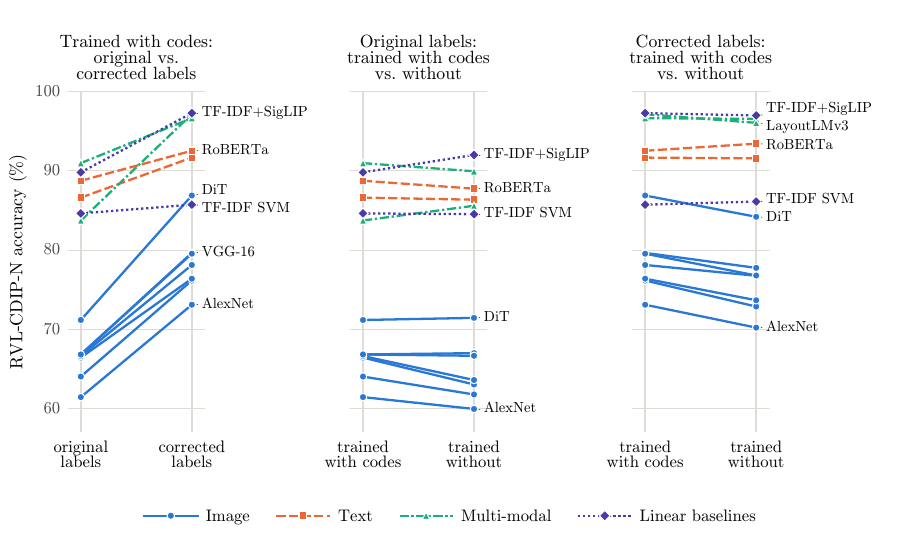}
  \caption{Accuracy on RVL-CDIP-\emph{N} by training variant, one line per
           model, from Table~\ref{tab:joint-remediation-n}. \emph{Left:}
           models trained on the original and on the corrected labels, both
           with codes. \emph{Middle} and \emph{right:} models trained with and
           without codes, under the original and under the corrected labels.}
  \label{fig:n-slopes}
\end{figure}

\begin{figure}[tbp]
  \centering
  \includegraphics[width=\linewidth]{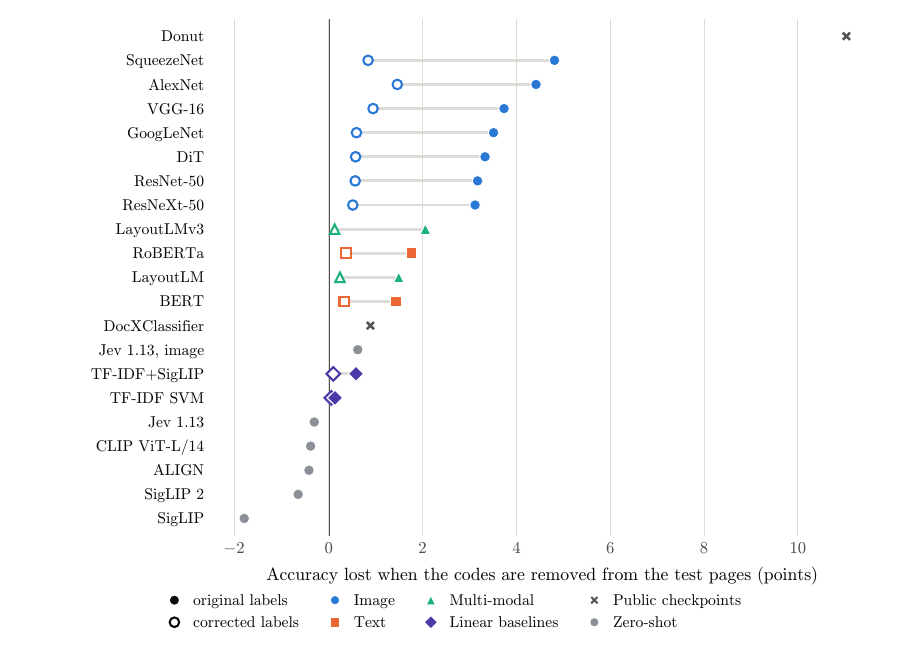}
  \caption{Test accuracy lost when the identification codes are removed from
           the test pages, for every model. Filled markers are models trained
           with codes on the original labels, hollow markers the same
           architectures trained with codes on the corrected labels; the
           public checkpoints and the zero-shot models have one value.}
  \label{fig:code-effect-dots}
\end{figure}

Figure~\ref{fig:n-slopes} shows the four training variants of Table~\ref{tab:joint-remediation-n} as connected pairs, and Figure~\ref{fig:code-effect-dots} charts the loss at test time shown in Figure~\ref{fig:before-after-codes} for every model, including the zero-shot models, which are 0.3 to 1.8 points more accurate on the pages without codes, except Jev with the page image, which is 0.6 points less accurate (Section~\ref{sec:joint}).

\begin{table}[tbp]
  \centering
  \caption{Accuracy (\%) of the twelve linear SVMs on the RVL-CDIP test split
           under the conditions of Table~\ref{tab:joint-remediation}, one run
           each. \emph{HW} adds two tokens that give the number of words on
           the page and the share that Textract marks as handwriting; the
           image features come from frozen encoders.}
  \label{tab:svm-baselines}
  \resizebox{\textwidth}{!}{
  \begin{tabular}{lccccccccccc}
    \toprule
    & \multicolumn{4}{c}{\textbf{All test pages}} & \multicolumn{4}{c}{\textbf{De-duplicated}} & \multicolumn{2}{c}{\textbf{Codes removed}} & \\
    \cmidrule(lr){2-5}\cmidrule(lr){6-9}\cmidrule(lr){10-11}
    \textbf{Model} & \textbf{Orig.} & \textbf{Drop.} & \textbf{Fixed} & \textbf{Fixed+RT} & \textbf{Orig.} & \textbf{Drop.} & \textbf{Fixed} & \textbf{Fixed+RT} & \textbf{Test} & \textbf{+RT} & \textbf{Full} \\
    \midrule
    $n$ & 40,000 & 35,280 & 36,446 & 36,446 & 25,923 & 22,530 & 23,132 & 23,132 & 40,000 & 40,000 & 23,132 \\
    \multicolumn{12}{l}{\emph{Text}} \\
    TF-IDF                  & 88.9 & 93.0 & 91.2 & 93.4 & 84.7 & 90.0 & 89.1 & 90.3 & 88.8 & 88.8 & 90.3 \\
    TF-IDF + HW             & 89.8 & 94.1 & 92.3 & 94.6 & 86.0 & 91.6 & 90.8 & 92.2 & 89.7 & 89.7 & 92.2 \\
    TF-IDF, Tesseract text  & 79.8 & 84.0 & 82.5 & 84.7 & 74.0 & 79.1 & 78.5 & 80.2 & 79.6 & 79.7 & 80.0 \\
    \midrule
    \multicolumn{12}{l}{\emph{Image features}} \\
    ResNet-50 (ImageNet)    & 76.9 & 81.5 & 80.4 & 82.3 & 74.7 & 80.0 & 79.4 & 80.8 & 75.3 & 76.4 & 80.4 \\
    CLIP                    & 87.5 & 92.8 & 91.5 & 93.8 & 85.4 & 91.8 & 91.1 & 93.0 & 87.1 & 87.4 & 92.8 \\
    SigLIP                  & 90.7 & 95.1 & 93.4 & 95.6 & 88.6 & 94.0 & 93.1 & 94.7 & 89.7 & 90.3 & 94.8 \\
    \midrule
    \multicolumn{12}{l}{\emph{Text and image features}} \\
    TF-IDF + ResNet-50      & 92.1 & 95.9 & 93.9 & 96.2 & 89.2 & 94.2 & 93.1 & 94.6 & 91.9 & 91.9 & 94.6 \\
    TF-IDF + CLIP           & 92.3 & 96.4 & 94.5 & 96.8 & 89.7 & 94.9 & 94.0 & 95.5 & 92.1 & 92.3 & 95.4 \\
    TF-IDF + SigLIP         & 93.1 & 96.8 & 94.8 & 97.1 & 90.6 & 95.5 & 94.4 & 96.0 & 92.5 & 92.8 & 96.0 \\
    TF-IDF + HW + ResNet-50 & 92.2 & 96.0 & 94.0 & 96.3 & 89.4 & 94.4 & 93.3 & 94.8 & 92.0 & 92.1 & 94.8 \\
    TF-IDF + HW + CLIP      & 92.4 & 96.4 & 94.5 & 96.9 & 89.7 & 95.0 & 94.1 & 95.6 & 92.2 & 92.4 & 95.5 \\
    TF-IDF + HW + SigLIP    & 93.1 & 96.9 & 94.9 & 97.1 & 90.7 & 95.6 & 94.5 & 96.1 & 92.5 & 92.8 & 96.0 \\
    \bottomrule
  \end{tabular}}
\end{table}

\begin{table}[tbp]
  \centering
  \footnotesize
  \caption{Accuracy (\%) of the linear SVMs on RVL-CDIP-\emph{N} by training
           variant, one run each. The SVM trained on Tesseract text is left
           out, because the RVL-CDIP-\emph{N} text is Textract output.}
  \label{tab:svm-baselines-n}
  \begin{tabular}{lcccc}
    \toprule
    \textbf{Model} & \textbf{Original} & \textbf{Fixed labels} & \textbf{Codes removed} & \textbf{Both} \\
    \midrule
    \multicolumn{5}{l}{\emph{Text}} \\
    TF-IDF                  & 84.6 & 85.7 & 84.5 & 86.1 \\
    TF-IDF + HW             & 90.5 & 92.1 & 90.3 & 92.1 \\
    \midrule
    \multicolumn{5}{l}{\emph{Image features}} \\
    ResNet-50 (ImageNet)    & 56.0 & 62.1 & 54.0 & 60.5 \\
    CLIP                    & 81.1 & 88.3 & 80.8 & 88.7 \\
    SigLIP                  & 87.7 & 95.6 & 90.1 & 95.8 \\
    \midrule
    \multicolumn{5}{l}{\emph{Text and image features}} \\
    TF-IDF + ResNet-50      & 85.9 & 91.9 & 86.3 & 91.9 \\
    TF-IDF + CLIP           & 88.3 & 95.6 & 88.0 & 95.1 \\
    TF-IDF + SigLIP         & 89.8 & 97.3 & 92.0 & 97.0 \\
    TF-IDF + HW + ResNet-50 & 86.1 & 93.2 & 86.3 & 93.6 \\
    TF-IDF + HW + CLIP      & 88.6 & 95.1 & 88.3 & 95.2 \\
    TF-IDF + HW + SigLIP    & 90.4 & 97.3 & 92.0 & 97.0 \\
    \bottomrule
  \end{tabular}
\end{table}

Tables~\ref{tab:svm-baselines} and~\ref{tab:svm-baselines-n} show all twelve linear SVMs, of which Tables~\ref{tab:joint-remediation} and~\ref{tab:joint-remediation-n} show the SVM on TF-IDF text and the SVM on TF-IDF text with SigLIP embeddings.
Each is a one-vs-rest linear SVM whose regularization strength, and for the fusion models the weight of the image block, is chosen on validation accuracy: three read the OCR text alone (TF-IDF features of the Textract text with and without the handwriting tokens, and of the Tesseract text), three read frozen image features alone (the pooled features of a ResNet-50 trained on ImageNet, CLIP embeddings and SigLIP embeddings), and six read the TF-IDF features concatenated with one of the three kinds of image features, with and without the handwriting tokens.
The handwriting tokens (\emph{HW} in the tables) are two pseudo-words added to the TF-IDF input of each page, both computed from its Textract output: the number of words on the page, in bins of powers of two, and the share of these words that Textract marks as handwritten, in tenths.
Because Textract decides whether a word is handwritten from the page image, an SVM with these tokens is no longer a text-only model.
With the corrected labels and retraining, the text-only SVMs obtain 93.4\% on Textract text, 94.6\% with the handwriting tokens and 84.7\% on Tesseract text, the image-only SVMs 82.3\% on the ResNet-50 features, 93.8\% on CLIP and 95.6\% on SigLIP embeddings, and the fusions 96.2, 96.8 and 97.1\% for the three encoders, to which the handwriting tokens add at most 0.2 points.
On RVL-CDIP-\emph{N} the same three fusions obtain 91.9, 95.6 and 97.3\%, against 85.7\% for the TF-IDF text alone and 62.1, 88.3 and 95.6\% for the three kinds of image features alone.
When the identification codes are removed from the test pages under the original labels, the SVM on ImageNet ResNet-50 features loses the most of the twelve (1.6 points, from 76.9 to 75.3\%), followed by the SVM on SigLIP embeddings (1.0 points), and the three text-only SVMs lose at most 0.2 points.

Table~\ref{tab:decision-models} extends the zero-shot text row of Table~\ref{tab:joint-remediation} to five decision models, each of which is given the Textract text of a page together with the sixteen category names and is run once without any training on RVL-CDIP; the columns are the test-side conditions of that table and the accuracy on RVL-CDIP-\emph{N}.

\begin{table}[tbp]
  \centering
  \caption{Accuracy (\%) of zero-shot decision models that read Textract text,
           under the test-side conditions of
           Table~\ref{tab:joint-remediation} and on RVL-CDIP-\emph{N}.
           \emph{Full} applies corrected labels, code removal and
           de-duplicated test pages together. Each model is run once.}
  \label{tab:decision-models}
  \resizebox{\textwidth}{!}{
  \begin{tabular}{lccccccccc}
    \toprule
    & \multicolumn{3}{c}{\textbf{All test pages}} & \multicolumn{3}{c}{\textbf{De-duplicated}} & & & \\
    \cmidrule(lr){2-4}\cmidrule(lr){5-7}
    \textbf{Model} & \textbf{Orig.} & \textbf{Drop.} & \textbf{Fixed} & \textbf{Orig.} & \textbf{Drop.} & \textbf{Fixed} & \textbf{Codes removed} & \textbf{Full} & \textbf{RVL-CDIP-\emph{N}} \\
    \midrule
    $n$ & 40,000 & 35,280 & 36,446 & 25,923 & 22,530 & 23,132 & 40,000 & 23,132 & 1,002 \\
    \midrule
    Jev 1.13          & 67.0 & 72.9 & 72.3 & 66.5 & 73.0 & 72.5 & 67.4 & 73.3 & 84.8 \\
    Bespoke-Nimble-9B & 63.0 & 68.3 & 67.7 & 62.6 & 68.8 & 68.4 & 63.3 & 69.0 & 80.4 \\
    clef-flash        & 59.3 & 64.0 & 63.5 & 57.5 & 62.8 & 62.3 & 59.4 & 62.3 & 82.2 \\
    Strands Decider 2B & 53.5 & 57.3 & 56.4 & 50.8 & 54.9 & 54.3 & 53.5 & 54.2 & 80.3 \\
    Laya (English)    & 35.2 & 37.6 & 37.1 & 31.3 & 33.7 & 33.4 & 35.4 & 33.6 & 52.2 \\
    \bottomrule
  \end{tabular}}
\end{table}

When Jev is given the page image in place of the OCR text, it is 6.2 points more accurate under the corrected test labels (78.5 against 72.3\%), and the two inputs lead to the same prediction on 84.1\% of the test pages.
Most of the gain is in categories that are recognized by their layout: the accuracy on \cat{handwritten} rises from 3 to 49\%, on \cat{file\_folder} from 63 to 83\%, on \cat{advertisement} from 84 to 93\% and on \cat{form} from 70 to 80\%, whereas \cat{presentation}, \cat{budget} and \cat{resume} remain between 32 and 65\% with either input.
To compare how much of a page the codes occupy in each category, we counted the pixels darker than half of the gray scale on 150 random test pages per category, with and without the codes.
The codes make up a median of 6.3\% of these pixels on the \cat{file\_folder} pages and 9.8\% on the 314 \cat{file\_folder} pages that Jev classifies correctly only with the codes (corrected labels), 6.2\% on the \cat{email} pages, on which the accuracy of Jev does not change, and between 0.1 and 1.9\% in the other fourteen categories.
Without the codes, Jev assigns 144 of these 314 pages to \cat{scientific\_publication} and 66 to \cat{handwritten}.

Table~\ref{tab:llm-prompts} shows the accuracy of the four large language models of Table~\ref{tab:joint-remediation} with each of the three prompts of Appendix~\ref{app:prompts}, under the test-side conditions that do not remove the identification codes and on RVL-CDIP-\emph{N}; Tables~\ref{tab:joint-remediation} and~\ref{tab:joint-remediation-n} show the rows of Prompt~1.
The models were run on the pages that have OCR text and are scored on those pages, so that the number of pages in each condition is smaller than in Table~\ref{tab:joint-remediation}, and Appendix~\ref{app:promptstats} tests the differences between the prompts under the original condition.

\begin{table}[tbp]
  \centering\footnotesize\setlength{\tabcolsep}{4pt}
  \caption{Accuracy (\%) of the zero-shot large language models by prompt, on
           the pages that have OCR text, under the test-side conditions of
           Table~\ref{tab:joint-remediation} without code removal and on
           RVL-CDIP-\emph{N}. Each model is run once per prompt.}
  \label{tab:llm-prompts}
  \begin{tabular}{llccccccc}
    \toprule
    & & \multicolumn{3}{c}{\textbf{All test pages}} & \multicolumn{3}{c}{\textbf{De-duplicated}} & \\
    \cmidrule(lr){3-5}\cmidrule(lr){6-8}
    \textbf{Model} & \textbf{Prompt} & \textbf{Orig.} & \textbf{Drop.} & \textbf{Fixed} & \textbf{Orig.} & \textbf{Drop.} & \textbf{Fixed} & \textbf{RVL-CDIP-\emph{N}} \\
    \midrule
    $n$ & & 38,377 & 33,796 & 34,922 & 24,324 & 21,062 & 21,624 & 1,001 \\
    \midrule
    Llama 4 Maverick & 1, names only      & 67.6 & 73.5 & 72.8 & 65.3 & 72.0 & 71.6 & 87.8 \\
                     & 2, corpus-grounded & 73.8 & 80.4 & 79.9 & 69.3 & 76.6 & 76.4 & 88.3 \\
                     & 3, generic         & 67.8 & 73.6 & 73.0 & 65.2 & 72.0 & 71.4 & 86.4 \\
    \midrule
    gpt-oss-20b      & 1, names only      & 64.6 & 70.2 & 69.7 & 63.8 & 70.1 & 69.7 & 84.3 \\
                     & 2, corpus-grounded & 70.6 & 76.7 & 76.2 & 67.9 & 74.8 & 74.5 & 83.2 \\
                     & 3, generic         & 64.8 & 70.4 & 69.8 & 63.3 & 69.6 & 69.2 & 82.3 \\
    \midrule
    Qwen 3 32B       & 1, names only      & 62.3 & 67.3 & 66.4 & 62.6 & 68.9 & 68.5 & 85.4 \\
                     & 2, corpus-grounded & 63.6 & 68.5 & 67.7 & 61.5 & 67.7 & 67.3 & 81.6 \\
                     & 3, generic         & 58.6 & 63.1 & 62.3 & 56.7 & 62.1 & 61.7 & 81.8 \\
    \midrule
    Mistral 7B       & 1, names only      & 57.0 & 61.6 & 60.9 & 54.3 & 59.8 & 59.4 & 78.2 \\
                     & 2, corpus-grounded & 54.6 & 58.6 & 58.1 & 54.5 & 59.7 & 59.4 & 76.4 \\
                     & 3, generic         & 55.1 & 59.5 & 58.8 & 53.0 & 58.3 & 58.0 & 68.3 \\
    \bottomrule
  \end{tabular}
\end{table}

Table~\ref{tab:macro-accuracy} shows macro accuracy, the mean of the per-category recalls over the categories present, for the models of Tables~\ref{tab:joint-remediation} and~\ref{tab:joint-remediation-n}; we did not compute it for the four large language models.
On RVL-CDIP-\emph{N}, which covers twelve categories with unequal numbers of pages, macro accuracy under the original labels is higher than micro accuracy for 13 of the 15 trained models and public checkpoints, by up to 7.2 points for Donut and 6.3 points for LayoutLMv3, because the 176 \cat{handwritten} pages, which make up 17.6\% of RVL-CDIP-\emph{N} and on which several of these models are weakest, count for one twelfth of the macro average.
Training on the corrected labels raises the macro accuracy of every retrained model on RVL-CDIP-\emph{N}, by 1.6 to 9.6 points, which is less than the gain in micro accuracy for every retrained model except the TF-IDF SVM (1.6 against 1.1 points), so that part of the micro gain is due to the larger weight of the \cat{handwritten} pages.

\begin{table}[tbp]
  \centering
  \footnotesize
  \caption{Macro accuracy (\%, mean per-category recall) on the RVL-CDIP test
           split, in three of the conditions of Table~\ref{tab:conditions},
           and on RVL-CDIP-\emph{N} by training variant, as in
           Table~\ref{tab:joint-remediation-n}.}
  \label{tab:macro-accuracy}
  \setlength{\tabcolsep}{4pt}
  \begin{tabular}{lccccccc}
    \toprule
    & \multicolumn{3}{c}{\textbf{Test split}} & \multicolumn{4}{c}{\textbf{RVL-CDIP-\emph{N}}} \\
    \cmidrule(lr){2-4}\cmidrule(lr){5-8}
    \textbf{Model} & \textbf{Orig.} & \textbf{Fixed+RT} & \textbf{Full} & \textbf{Original} & \textbf{Fixed} & \textbf{No codes} & \textbf{Both} \\
    \midrule
    \multicolumn{8}{l}{\emph{Image}} \\
    AlexNet             & 89.3 & 92.7 & 87.0 & 62.9 & 71.7 & 63.4 & 70.1 \\
    GoogLeNet           & 89.2 & 93.0 & 87.5 & 66.2 & 74.2 & 65.1 & 72.2 \\
    SqueezeNet          & 88.6 & 92.5 & 87.3 & 66.2 & 74.9 & 64.8 & 73.1 \\
    VGG-16              & 91.2 & 94.5 & 89.5 & 69.4 & 77.8 & 70.3 & 77.1 \\
    ResNet-50           & 90.8 & 94.2 & 89.3 & 69.7 & 77.2 & 67.1 & 76.5 \\
    ResNeXt-50          & 91.3 & 94.6 & 89.9 & 68.8 & 78.3 & 69.5 & 76.2 \\
    DiT-base            & 92.7 & 95.9 & 92.4 & 76.6 & 86.2 & 76.2 & 84.7 \\
    \midrule
    \multicolumn{8}{l}{\emph{Text}} \\
    BERT-base           & 92.3 & 95.6 & 92.8 & 89.5 & 93.6 & 88.7 & 93.5 \\
    RoBERTa-base        & 92.5 & 96.0 & 93.2 & 91.2 & 94.1 & 89.8 & 94.4 \\
    \midrule
    \multicolumn{8}{l}{\emph{Multi-modal}} \\
    LayoutLM-base       & 93.4 & 96.7 & 94.4 & 92.8 & 96.3 & 90.6 & 95.8 \\
    LayoutLMv3-base     & 94.8 & 97.5 & 95.6 & 90.0 & 97.2 & 90.5 & 96.1 \\
    \midrule
    \multicolumn{8}{l}{\emph{Linear baselines}} \\
    TF-IDF SVM          & 88.9 & 93.4 & 89.4 & 86.4 & 88.0 & 86.5 & 88.3 \\
    TF-IDF + SigLIP SVM & 93.1 & 97.1 & 94.9 & 89.3 & 96.7 & 91.7 & 96.2 \\
    \midrule
    \multicolumn{8}{l}{\emph{Public checkpoints}} \\
    DocXClassifier      & 94.0 & -- & 91.8 & 79.7 & -- & -- & -- \\
    Donut               & 95.2 & -- & 87.3 & 86.1 & -- & -- & -- \\
    \midrule
    \multicolumn{8}{l}{\emph{Zero-shot, image}} \\
    CLIP ViT-L/14       & 43.7 & -- & 48.3 & 81.6 & -- & -- & -- \\
    ALIGN               & 47.0 & -- & 52.1 & 86.0 & -- & -- & -- \\
    SigLIP              & 56.3 & -- & 62.8 & 92.8 & -- & -- & -- \\
    SigLIP 2            & 54.0 & -- & 58.9 & 89.5 & -- & -- & -- \\
    \midrule
    \multicolumn{8}{l}{\emph{Zero-shot, text}} \\
    Jev 1.13            & 67.1 & -- & 74.5 & 91.6 & -- & -- & -- \\
    \bottomrule
  \end{tabular}
\end{table}

\section{Out-of-Distribution Supplementary Results}\label{app:ood}

\begin{table}[tbp]
  \centering\footnotesize\setlength{\tabcolsep}{3.8pt}
  \caption{Out-of-scope separability (AUC under the energy score) by training
           variant, as Table~\ref{tab:ood-variants}, for the models that
           produce logits.}
  \label{tab:ood-variants-energy}
  \begin{tabular}{lcccccccc}
    \toprule
    & \multicolumn{4}{c}{\textbf{\emph{T-O}}} & \multicolumn{4}{c}{\textbf{\emph{N-O}}} \\
    \cmidrule(lr){2-5}\cmidrule(lr){6-9}
    \textbf{Model} & \textbf{Original} & \textbf{Fixed} & \textbf{No codes} & \textbf{Both} & \textbf{Original} & \textbf{Fixed} & \textbf{No codes} & \textbf{Both} \\
    \midrule
    \multicolumn{9}{l}{\emph{Image}} \\
    AlexNet           & 0.912 & 0.931 & 0.894 & 0.923 & 0.676 & 0.741 & 0.661 & 0.723 \\
    GoogLeNet         & 0.874 & 0.879 & 0.860 & 0.856 & 0.627 & 0.691 & 0.626 & 0.660 \\
    SqueezeNet        & 0.885 & 0.882 & 0.852 & 0.869 & 0.644 & 0.719 & 0.665 & 0.722 \\
    VGG-16            & 0.924 & 0.943 & 0.907 & 0.936 & 0.683 & 0.770 & 0.664 & 0.751 \\
    ResNet-50         & 0.897 & 0.895 & 0.887 & 0.884 & 0.645 & 0.709 & 0.634 & 0.701 \\
    ResNeXt-50        & 0.907 & 0.908 & 0.887 & 0.894 & 0.641 & 0.728 & 0.634 & 0.705 \\
    DiT-base          & 0.911 & 0.926 & 0.892 & 0.927 & 0.714 & 0.800 & 0.704 & 0.793 \\
    \midrule
    \multicolumn{9}{l}{\emph{Text}} \\
    BERT-base         & 0.837 & 0.869 & 0.821 & 0.864 & 0.834 & 0.888 & 0.817 & 0.884 \\
    RoBERTa-base      & 0.848 & 0.870 & 0.838 & 0.877 & 0.848 & 0.895 & 0.839 & 0.902 \\
    \midrule
    \multicolumn{9}{l}{\emph{Multi-modal}} \\
    LayoutLM-base     & 0.851 & 0.879 & 0.846 & 0.880 & 0.848 & 0.907 & 0.840 & 0.919 \\
    LayoutLMv3-base   & 0.905 & 0.943 & 0.905 & 0.933 & 0.840 & 0.928 & 0.845 & 0.899 \\
    \midrule
    \multicolumn{9}{l}{\emph{Public checkpoints}} \\
    DocXClassifier    & 0.809 & -- & -- & -- & 0.626 & -- & -- & -- \\
    Donut             & 0.960 & -- & -- & -- & 0.840 & -- & -- & -- \\
    \midrule
    \multicolumn{9}{l}{\emph{Zero-shot, image}} \\
    CLIP ViT-L/14     & 0.861 & -- & -- & -- & 0.756 & -- & -- & -- \\
    ALIGN             & 0.611 & -- & -- & -- & 0.666 & -- & -- & -- \\
    SigLIP            & 0.864 & -- & -- & -- & 0.812 & -- & -- & -- \\
    SigLIP 2          & 0.796 & -- & -- & -- & 0.820 & -- & -- & -- \\
    \bottomrule
  \end{tabular}
\end{table}

All models are evaluated on the same three sets: the 40,000 documents of the RVL-CDIP test split, the 1,002 of RVL-CDIP-\emph{N}, and the 2,915 of RVL-CDIP-\emph{O}.
Every image is converted to grayscale on load, for every model and every set, because the released RVL-CDIP-\emph{O} images are in color while RVL-CDIP and RVL-CDIP-\emph{N} are grayscale, and color alone would otherwise mark the out-of-scope documents.
The text and multi-modal models use Amazon Textract OCR.
The five CNNs, BERT, RoBERTa, LayoutLM and LayoutLMv3 are the models of Section~\ref{sec:joint}, each trained once on the original and once on the corrected labels; DiT, DocXClassifier and Donut are their authors' published RVL-CDIP checkpoints, for which no corrected-label version exists.
Donut is generative, so its logits are the decoder's next-token logits over the sixteen category tokens that follow its task prefix, and its prediction is the category it would generate if restricted to a valid one.
The zero-shot vision-language models score each image against sixteen captions of the form \emph{This is a photo of a letter document.}, one per category. Their logits are the image--text similarities scaled by each model's learned temperature (and, for SigLIP and SigLIP~2, offset by its learned bias), so the scores computed through a softmax depend on that temperature; we use each model's own.

Section~\ref{sec:joint-abstention} reports maximum softmax probability and the energy score.
We also compute two further post-hoc scores from the same logits.
MaxLogit \citep{hendrycks2022scaling} is the largest unnormalized logit, $\max_i z_i$, which keeps the magnitude information that the softmax discards.
GEN \citep{liu2023gen} is the negated generalized entropy $-\sum_{i=1}^{K} p_i^{\gamma}(1-p_i)^{\gamma}$ of the softmax probabilities $p$, with $\gamma=0.1$; it weights small probabilities heavily, and so remains informative when the largest is close to one.
For the zero-shot models we add MCM \citep{ming2022mcm}, the confidence function usually used with vision-language models: maximum softmax probability over the raw cosine similarities at temperature 1 and not at the model's learned temperature.
Table~\ref{tab:ood-auc-all} shows AUC under every score, and Table~\ref{tab:ood-fpr95} the false positive rate at 95\% true positive rate (FPR95): the share of RVL-CDIP-\emph{O} accepted at the threshold that accepts 95\% of in-domain documents.

\begin{table}[tbp]
\centering\footnotesize\setlength{\tabcolsep}{3.8pt}
\caption{AUC under each confidence function. (corrected): retrained on the corrected labels. MCM needs cosine similarities, so is defined for the zero-shot models only.}\label{tab:ood-auc-all}
\begin{tabular}{lcccccccccc}\toprule
 & \multicolumn{5}{c}{\textbf{\emph{T-O}}} & \multicolumn{5}{c}{\textbf{\emph{N-O}}} \\
\cmidrule(lr){2-6}\cmidrule(lr){7-11}
\textbf{Model} & \textbf{MSP} & \textbf{Energy} & \textbf{MaxLogit} & \textbf{GEN} & \textbf{MCM} & \textbf{MSP} & \textbf{Energy} & \textbf{MaxLogit} & \textbf{GEN} & \textbf{MCM} \\
\midrule
AlexNet & 0.880 & 0.914 & 0.914 & 0.912 & -- & 0.645 & 0.673 & 0.673 & 0.673 & -- \\
AlexNet (corrected) & 0.886 & 0.919 & 0.919 & 0.917 & -- & 0.740 & 0.751 & 0.751 & 0.755 & -- \\
GoogLeNet & 0.837 & 0.845 & 0.845 & 0.849 & -- & 0.604 & 0.594 & 0.595 & 0.599 & -- \\
GoogLeNet (corrected) & 0.842 & 0.829 & 0.829 & 0.840 & -- & 0.684 & 0.661 & 0.662 & 0.674 & -- \\
ResNet-50 & 0.874 & 0.875 & 0.875 & 0.884 & -- & 0.624 & 0.600 & 0.601 & 0.611 & -- \\
ResNet-50 (corrected) & 0.868 & 0.861 & 0.862 & 0.871 & -- & 0.689 & 0.682 & 0.683 & 0.690 & -- \\
ResNeXt-50 & 0.882 & 0.889 & 0.889 & 0.895 & -- & 0.618 & 0.612 & 0.613 & 0.619 & -- \\
ResNeXt-50 (corrected) & 0.877 & 0.874 & 0.874 & 0.881 & -- & 0.724 & 0.702 & 0.703 & 0.715 & -- \\
VGG-16 & 0.889 & 0.930 & 0.930 & 0.928 & -- & 0.661 & 0.668 & 0.669 & 0.672 & -- \\
VGG-16 (corrected) & 0.881 & 0.925 & 0.925 & 0.923 & -- & 0.759 & 0.765 & 0.765 & 0.771 & -- \\
DiT-base & 0.895 & 0.892 & 0.893 & 0.895 & -- & 0.733 & 0.759 & 0.749 & 0.745 & -- \\
DiT-large & 0.814 & 0.825 & 0.831 & 0.787 & -- & 0.691 & 0.731 & 0.725 & 0.687 & -- \\
DocXClassifier & 0.897 & 0.808 & 0.857 & 0.867 & -- & 0.707 & 0.626 & 0.672 & 0.688 & -- \\
Donut & 0.909 & 0.960 & 0.960 & 0.954 & -- & 0.793 & 0.840 & 0.840 & 0.828 & -- \\
\midrule
BERT & 0.859 & 0.845 & 0.845 & 0.849 & -- & 0.841 & 0.858 & 0.857 & 0.851 & -- \\
BERT (corrected) & 0.854 & 0.834 & 0.834 & 0.846 & -- & 0.866 & 0.872 & 0.871 & 0.876 & -- \\
RoBERTa & 0.868 & 0.857 & 0.857 & 0.861 & -- & 0.838 & 0.871 & 0.871 & 0.865 & -- \\
RoBERTa (corrected) & 0.851 & 0.832 & 0.833 & 0.841 & -- & 0.908 & 0.908 & 0.908 & 0.908 & -- \\
\midrule
LayoutLM & 0.861 & 0.865 & 0.867 & 0.871 & -- & 0.834 & 0.858 & 0.858 & 0.855 & -- \\
LayoutLM (corrected) & 0.820 & 0.827 & 0.828 & 0.832 & -- & 0.802 & 0.829 & 0.828 & 0.820 & -- \\
LayoutLMv3 & 0.893 & 0.907 & 0.906 & 0.906 & -- & 0.827 & 0.838 & 0.838 & 0.836 & -- \\
LayoutLMv3 (corrected) & 0.894 & 0.909 & 0.908 & 0.910 & -- & 0.876 & 0.910 & 0.909 & 0.899 & -- \\
\midrule
CLIP ViT-L/14 & 0.623 & 0.861 & 0.864 & 0.606 & 0.616 & 0.806 & 0.756 & 0.785 & 0.852 & 0.845 \\
ALIGN & 0.655 & 0.612 & 0.633 & 0.678 & 0.677 & 0.861 & 0.666 & 0.731 & 0.891 & 0.894 \\
SigLIP & 0.710 & 0.862 & 0.863 & 0.807 & 0.800 & 0.890 & 0.812 & 0.838 & 0.925 & 0.926 \\
SigLIP 2 & 0.673 & 0.797 & 0.799 & 0.671 & 0.681 & 0.916 & 0.820 & 0.861 & 0.946 & 0.945 \\
\bottomrule\end{tabular}\end{table}
\begin{table}[tbp]
\centering\footnotesize\setlength{\tabcolsep}{3.8pt}
\caption{FPR95 (\%) under each confidence function: the share of RVL-CDIP-\emph{O} accepted at the threshold that accepts 95\% of in-domain documents. Lower is better. (corrected): retrained on the corrected labels. MCM needs cosine similarities, so is defined for the zero-shot models only.}\label{tab:ood-fpr95}
\begin{tabular}{lcccccccccc}\toprule
 & \multicolumn{5}{c}{\textbf{\emph{T-O}}} & \multicolumn{5}{c}{\textbf{\emph{N-O}}} \\
\cmidrule(lr){2-6}\cmidrule(lr){7-11}
\textbf{Model} & \textbf{MSP} & \textbf{Energy} & \textbf{MaxLogit} & \textbf{GEN} & \textbf{MCM} & \textbf{MSP} & \textbf{Energy} & \textbf{MaxLogit} & \textbf{GEN} & \textbf{MCM} \\
\midrule
AlexNet & 69.0 & 51.0 & 51.3 & 50.8 & -- & 90.5 & 91.6 & 91.2 & 90.5 & -- \\
AlexNet (corrected) & 66.2 & 49.8 & 50.1 & 50.6 & -- & 86.3 & 86.7 & 85.6 & 85.5 & -- \\
GoogLeNet & 74.5 & 62.8 & 63.1 & 62.8 & -- & 93.3 & 92.8 & 93.0 & 92.9 & -- \\
GoogLeNet (corrected) & 73.8 & 65.5 & 65.9 & 65.7 & -- & 90.7 & 87.2 & 86.8 & 86.6 & -- \\
ResNet-50 & 67.1 & 56.3 & 56.0 & 55.4 & -- & 92.5 & 93.6 & 93.4 & 93.2 & -- \\
ResNet-50 (corrected) & 68.6 & 58.3 & 58.7 & 58.4 & -- & 91.9 & 90.4 & 90.5 & 90.5 & -- \\
ResNeXt-50 & 63.9 & 49.4 & 50.1 & 50.7 & -- & 92.4 & 92.1 & 92.4 & 91.9 & -- \\
ResNeXt-50 (corrected) & 66.0 & 54.9 & 55.1 & 54.4 & -- & 84.8 & 87.1 & 86.8 & 87.3 & -- \\
VGG-16 & 61.3 & 42.2 & 42.2 & 43.6 & -- & 89.2 & 90.7 & 90.8 & 90.3 & -- \\
VGG-16 (corrected) & 61.4 & 48.2 & 48.1 & 48.2 & -- & 85.3 & 86.4 & 85.8 & 85.3 & -- \\
DiT-base & 57.8 & 49.5 & 53.3 & 49.7 & -- & 84.6 & 85.2 & 83.7 & 83.4 & -- \\
DiT-large & 58.4 & 55.5 & 55.7 & 53.0 & -- & 86.6 & 84.0 & 84.9 & 84.3 & -- \\
DocXClassifier & 48.3 & 66.5 & 52.1 & 56.0 & -- & 87.9 & 91.9 & 90.4 & 89.7 & -- \\
Donut & 41.2 & 19.8 & 20.5 & 25.0 & -- & 83.4 & 74.8 & 74.4 & 81.8 & -- \\
\midrule
BERT & 82.9 & 88.7 & 88.0 & 87.3 & -- & 75.2 & 64.8 & 65.4 & 67.2 & -- \\
BERT (corrected) & 80.4 & 83.0 & 83.6 & 82.5 & -- & 69.6 & 63.5 & 64.3 & 64.2 & -- \\
RoBERTa & 81.8 & 87.2 & 87.4 & 87.2 & -- & 69.3 & 60.1 & 61.0 & 61.6 & -- \\
RoBERTa (corrected) & 83.7 & 83.5 & 84.6 & 83.0 & -- & 60.0 & 53.5 & 53.5 & 53.4 & -- \\
\midrule
LayoutLM & 58.1 & 83.4 & 82.3 & 75.3 & -- & 67.6 & 59.5 & 60.7 & 59.1 & -- \\
LayoutLM (corrected) & 65.1 & 89.5 & 88.9 & 80.5 & -- & 77.5 & 68.9 & 68.7 & 70.4 & -- \\
LayoutLMv3 & 53.7 & 39.1 & 40.9 & 40.8 & -- & 79.1 & 74.1 & 76.3 & 74.8 & -- \\
LayoutLMv3 (corrected) & 54.3 & 43.1 & 43.9 & 44.9 & -- & 59.5 & 47.9 & 50.1 & 50.8 & -- \\
\midrule
CLIP ViT-L/14 & 91.8 & 54.9 & 51.8 & 94.1 & 93.7 & 66.4 & 81.6 & 79.5 & 57.2 & 58.9 \\
ALIGN & 89.5 & 84.7 & 84.8 & 92.2 & 91.0 & 49.8 & 81.1 & 72.6 & 46.7 & 45.6 \\
SigLIP & 82.0 & 53.3 & 54.5 & 76.4 & 77.7 & 48.7 & 73.2 & 65.0 & 36.7 & 34.4 \\
SigLIP 2 & 92.9 & 63.5 & 64.8 & 98.4 & 96.9 & 37.0 & 80.2 & 71.9 & 24.5 & 26.0 \\
\bottomrule\end{tabular}\end{table}

None of the alternative scores changes a conclusion of Section~\ref{sec:joint-abstention} or Section~\ref{sec:interaction}: under every score, the CNNs separate the \emph{N-O} setting worse than every text and multi-modal model, and retraining on the corrected labels raises \emph{N-O} AUC in eight of the nine pairs.
With sixteen classes MaxLogit cannot differ from the energy score by more than $\log 16$, and its AUC is within 0.009 of energy's for every supervised model except DocXClassifier, whose energy score separates unusually poorly (\emph{T-O} AUC 0.808, against 0.897 under maximum softmax probability) and which MaxLogit partly recovers (0.857).
Among the supervised models, GEN differs from the better of maximum softmax probability and energy by between $-0.044$ (DiT-large, \emph{N-O}) and $+0.009$ (ResNet-50, \emph{T-O}).
For the zero-shot models the four scores disagree far more.
In the \emph{N-O} setting GEN is the best of the four for every zero-shot model (0.852 to 0.946) and energy the worst, while in the \emph{T-O} setting MaxLogit is the best for three of the four.
MCM raises every zero-shot model's \emph{N-O} AUC above maximum softmax probability at the model's own temperature, by 0.029 to 0.039, and agrees with GEN to within 0.010 in both settings: GEN's small exponent flattens the probabilities much as a higher temperature does.
The scores agree less closely at a fixed operating point than across all thresholds: LayoutLM's \emph{T-O} AUC varies by only 0.011 across the four scores, yet at the 95\% threshold it accepts between 58.1\% and 83.4\% of RVL-CDIP-\emph{O}, depending on the score.

Six RVL-CDIP-\emph{O} documents do not yield OCR text, so text models are shown an empty document for them.
Excluding them does not change any AUC by as much as 0.001 or any FPR95 by as much as 0.2 percentage points.

\begin{figure}[tbp]
  \centering
  \includegraphics[width=\linewidth]{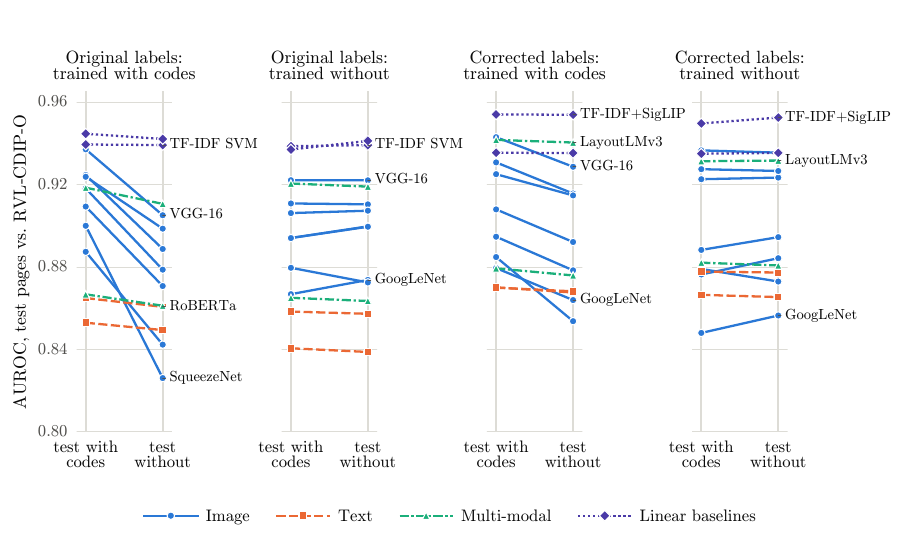}
  \caption{Out-of-scope separability (AUC, maximum logit) of the RVL-CDIP test
           split against RVL-CDIP-\emph{O} with the identification codes on
           the test pages and with the codes removed, for models trained with
           and without codes under the original and the corrected labels. One
           line per model; retrained models are means over three seeds (one
           run for the linear baselines).}
  \label{fig:ood-code-slopes}
\end{figure}

Figure~\ref{fig:ood-code-slopes} repeats the \emph{T-O} comparison with the identification codes removed from the test pages while the model and the RVL-CDIP-\emph{O} pages stay the same.
For the image models trained with codes the AUC decreases by 2.3 to 8.1 points under the original labels and by 1.0 to 3.3 points under the corrected labels, in every seed, whereas for the same architectures trained without codes it changes by between $-0.9$ and $+0.7$ points, so that part of what separates the test pages from out-of-scope documents for a model trained with codes is the code itself.

\section{Zero-Shot Classification Prompts}\label{app:prompts}

The zero-shot large language models evaluated in Section~\ref{sec:joint} are prompted with one of three prompts, reproduced below exactly as issued.
All three have the same structure (a system role, the enumerated list of the sixteen categories, an instruction to answer with a category name and nothing else, and a JSON response schema) and differ only in how much the categories are described.
Prompt~1 gives the category names alone.
Prompt~2 defines each category against the RVL-CDIP corpus itself, so that the sub-types present in the data (for example, that \cat{budget} covers campaign contribution requests and check stubs) are stated explicitly.
Prompt~3 gives generic dictionary definitions written without reference to the corpus.
The contrast between Prompts~2 and~3 is the one of interest: it separates knowing what a category is called from knowing what the benchmark's annotators put in it, a distinction that, as Section~\ref{sec:guidelines} argues, the label names alone do not express.

In every case the OCR'd document text is appended directly after the final line of the prompt, and the model's reply is parsed as JSON and matched against the sixteen category names after lower-casing and mapping interior whitespace to underscores.
Replies that do not parse, or that do not name a category, are counted as errors and are not discarded.

\begin{promptbox}[title={Prompt 1: category names only}]
\begin{lstlisting}[style=promptstyle]
You are a document categorization system.

Classify the text of a document into one of the following categories:

    1. advertisement
    2. budget
    3. email
    4. file_folder
    5. form
    6. handwritten
    7. invoice
    8. letter
    9. memo
    10. news_article
    11. presentation
    12. questionnaire
    13. resume
    14. scientific_publication
    15. scientific_report
    16. specification

Respond only with the category name. Do not provide an explanation.

Respond with the following JSON format:
{"prediction": <category name>}

Here is the document text to classify:
\end{lstlisting}
\end{promptbox}

\begin{promptbox}[title={Prompt 2: corpus-grounded definitions}]
\begin{lstlisting}[style=promptstyle]
You are a document categorization system.

Classify the text of a document into one of the following categories:

    1. advertisement: advertisements from print-form media like newspapers, magazines, and radio/television scripts.
    2. budget: various budget documents such as expense, spending, sales, cash, and accounting reports and forecasts; campaign contribution requests; checks and check stubs.
    3. email: printed emails.
    4. file_folder: folders and binders. These will typically have little text.
    5. form: form documents with form-like elements. These include fax forms.
    6. handwritten: handwritten documents.
    7. invoice: invoices, bills, quotations, and estimates.
    8. letter: letters, often with letterhead and "Dear..." salutations.
    9. memo: memo or memoranda documents or inter-office correspondence documents, often with clear "TO", "FROM" headings.
    10. news_article: news articles in the form of clippings from newspapers and other print-form news media.
    11. presentation: presentation and overhead slides, transcripts of speeches and statements, and press releases.
    12. questionnaire: customer surveys and questionnaires, survey prompts.
    13. resume: resumes, curricula vitae (CVs), biographical sketches, executive biographies, business cards.
    14. scientific_publication: scientific publications or articles from scientific journals and book chapters; includes book title pages.
    15. scientific_report: scientific reports like bioassay, pathology, test reports, charts, graphs, tables, research progress reports, research proposals.
    16. specification: specifications like data sheets (including material safety data sheets); product, material, and test specifications.

Respond only with the category name. Do not provide an explanation.

Respond with the following JSON format:
{"prediction": <category name>}

Here is the document text to classify:
\end{lstlisting}
\end{promptbox}

\begin{promptbox}[title={Prompt 3: generic definitions}]
\begin{lstlisting}[style=promptstyle]
You are a document categorization system.

Classify the text of a document into one of the following categories:

    1. advertisement: a notice or announcement in a public medium promoting a product.
    2. budget: an estimate, often itemized, of expected income and expense for a given period in the future.
    3. email: a message sent by email.
    4. file_folder: a folded sheet of light cardboard used to cover or hold documents.
    5. form: a document with blank spaces to be filled in with particulars before it is executed.
    6. handwritten: handwritten documents.
    7. invoice: an itemized bill for goods sold or services provided, containing individual prices, the total charge, and the terms.
    8. letter: a written or printed communication addressed to a person or organization and usually transmitted by mail.
    9. memo: an informal message, especially one sent between two or more employees of the same company, concerning company business.
    10. news_article: the presentation of a report on recent or new events in a newspaper or other periodical or on radio or television.
    11. presentation: a document for a formal talk given to an audience to share information, persuade, inspire, or demonstrate something, often using visuals like slides to support the spoken message.
    12. questionnaire: a list of questions, usually printed, submitted for replies that can be analyzed for usable information.
    13. resume: a brief written account of personal, educational, and professional qualifications and experience, as that prepared by an applicant for a job.
    14. scientific_publication: periodical that shares new scientific research and knowledge with the scientific community and the public.
    15. scientific_report: document that details a scientific experiment or research project, presenting the methodology, data, results, and conclusions to inform readers and allow for reproduction of the study.
    16. specification: a detailed description or assessment of requirements, dimensions, materials, etc.

Respond only with the category name. Do not provide an explanation.

Respond with the following JSON format:
{"prediction": <category name>}

Here is the document text to classify:
\end{lstlisting}
\end{promptbox}

\subsection*{Rejection Variant for Out-of-Scope Detection}

For the out-of-scope evaluation on RVL-CDIP-\emph{O} (Appendix~\ref{app:promptstats}), each prompt is issued in a rejection variant that adds a seventeenth option, \cat{none\_of\_the\_above}, and changes nothing else.
The variants are derived from the prompts above by program, so the structure, the sixteen category entries, and the response instructions are identical to those shown.
The new entry follows the style of the prompt it extends (a bare name in Prompt~1, a one-line gloss in Prompts~2 and~3), so that the rejection option is never the only entry with, or without, a definition.
Each variant also gains a guardrail sentence immediately before the response instruction, intended to stop a model from using the rejection option to express uncertainty between two of the sixteen categories.
The box below shows the end of the Prompt~2 variant exactly as issued.
In Prompt~1, item 17 reads \cat{17. none\_of\_the\_above}; in Prompt~3 it reads \cat{17. none\_of\_the\_above: a document that does not belong to any of the sixteen categories above.}

\begin{promptbox}[title={Rejection variant of Prompt 2, from item 16 to the end}]
\begin{lstlisting}[style=promptstyle]
    16. specification: specifications like data sheets (including material safety data sheets); product, material, and test specifications.
    17. none_of_the_above: the document is not any of the sixteen types above.

Choose none_of_the_above only if the document belongs to none of the sixteen
categories. Do not choose it because you are uncertain between two of the
sixteen; in that case choose the more likely of the two.

Respond only with the category name. Do not provide an explanation.

Respond with the following JSON format:
{"prediction": <category name>}

Here is the document text to classify:
\end{lstlisting}
\end{promptbox}

The document text is Amazon Textract OCR, one line of text per detected line, appended after the final line of the prompt as before.
Each model is queried once per document at temperature 0 with an output limit of 8,192 tokens.

Replies are normalized as in the main evaluation, and scored with the following additional rules.
A reply naming \cat{none\_of\_the\_above} is an explicit rejection.
A reply naming a label outside the seventeen offered, such as \cat{legal\_document} or \cat{certificate}, is an implicit rejection: the model has recognized that the document fits none of the categories and said so in its own words, so it is counted as a detection, and Appendix~\ref{app:promptstats} reports the two routes separately.
Where a reply contains several prediction objects, as when a model revises its answer, the last one naming one of the seventeen labels is taken.
A reply that hedges between labels (\texttt{"news\_article" or "presentation"}) does not name a single answer and is unparseable, whereas a single label with a parenthetical gloss counts as that label.
A reply cut off at the output limit is almost always one answer repeated until the budget runs out; it is accepted only if one label accounts for at least 90\% of the repetitions, and is otherwise unparseable.
Unparseable replies count as not rejected, which is the conservative choice for detection.
Pages on which OCR does not find text are excluded from every denominator, since the model is never shown a document.

\subsection*{Input of the Decision Models}

The decision models of Table~\ref{tab:decision-models} do not take a free-text prompt.
Each request consists of three fields, the Textract text of the page, a question and a list of options, and the model assigns a probability to every option, of which the highest is taken as the prediction.
The box below shows the question and the options as they were issued to all five models: the options are the sixteen category names with the underscores replaced by spaces, in the order of the label indices, and no definitions of the categories are given.
A page on which OCR does not find text is passed as the string ``[no text was recognized on this page]''.
For the run of Jev on the page image, the text field is replaced by the page image of the de-identified corpus as we release it, at its stored resolution, and the question reads ``This is one scanned document page. What type of document is it?'', with the same options; the pages outside the range of 280 to 1,568 pixels accepted by the API (one test page and three pages of RVL-CDIP-\emph{N}) are rescaled into that range.

\begin{promptbox}[title={Input of the decision models: question and options}]
\begin{lstlisting}[style=promptstyle]
Text:
    <Textract text of the page>

Question:
    This is the OCR text of one scanned document page. What type of
    document is it?

Options:
    letter
    form
    email
    handwritten
    advertisement
    scientific report
    scientific publication
    specification
    file folder
    news article
    budget
    invoice
    presentation
    questionnaire
    resume
    memo
\end{lstlisting}
\end{promptbox}

\section{Prompt Sensitivity: Statistical Analysis}\label{app:promptstats}

This appendix reports the significance testing behind the zero-shot prompt comparison, and is written to be read on its own.
The three prompts of Appendix~\ref{app:prompts} are issued to each model over the same documents in the same order, so every within-dataset comparison is paired at the document level.
We therefore use McNemar's exact test for pairwise comparisons, which conditions on the discordant documents (those that exactly one prompt classifies correctly), and Cochran's $Q$ as the omnibus test across all three prompts.
The exact binomial form is used in place of the $\chi^2$ approximation because several discordant counts on RVL-CDIP-\emph{N} are small.
Accuracy intervals are Wilson score intervals (Table~\ref{tab:prompt-acc}); $p$-values are corrected within each dataset by the Holm--Bonferroni step-down procedure (Table~\ref{tab:prompt-mcnemar}).
An unpaired two-proportion test would be both invalid and less powerful on this design.
The gpt-oss-120b model of Section~\ref{sec:joint} was evaluated on RVL-CDIP-\emph{N} only; because the analysis below turns on comparing behavior across the two test sets, it is excluded throughout.

Two features of the data govern how the results should be read.
First, statistical power is highly asymmetric: RVL-CDIP contributes 38,377 documents and RVL-CDIP-\emph{N} only 1,001, so paired differences are resolved to roughly $\pm0.35$ percentage points on the former and $\pm1.2$ to $\pm2.4$ on the latter.
A null result on RVL-CDIP-\emph{N} is therefore weak evidence of no effect, and we do not report it as one.
Second, accuracy on RVL-CDIP-\emph{N} is some twenty points higher than on RVL-CDIP, so a given absolute gain is less available there; comparisons across the two datasets are consequently made on the log-odds scale, which is free of that ceiling.

\begin{table}[t]\centering\footnotesize\setlength{\tabcolsep}{3.8pt}
\caption{Zero-shot accuracy (\%) by prompt, with 95\% Wilson intervals as half-widths, and the Cochran $Q$ omnibus test across the three prompts. RVL-CDIP contributes 38,377 documents, RVL-CDIP-\emph{N} 1,001. All eight omnibus tests reject.}\label{tab:prompt-acc}
\begin{tabular}{llccccc}\toprule
\textbf{Test set} & \textbf{Model} & \textbf{Prompt 1} & \textbf{Prompt 2} & \textbf{Prompt 3} & \textbf{\boldmath$Q$} & \textbf{\boldmath$p$} \\ \midrule
RVL-CDIP & Llama 4 Maverick & 67.61\,$\pm$\,0.47 & 73.79\,$\pm$\,0.44 & 67.77\,$\pm$\,0.47 & 1787.9 & $<\!10^{-15}$ \\
 & Mistral 7B & 56.96\,$\pm$\,0.50 & 54.56\,$\pm$\,0.50 & 55.07\,$\pm$\,0.50 & 175.9 & $<\!10^{-15}$ \\
 & gpt-oss-20b & 64.63\,$\pm$\,0.48 & 70.62\,$\pm$\,0.46 & 64.80\,$\pm$\,0.48 & 1642.1 & $<\!10^{-15}$ \\
 & Qwen 3 32B & 62.28\,$\pm$\,0.48 & 63.56\,$\pm$\,0.48 & 58.60\,$\pm$\,0.49 & 853.0 & $<\!10^{-15}$ \\
\midrule
RVL-CDIP-\emph{N} & Llama 4 Maverick & 87.81\,$\pm$\,2.03 & 88.31\,$\pm$\,1.99 & 86.41\,$\pm$\,2.12 & 8.4 & 0.015 \\
 & Mistral 7B & 78.22\,$\pm$\,2.55 & 76.42\,$\pm$\,2.63 & 68.33\,$\pm$\,2.88 & 83.4 & $<\!10^{-15}$ \\
 & gpt-oss-20b & 84.32\,$\pm$\,2.25 & 83.22\,$\pm$\,2.31 & 82.32\,$\pm$\,2.36 & 9.1 & 0.010 \\
 & Qwen 3 32B & 85.41\,$\pm$\,2.19 & 81.62\,$\pm$\,2.40 & 81.82\,$\pm$\,2.39 & 33.5 & $5.4\times10^{-8}$ \\
\bottomrule
\end{tabular}\end{table}
\begin{table}[t]\centering\footnotesize\setlength{\tabcolsep}{3.8pt}
\caption{Pairwise comparisons by McNemar's exact test. $\Delta$ is the accuracy of the second prompt minus the first, in percentage points, with a 95\% paired interval; $b/c$ are the discordant counts. $p_{\text{Holm}}$ is corrected within each test set.}\label{tab:prompt-mcnemar}
\begin{tabular}{llcrcc}\toprule
\textbf{Test set} & \textbf{Model} & \textbf{Contrast} & \textbf{\boldmath$\Delta$ (pp)} & \textbf{\boldmath$b/c$} & \textbf{\boldmath$p_{\text{Holm}}$} \\ \midrule
RVL-CDIP & Llama 4 Maverick & P1 vs.\ P2 & $+6.17$ $[+5.82, +6.53]$ & 1285/3654 & $<\!10^{-15}$~*** \\
 &  & P1 vs.\ P3 & $+0.16$ $[-0.09, +0.40]$ & 1147/1207 & 0.448~\textsc{ns} \\
 &  & P2 vs.\ P3 & $-6.02$ $[-6.37, -5.66]$ & 3631/1322 & $<\!10^{-15}$~*** \\
 & Mistral 7B & P1 vs.\ P2 & $-2.40$ $[-2.80, -2.00]$ & 3530/2608 & $<\!10^{-15}$~*** \\
 &  & P1 vs.\ P3 & $-1.90$ $[-2.24, -1.56]$ & 2575/1847 & $<\!10^{-15}$~*** \\
 &  & P2 vs.\ P3 & $+0.51$ $[+0.12, +0.89]$ & 2682/2876 & 0.029~* \\
 & gpt-oss-20b & P1 vs.\ P2 & $+5.99$ $[+5.64, +6.33]$ & 1194/3491 & $<\!10^{-15}$~*** \\
 &  & P1 vs.\ P3 & $+0.17$ $[-0.11, +0.45]$ & 1470/1534 & 0.448~\textsc{ns} \\
 &  & P2 vs.\ P3 & $-5.82$ $[-6.17, -5.47]$ & 3524/1291 & $<\!10^{-15}$~*** \\
 & Qwen 3 32B & P1 vs.\ P2 & $+1.28$ $[+0.94, +1.62]$ & 2029/2520 & $1.4\times10^{-12}$~*** \\
 &  & P1 vs.\ P3 & $-3.68$ $[-4.01, -3.34]$ & 2841/1430 & $<\!10^{-15}$~*** \\
 &  & P2 vs.\ P3 & $-4.96$ $[-5.31, -4.60]$ & 3399/1497 & $<\!10^{-15}$~*** \\
\midrule
RVL-CDIP-\emph{N} & Llama 4 Maverick & P1 vs.\ P2 & $+0.50$ $[-0.78, +1.78]$ & 19/24 & 1.000~\textsc{ns} \\
 &  & P1 vs.\ P3 & $-1.40$ $[-2.60, -0.19]$ & 26/12 & 0.201~\textsc{ns} \\
 &  & P2 vs.\ P3 & $-1.90$ $[-3.37, -0.42]$ & 38/19 & 0.114~\textsc{ns} \\
 & Mistral 7B & P1 vs.\ P2 & $-1.80$ $[-3.79, +0.20]$ & 61/43 & 0.475~\textsc{ns} \\
 &  & P1 vs.\ P3 & $-9.89$ $[-12.10, -7.68]$ & 118/19 & $<\!10^{-15}$~*** \\
 &  & P2 vs.\ P3 & $-8.09$ $[-10.51, -5.67]$ & 120/39 & $9.5\times10^{-10}$~*** \\
 & gpt-oss-20b & P1 vs.\ P2 & $-1.10$ $[-2.35, +0.15]$ & 26/15 & 0.475~\textsc{ns} \\
 &  & P1 vs.\ P3 & $-2.00$ $[-3.35, -0.65]$ & 34/14 & 0.044~* \\
 &  & P2 vs.\ P3 & $-0.90$ $[-2.18, +0.38]$ & 26/17 & 0.666~\textsc{ns} \\
 & Qwen 3 32B & P1 vs.\ P2 & $-3.80$ $[-5.29, -2.30]$ & 49/11 & $6.8\times10^{-6}$~*** \\
 &  & P1 vs.\ P3 & $-3.60$ $[-4.93, -2.26]$ & 42/6 & $1.0\times10^{-6}$~*** \\
 &  & P2 vs.\ P3 & $+0.20$ $[-1.27, +1.67]$ & 27/29 & 1.000~\textsc{ns} \\
\bottomrule
\end{tabular}\end{table}

\subsection*{Transfer of the Corpus-Grounded Prompt}

Prompt~2 differs from Prompt~3 in that its category definitions are written to the RVL-CDIP corpus itself, naming the sub-types its annotators actually filed under each label.
If that information is what the prompt provides, its benefit should be specific to RVL-CDIP and should not apply to RVL-CDIP-\emph{N}, whose documents were collected separately.
Table~\ref{tab:prompt-transfer} tests this directly, comparing the odds of a correct classification under Prompt~2 against Prompt~1 on each dataset and testing the difference on the log-odds scale.

\begin{table}[t]\centering\footnotesize\setlength{\tabcolsep}{3.8pt}
\caption{Transfer of the Prompt~2 benefit. Relative error reduction is $(\text{err}_{P1}-\text{err}_{P2})/\text{err}_{P1}$; the odds ratio is McNemar's conditional odds ratio $c/b$, above 1 where Prompt~2 helps. The final columns test whether the log-odds benefit differs between the two test sets, a scale on which the ceiling difference does not have an effect.}\label{tab:prompt-transfer}
\begin{tabular}{lccccr@{}lc}\toprule
 & \multicolumn{2}{c}{\textbf{Rel.\ error reduction}} & \multicolumn{2}{c}{\textbf{Odds ratio (P2 vs.\ P1)}} & \multicolumn{3}{c}{\textbf{Interaction}} \\
\cmidrule(lr){2-3}\cmidrule(lr){4-5}\cmidrule(lr){6-8}
\textbf{Model} & \textbf{RVL-CDIP} & \textbf{-\emph{N}} & \textbf{RVL-CDIP} & \textbf{-\emph{N}} & \multicolumn{2}{c}{\textbf{\boldmath$\Delta\log\text{OR}$}} & \textbf{\boldmath$p$} \\ \midrule
Llama 4 Maverick & $+19.1\%$ & $+4.1\%$ & 2.84 & 1.26 & $-0.81$&\,$[-1.42, -0.21]$ & 0.009~** \\
Mistral 7B & $-5.6\%$ & $-8.3\%$ & 0.74 & 0.70 & $-0.05$&\,$[-0.44, +0.35]$ & 0.815~\textsc{ns} \\
gpt-oss-20b & $+16.9\%$ & $-7.0\%$ & 2.92 & 0.58 & $-1.62$&\,$[-2.26, -0.98]$ & $6.4\times10^{-7}$~*** \\
Qwen 3 32B & $+3.4\%$ & $-26.0\%$ & 1.24 & 0.22 & $-1.71$&\,$[-2.37, -1.05]$ & $3.3\times10^{-7}$~*** \\
\bottomrule\end{tabular}\end{table}

The pattern is consistent across three of the four models.
On RVL-CDIP, Prompt~2 removes 19.1\% of Llama~4 Maverick's errors and 16.9\% of gpt-oss-20b's; on RVL-CDIP-\emph{N} the same prompt removes 4.1\% and \emph{adds} 7.0\%.
For Qwen the reversal is largest: a 3.4\% error reduction on RVL-CDIP becomes a 26.0\% increase on RVL-CDIP-\emph{N}, an odds ratio of 0.22 (95\% CI [0.12, 0.43]), which results from 49 documents that Prompt~1 classifies correctly and Prompt~2 does not against 11 in the other direction.
The interaction is significant for all three on the log-odds scale, so it is not an artifact of the differing accuracy ceilings.
Mistral~7B is the exception only in that Prompt~2 never helps it on either dataset, leaving no benefit to fail to transfer.

We interpret this as a measurement of how much of the RVL-CDIP task consists of corpus-specific annotation convention, as distinct from document category.
Definitions that encode those conventions (that \cat{budget} includes check stubs and campaign contribution requests, that \cat{presentation} includes press releases) remove up to a fifth of a model's errors on the corpus that follows them, and do not have a benefit, or are harmful, on documents that do not.
Prompt~3 is the control for this interpretation: it provides definitions of comparable length and specificity written without reference to the corpus, and on RVL-CDIP it is statistically indistinguishable from not providing definitions for both Maverick ($+0.16$ pp, $p_{\text{Holm}}=0.45$) and gpt-oss-20b ($+0.17$ pp, $p_{\text{Holm}}=0.45$).
Those two intervals exclude any effect larger than $0.45$ pp, so this is a tightly bounded null result and not an underpowered one.
The gain from Prompt~2 is thus not a gain from longer or more detailed prompting; it is a gain from knowing the corpus.

A caution on the two datasets' differing precision applies to Table~\ref{tab:prompt-transfer}.
The discordant counts underlying the RVL-CDIP-\emph{N} odds ratios are small (19/24 for Maverick and 26/15 for gpt-oss-20b), and neither of those pairwise comparisons is significant on its own.
Only Qwen's reversal is established by its own test; for the other two the evidence is the interaction, which borrows precision from the much larger RVL-CDIP sample.

\subsection*{Sensitivity to Unparseable Responses}

Throughout, a response that cannot be parsed as one of the sixteen category names is scored as an error.
This is the conservative choice for a benchmark comparison, but it conflates classification accuracy with instruction-following, and the models differ sharply in how often it applies: gpt-oss-20b fails to parse on 9 of 38,377 documents, Mistral~7B on 2,771.
We therefore repeated every comparison on complete cases only, discarding any document that at least one of the three prompts failed to parse (Table~\ref{tab:prompt-parse}).

\begin{table}[t]\centering\footnotesize\setlength{\tabcolsep}{3.8pt}
\caption{Parse-failure sensitivity on RVL-CDIP: $\Delta$ for each contrast with unparseable responses scored as errors, and with the affected documents removed. No comparison changes sign or significance verdict under either treatment. Significance marks here use uncorrected $p$-values, and Table~\ref{tab:prompt-mcnemar} Holm-corrected ones.}\label{tab:prompt-parse}
\begin{tabular}{lrccc}\toprule
\textbf{Model} & \textbf{Dropped} & \textbf{Contrast} & \textbf{\boldmath$\Delta$ scored as error} & \textbf{\boldmath$\Delta$ complete cases} \\ \midrule
Llama 4 Maverick &   278 & P1 vs.\ P2 & $+6.17$~*** & $+5.80$~*** \\
 &       & P1 vs.\ P3 & $+0.16$~\textsc{ns} & $+0.15$~\textsc{ns} \\
 &       & P2 vs.\ P3 & $-6.02$~*** & $-5.65$~*** \\
Mistral 7B &  2771 & P1 vs.\ P2 & $-2.40$~*** & $-2.90$~*** \\
 &       & P1 vs.\ P3 & $-1.90$~*** & $-1.70$~*** \\
 &       & P2 vs.\ P3 & $+0.51$~** & $+1.20$~*** \\
gpt-oss-20b &     9 & P1 vs.\ P2 & $+5.99$~*** & $+5.98$~*** \\
 &       & P1 vs.\ P3 & $+0.17$~\textsc{ns} & $+0.17$~\textsc{ns} \\
 &       & P2 vs.\ P3 & $-5.82$~*** & $-5.81$~*** \\
Qwen 3 32B &   876 & P1 vs.\ P2 & $+1.28$~*** & $+0.79$~*** \\
 &       & P1 vs.\ P3 & $-3.68$~*** & $-4.12$~*** \\
 &       & P2 vs.\ P3 & $-4.96$~*** & $-4.91$~*** \\
\bottomrule\end{tabular}\end{table}

No comparison changes sign, and none crosses a significance boundary.
The largest movement is Qwen's Prompt~1 versus Prompt~2 contrast, which decreases from $+1.28$ to $+0.79$ pp and remains significant.
Mistral~7B's accuracy increases by roughly three points on every prompt once its unparseable responses are removed, but the ordering of its prompts is unchanged and its preference for Prompt~1 over Prompt~2 \emph{widens}, from $-2.40$ to $-2.90$ pp.
Its atypical behavior is therefore a property of how it classifies under corpus-grounded definitions, not of how often it fails to answer in the requested format.

\subsection*{Out-of-Scope Detection on RVL-CDIP-\emph{O}}

The rejection variants of the three prompts (Appendix~\ref{app:prompts}) were issued to each model over RVL-CDIP-\emph{O}, none of whose documents belongs to any of the sixteen categories, so the correct reply is always a rejection.
The share of documents a model rejects is therefore its recall on the out-of-scope class, which we call its detection rate.
Of the 2,915 documents, 6 do not yield OCR text and are excluded, leaving 2,909.
As before, every prompt sees the same documents in the same order, so the comparisons are paired and use the same tests, with Holm correction across the twelve pairwise contrasts.
A model that rejects every document would score 100\% here, and we did not measure how often the models reject in-distribution documents under the rejection variants.
The rates in this subsection therefore show how readily each model rejects a document, and they do not measure how well it separates out-of-scope documents from in-distribution ones.

\begin{table}[t]\centering\footnotesize\setlength{\tabcolsep}{3.8pt}
\caption{Out-of-scope detection on RVL-CDIP-\emph{O} (\%): the share of its 2,909 documents that a model rejects, explicitly or with a label outside the seventeen offered, with 95\% Wilson intervals as half-widths, and the Cochran $Q$ omnibus test across the three prompts. Unparseable replies count as not rejected.}\label{tab:oos-detect}
\begin{tabular}{lccccc}\toprule
\textbf{Model} & \textbf{Prompt 1} & \textbf{Prompt 2} & \textbf{Prompt 3} & \textbf{\boldmath$Q$} & \textbf{\boldmath$p$} \\ \midrule
Llama 4 Maverick & 3.09\,$\pm$\,0.63 & 6.53\,$\pm$\,0.90 & 8.04\,$\pm$\,0.99 & 129.7 & $<\!10^{-15}$ \\
Mistral 7B & 13.13\,$\pm$\,1.23 & 8.25\,$\pm$\,1.00 & 20.14\,$\pm$\,1.46 & 285.8 & $<\!10^{-15}$ \\
gpt-oss-20b & 37.19\,$\pm$\,1.76 & 52.35\,$\pm$\,1.81 & 59.02\,$\pm$\,1.79 & 575.0 & $<\!10^{-15}$ \\
Qwen 3 32B & 62.22\,$\pm$\,1.76 & 50.74\,$\pm$\,1.82 & 64.15\,$\pm$\,1.74 & 282.0 & $<\!10^{-15}$ \\
\bottomrule\end{tabular}\end{table}

Detection divides the models into two groups (Table~\ref{tab:oos-detect}).
Two of them, gpt-oss-20b and Qwen~3~32B, reject between 37.2\% and 64.2\% of documents depending on the prompt; Llama~4 Maverick and Mistral~7B reject between 3.1\% and 20.1\%, assigning the great majority of out-of-scope documents to one of the sixteen categories.
Prompt~3 yields the highest detection for all four models, every omnibus test rejects, and all twelve pairwise contrasts remain significant after correction (Table~\ref{tab:oos-mcnemar}).
Because we did not measure false rejections, these results do not show whether the generic definitions of Prompt~3 help a model recognize documents outside the label set or only make the model more ready to reject.

\begin{table}[t]\centering\footnotesize\setlength{\tabcolsep}{3.8pt}
\caption{Pairwise comparisons of out-of-scope detection by McNemar's exact test. $\Delta$ is the detection rate of the second prompt minus the first, in percentage points, with a 95\% paired interval; $b/c$ are the discordant counts. $p_{\text{Holm}}$ is corrected across the twelve contrasts.}\label{tab:oos-mcnemar}
\begin{tabular}{lcrcc}\toprule
\textbf{Model} & \textbf{Contrast} & \textbf{\boldmath$\Delta$ (pp)} & \textbf{\boldmath$b/c$} & \textbf{\boldmath$p_{\text{Holm}}$} \\ \midrule
Llama 4 Maverick & P1 vs.\ P2 & $+3.44$ $[+2.59, +4.29]$ & 31/131 & $3.9\times10^{-15}$~*** \\
 & P1 vs.\ P3 & $+4.95$ $[+4.09, +5.81]$ & 13/157 & $<\!10^{-15}$~*** \\
 & P2 vs.\ P3 & $+1.51$ $[+0.63, +2.39]$ & 64/108 & 0.002~** \\
Mistral 7B & P1 vs.\ P2 & $-4.88$ $[-6.17, -3.59]$ & 257/115 & $4.0\times10^{-13}$~*** \\
 & P1 vs.\ P3 & $+7.01$ $[+5.63, +8.40]$ & 117/321 & $<\!10^{-15}$~*** \\
 & P2 vs.\ P3 & $+11.89$ $[+10.52, +13.27]$ & 57/403 & $<\!10^{-15}$~*** \\
gpt-oss-20b & P1 vs.\ P2 & $+15.16$ $[+13.40, +16.92]$ & 155/596 & $<\!10^{-15}$~*** \\
 & P1 vs.\ P3 & $+21.83$ $[+20.09, +23.57]$ & 86/721 & $<\!10^{-15}$~*** \\
 & P2 vs.\ P3 & $+6.67$ $[+4.97, +8.37]$ & 229/423 & $1.1\times10^{-13}$~*** \\
Qwen 3 32B & P1 vs.\ P2 & $-11.48$ $[-13.28, -9.68]$ & 544/210 & $<\!10^{-15}$~*** \\
 & P1 vs.\ P3 & $+1.93$ $[+0.37, +3.48]$ & 240/296 & 0.017~* \\
 & P2 vs.\ P3 & $+13.41$ $[+11.83, +14.99]$ & 106/496 & $<\!10^{-15}$~*** \\
\bottomrule\end{tabular}\end{table}

The two routes to rejection are used very differently (Table~\ref{tab:oos-routes}).
Llama~4 Maverick and gpt-oss-20b never answer outside the seventeen labels, so every detection they make is an explicit rejection.
Mistral~7B does the reverse: labels of its own, most often \cat{legal\_document} and \cat{educational\_material}, account for 53--64\% of its detections, and it names the option it was offered less often than it invents one.
Qwen~3~32B uses both, with its own labels a minority of its detections (4--8\%).
Unparseable replies occur only for Mistral~7B (3.6--7.5\% of documents) and Llama~4 Maverick (at most 4 documents per prompt).
Counting them as detections instead would raise Mistral~7B's rate to at most 27.6\%, which leaves the grouping unchanged.

\begin{table}[t]\centering\footnotesize\setlength{\tabcolsep}{3.8pt}
\caption{Routes to rejection on RVL-CDIP-\emph{O}, as a percentage of documents: an explicit \cat{none\_of\_the\_above}, a label outside the seventeen offered, and unparseable replies, which count as not rejected. Detection in Table~\ref{tab:oos-detect} is the sum of the first two columns.}\label{tab:oos-routes}
\begin{tabular}{lcrrr}\toprule
\textbf{Model} & \textbf{Prompt} & \textbf{Explicit} & \textbf{Own label} & \textbf{Unparseable} \\ \midrule
Llama 4 Maverick & 1 & 3.09 & 0.00 & 0.03 \\
 & 2 & 6.53 & 0.00 & 0.14 \\
 & 3 & 8.04 & 0.00 & 0.07 \\
Mistral 7B & 1 & 4.78 & 8.35 & 3.64 \\
 & 2 & 3.88 & 4.37 & 4.43 \\
 & 3 & 8.42 & 11.72 & 7.46 \\
gpt-oss-20b & 1 & 37.19 & 0.00 & 0.00 \\
 & 2 & 52.35 & 0.00 & 0.00 \\
 & 3 & 59.02 & 0.00 & 0.00 \\
Qwen 3 32B & 1 & 57.24 & 4.98 & 0.00 \\
 & 2 & 48.30 & 2.44 & 0.00 \\
 & 3 & 61.57 & 2.58 & 0.00 \\
\bottomrule\end{tabular}\end{table}

\section{Models and Training Details}\label{app:training}

\begin{table}[tbp]
  \centering\footnotesize\setlength{\tabcolsep}{4pt}
  \caption{Training setup of the eleven fine-tuned models of
           Section~\ref{sec:joint}. Every model is trained for the full number
           of epochs and the epoch with the highest validation accuracy is
           kept.}
  \label{tab:training-details}
  \resizebox{\textwidth}{!}{
  \begin{tabular}{llllccc}
    \toprule
    \textbf{Model} & \textbf{Initial weights} & \textbf{Optimizer} & \textbf{Schedule} & \textbf{LR} & \textbf{Batch} & \textbf{Epochs} \\
    \midrule
    Six CNNs        & ImageNet (torchvision)             & SGD, momentum 0.9 & polynomial, power 0.5       & $10^{-3}$         & 64 & 30 \\
    DiT-base        & \texttt{microsoft/dit-base}          & AdamW             & cosine, 10\% warmup         & $3\times10^{-5}$  & 32 & 10 \\
    BERT-base       & \texttt{bert-base-uncased}           & AdamW             & linear, 10\% warmup         & $5\times10^{-5}$  & 16 & 5 \\
    RoBERTa-base    & \texttt{roberta-base}                & AdamW             & linear, 10\% warmup         & $5\times10^{-5}$  & 16 & 5 \\
    LayoutLM-base   & \texttt{microsoft/layoutlm-base-uncased} & AdamW         & linear, 10\% warmup         & $5\times10^{-5}$  & 16 & 5 \\
    LayoutLMv3-base & \texttt{microsoft/layoutlmv3-base}   & Adam              & constant                    & $10^{-5}$         & 64 & 10 \\
    \bottomrule
  \end{tabular}}
\end{table}

Table~\ref{tab:training-details} lists the training setup of the fine-tuned models, each of which is trained with the settings of its publication or of our earlier work, and only the label set and the pages change between the conditions of Section~\ref{sec:joint}.
The input of the image models is the page converted to grayscale, resized to 224 by 224 pixels and replicated to three channels, and these models are trained with a cross-entropy loss and mixed precision; the six convolutional networks (AlexNet, GoogLeNet, SqueezeNet, VGG-16, ResNet-50 and ResNeXt-50) are trained with the same settings.
The input of BERT and RoBERTa is the Textract text of the page truncated to 512 tokens, that of LayoutLM is the Textract words with their bounding boxes, truncated to 512 tokens, with gradients clipped at a norm of 1.0 in training, and that of LayoutLMv3 is the same words and boxes together with the page image; LayoutLMv3 is trained with mixed precision on two GPUs with 32 pages each.
Each of the eleven models is trained with the seeds 42, 1 and 2 for every combination of label set and code condition.

The TF-IDF features are computed on lowercased word unigrams and bigrams that occur in at least two training pages, limited to the 50,000 most frequent and with sublinear term-frequency scaling, and the SVM is a one-vs-rest linear SVM, trained with scikit-learn \citep{scikit-learn}, whose regularization constant (0.1, 0.3, 1 or 3) is chosen on validation accuracy.
For the SVM on text and image features the SigLIP embedding (\texttt{google/siglip-so400m-patch14-384}) is standardized, scaled so that its norm is comparable to that of the TF-IDF vector and multiplied by a weight, and the weight (0.5, 1 or 2) and the regularization constant are chosen together on validation accuracy.

The public checkpoints are \texttt{naver-clova-ix/donut-base-finetuned-rvlcdip} for Donut and the ConvNeXt-base weights released by the authors for DocXClassifier.
The zero-shot image models are \texttt{openai/clip-vit-large-patch14-336}, \texttt{kakaobrain/align-base}, \texttt{google/siglip-so400m-patch14-384} and \texttt{google/siglip2-so400m-patch16-naflex}, each of which computes the similarity between the page image and the sixteen prompts ``This is a photo of \emph{category}.'' and predicts the category with the highest similarity.
Jev~1.13 \citep{typesafe-jev-2026} was queried through the API of its developer in October 2026 with the full Textract text of the page and the question and options given in Appendix~\ref{app:prompts}, and the option to which it assigns the highest probability is taken as its prediction.
We queried the same API in October 2026 with the page images, an input that the API offers as a preview feature, and sent only the de-identified images that we release.
The four large language models were queried through Amazon Bedrock in April 2026 with each of the three prompts of Appendix~\ref{app:prompts} (Prompt~1 in the main tables) followed by the Textract text of the page, and the category named in the reply is taken as the prediction.
Bespoke-Nimble-9B \citep{nimble2026}, a LoRA adapter on Qwen3.5-9B that scores the answer options without generating text, was run on our GPUs with the inference code of its developers, the same text and the same options, and the option with the highest score is taken as its prediction; no page exceeds its input limit of 8,192 tokens.
Strands Decider~2B \citep{strands-decider-2026}, which replaces the language-model head of Qwen3.5-2B with a pointer head that scores each option against the answer position, was run on our GPUs in its release \texttt{StrandsAgents/strands-decider-2B-hobson-v21} with the \texttt{strands-decider} package (version 0.1.0), one request per page with the same text and options, and the option with the highest score is taken as its prediction.
Laya \citep{laya-2026} pairs a ModernBERT-large encoder with a decision head that scores each option at its own mask token; we ran its English model on our GPUs with the \texttt{laya} package (version 0.4.0), the same text and the same options, and the option with the highest score is taken as its prediction, with the text truncated to the model's input limit of 512 tokens.
clef-flash \citep{cloudflare-clef-2026} adds low-rank adapters and a routing head to a frozen Qwen3.5-9B and scores all options in a single forward pass over the input; we ran its release \texttt{Cloudflare/clef-flash} on our GPUs with the code of its repository, on the same text and options and without the page image, and the option with the highest score is taken as its prediction.

\end{document}